\documentclass{article}

\usepackage[nonatbib, final]{neurips_2024}

\usepackage[utf8]{inputenc} 
\usepackage[T1]{fontenc}    
\usepackage{url}            
\usepackage{booktabs}       
\usepackage{amsfonts}       
\usepackage{nicefrac}       
\usepackage{microtype}      
\usepackage{xcolor}         

\usepackage{hyperref}
\hypersetup{pagebackref=true,breaklinks=true,letterpaper=true,colorlinks,bookmarks=false,citecolor=cyan}

\usepackage{mwe} 
\usepackage{tabularx}
\usepackage{amsmath}
\usepackage{multirow}
\usepackage{siunitx}
\usepackage{makecell}
\usepackage{adjustbox}
\usepackage{cuted}
\usepackage{caption}

\title{Face2QR: A Unified Framework for Aesthetic, Face-Preserving, and Scannable QR Code Generation}

\makeatletter
\def\thanks#1{\protected@xdef\@thanks{\@thanks
        \protect\footnotetext{#1}}}
\makeatother

\author{
    Xuehao Cui$^*$, Guangyang Wu$^*$, Zhenghao Gan, Guangtao Zhai, Xiaohong Liu$^{\dag}$\\
  Shanghai Jiao Tong University \\
  \texttt{\{cavosamir, wu.guang.young, ganzhenghao,}\\
  \texttt{zhaiguangtao, xiaohongliu\}@sjtu.edu.cn}\\
  \thanks{$^{*}$ Equal contribution.}
  \thanks{$^{\dag}$ Corresponding author.}
}

\begin{document}

\maketitle

\begin{abstract}
  Existing methods to generate aesthetic QR codes, such as image and style transfer techniques, tend to compromise either the visual appeal or the scannability of QR codes when they incorporate human face identity. Addressing these imperfections, we present Face2QR—a novel pipeline specifically designed for generating personalized QR codes that harmoniously blend aesthetics, face identity, and scannability. Our pipeline introduces three innovative components. First, the ID-refined QR integration (IDQR) seamlessly intertwines the background styling with face ID, utilizing a unified Stable Diffusion (SD)-based framework with control networks. Second, the ID-aware QR ReShuffle (IDRS) effectively rectifies the conflicts between face IDs and QR patterns, rearranging QR modules to maintain the integrity of facial features without compromising scannability. Lastly, the ID-preserved Scannability Enhancement (IDSE) markedly boosts scanning robustness through latent code optimization, striking a delicate balance between face ID, aesthetic quality and QR functionality. In comprehensive experiments, Face2QR demonstrates remarkable performance, outperforming existing approaches, particularly in preserving facial recognition features within custom QR code designs. Codes are available at \url{https://github.com/cavosamir/Face2QR}.
\end{abstract}
\section{Introduction}\label{sec:intro}

Quick Response (QR) codes, due to their capability to store a substantial amount of data and their ease of accessibility through basic camera devices, have become an exceedingly widespread medium for the representation of information in the digital era~\cite{garateguy2014qr,stylized,picode,racode,su2021artcoder}. With the wide application of QR codes in social context, there has been increasing needs for customizing QR codes to include \textbf{personal identity} and \textbf{aesthetic allure}. 
However, such needs cannot be fulfilled by the dull appearance of common QR codes, which contain only black and white modules. 

With the widespread application of QR codes across diverse fields, related technologies are also developing at a rapid pace. While techniques employing image transformation~\cite{chu2013halftone,qart,garateguy2014qr,artup} and style transferring~\cite{su2021artcoder,stylized} can partially retain facial features, their perceptual quality and aesthetic adaptability are limited. On the other hand, generative model-based approaches~\cite{antfu,wu2024text2qr} can produce QR code images of superior quality and diversity, yet they pose challenges in controlling the generated content, particularly in preserving human facial characteristics
. To address these limitations and ensure faithful preservation of face identity within a customized and scannable QR code image, we introduce a novel pipeline, named Face2QR. This approach achieves a balanced compromise between face ID preservation, aesthetic appeal, and scannability for QR code images.

\begin{figure}[t]
\vspace{15pt}
\includegraphics[width=\linewidth]{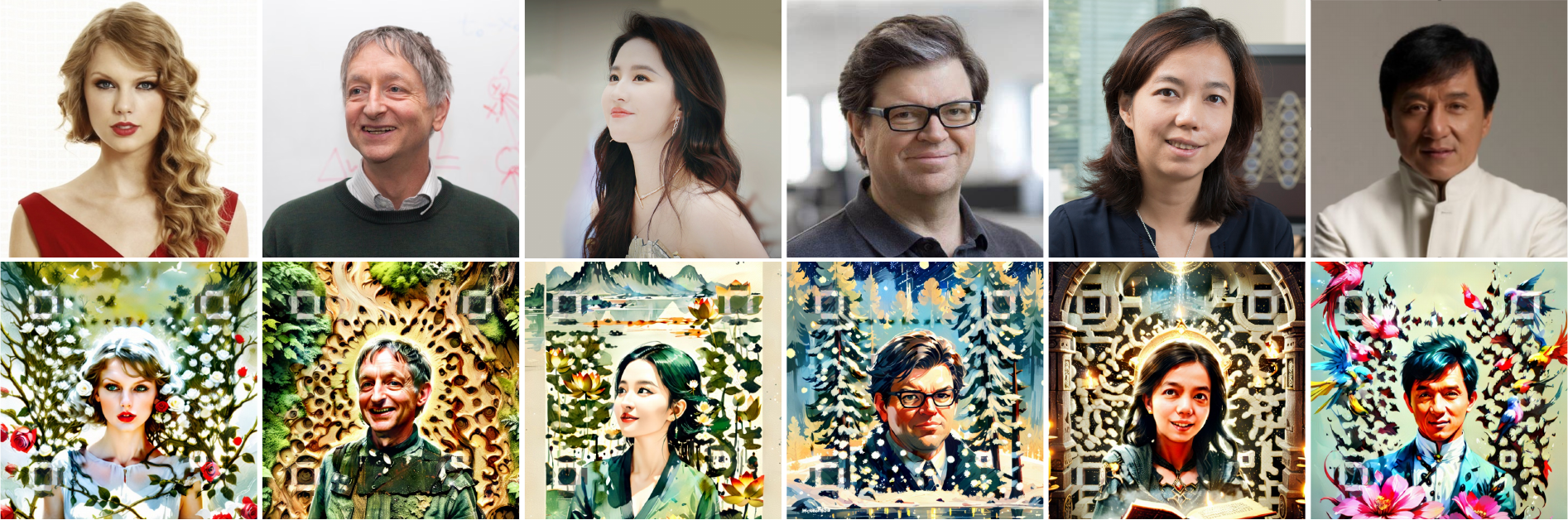}
\caption{Face images (first row) and QR code images (second row) generated by Face2QR. Our QR codes not only faithfully maintain face ID, but also showcase remarkable scanning resilience and aesthetic quality.}
\vspace{-20pt}
\label{teaser}
\end{figure}

The primary challenges lie in effectively integrating three key aspects: face ID, aesthetic quality, and scanable QR pattern, which can be summarized as follows: 
\textbf{(1) Combination of face ID and background.} Achieving a harmonious balance between strict facial ID preservation and diverse customized background styles within a unified pipeline presents a notable challenge. Methods reliant on style transfer~\cite{su2021artcoder,stylized} often yield facial textures that appear unnatural, while those based on image transfer~\cite{chu2013halftone,qart,garateguy2014qr,artup} may introduce visible artifacts in the facial region;
\textbf{(2) Conflict between face ID and QR code pattern.} While prior generative model-based techniques~\cite{wu2024text2qr} have demonstrated the ability to control the QR code pattern using QR blueprints, they have struggled to exclude these patterns from the facial region, resulting in unnatural shadows and undesirable artifacts. However, directly removing these patterns from the facial region can make the image unscannable. Thus, achieving a balance between maintaining visual quality in the facial region and ensuring the correctness of the QR pattern presents a formidable obstacle;
\textbf{(3) Balance between aesthetics and scannability.} As revealed in~\cite{wu2024text2qr}, generated images often exhibit a tendency towards being unscannable, necessitating enhancements to their scannability through post-processing. However, globally adjusting brightness can compromise the natural appeal of the facial region. Thus, novel region-based enhancement methods are worth considering to address this challenge.

To address these challenges, the proposed Face2QR pipeline offers a solution for generating personalized QR codes that strike a balance between aesthetics, facial ID preservation, and scannability. We propose ID-refined QR integration (IDQR) to seamlessly combine background and face ID, and ID-aware QR ReShuffle (IDRS) to solve the conflict between face ID and QR code pattern. Specifically, IDQR applies a unified SD-based framework to ensure that the generated images have a uniform style. Stable Diffusion (SD) models are guided by two sets of control networks, corresponding to face refinement and QR pattern respectively, to achieve separate control in face region and background. IDRS utilizes the flexibility of QR code encoding and reshuffles the modules to make the QR pattern compatible with face ID. Finally, we use ID-preserved Scannability Enhancement (IDSE) to enhance scan robustness through latent code optimization, achieving a new trade-off between face ID, aesthetics and scannability. Figure~\ref{teaser} shows the QR images generated by Face2QR. It is worth noting that the generated QR images are not only the reprints of the provided references, but also have improved aesthetics to align with the generated background, guided by text prompts (\textit{e.g.}, the style and color of clothes have been  adjusted accordingly). 

The contributions of this work can be summarized as:

\noindent$\bullet$ We propose a novel pipeline that holistically integrates aesthetic appealing, facial ID, and scannability to deliver a customized personal representation in QR codes.

\noindent$\bullet$ We introduce the ID-refined QR integration (IDQR) for seamlessly integrating face ID with background, the ID-aware QR ReShuffle (IDRS) for solving conflicts  between face ID and QR pattern, and the ID-preserved Scannability Enhancement (IDSE) for optimizing scan robustness while maintaining face ID and aesthetic quality.

\noindent$\bullet$ Our Face2QR achieves the State-Of-The-Art (SOTA) performance in generating the ID-preserved aesthetic QR codes, compared with previous methods.

\section{Related Works}\label{sec:related}

\paragraph{Quick Response (QR) Code.} 
As QR codes emerging as a key connector between real and virtual worlds, there is increasing interest in enhancing the visual appeal of normally monochromatic QR codes. Halftone QR codes~\cite{chu2013halftone} offers a design where QR code patterns align with a given image in a thematically cohesive manner. QRImage and Artup ~\cite{garateguy2014qr, artup} explore ways to encode colorful imagery within a QR code. Other advances~\cite{su2021q,su2021artcoder} have been made in artistic style transfer to increase aesthetic appearance of QR codes. To further customize QR code and obscure overt QR code markers, Chen et al.~\cite{chen2018robust,racode,ma2023oacode} crafted encoding schemes that consider human visual perception, thus making these patterns less intrusive. TPVM~\cite{TPVM} has gone further to conceal QR codes within video content, exploiting the discrepancies in frame capture rates between human vision and digital screens. Similarly, advancements have sought to keep data imperceptible yet accessible through various stealth mechanisms~\cite{fang2018screen,TERA,stegastamp,RIHOOP,wengrowski2019light,jia2022learning}.

\paragraph{Diffusion Based Models.}
Image manipulation and generation techniques powered by deep learning have made strides in recent years~\cite{RPSRMD,pred,fastllve,accflow,stsr,vfiformer,vfigdc}, with generative models being at the forefront of this development~\cite{a3gan,ciagan,dalle,glide}. Novel diffusion-based models such as GLIDE~\cite{glide}, DALLE-2~\cite{dalle}, and Latent Diffusion models~\cite{latent_diffusion} have come into prominence. Notably, the Stable Diffusion model~\cite{latent_diffusion} moves the denoising steps into the latent dimension of a variational autoencoder, which significantly optimizes the generation process in terms of data volume and training time. In parallel, new research has introduced various techniques for modulating the diffusion process. Structural condition interventions have been successfully implemented by ControlNet~\cite{controlnet} and T2I-Adapter~\cite{t2i_adapter}. 
On a different note, BLIP-Diffusion~\cite{blip} and SeeCoder~\cite{SeeCoder} have made progress on steering generative outcomes based on stylistic aspects.

\paragraph{Identity Preserved Generative Models.}
In the field of ID-preserving image generation, research focuses on maintaining semantic face attributes while generating images that have wide real-world applications. Studies have generally split between techniques requiring test-time fine-tuning, such as Low-Rank Adaptation~\cite{hu2021lora}, and newer optimization-free methods such as Face0~\cite{valevski2023face0}, PhotoMaker~\cite{li2023photomaker}, and FaceStudio~\cite{yan2023facestudio}, which integrate facial embeddings into the generation process in different ways. While techniques like IP-Adapter~\cite{ip_adapter} strive for identity consistency by using embeddings from recognition models, they face challenges in compatibility with pre-trained models and ensuring facial fidelity. Most recent work like InstantID~\cite{wang2024instantid} use a pluggable module that does not demand fine-tuning and can work seamlessly with available pre-trained diffusion models to achieve high-quality face preservation in generated images.
\section{Method}\label{sec:method}

\begin{figure*}[ht]
    \centering
    \includegraphics[width=\linewidth]{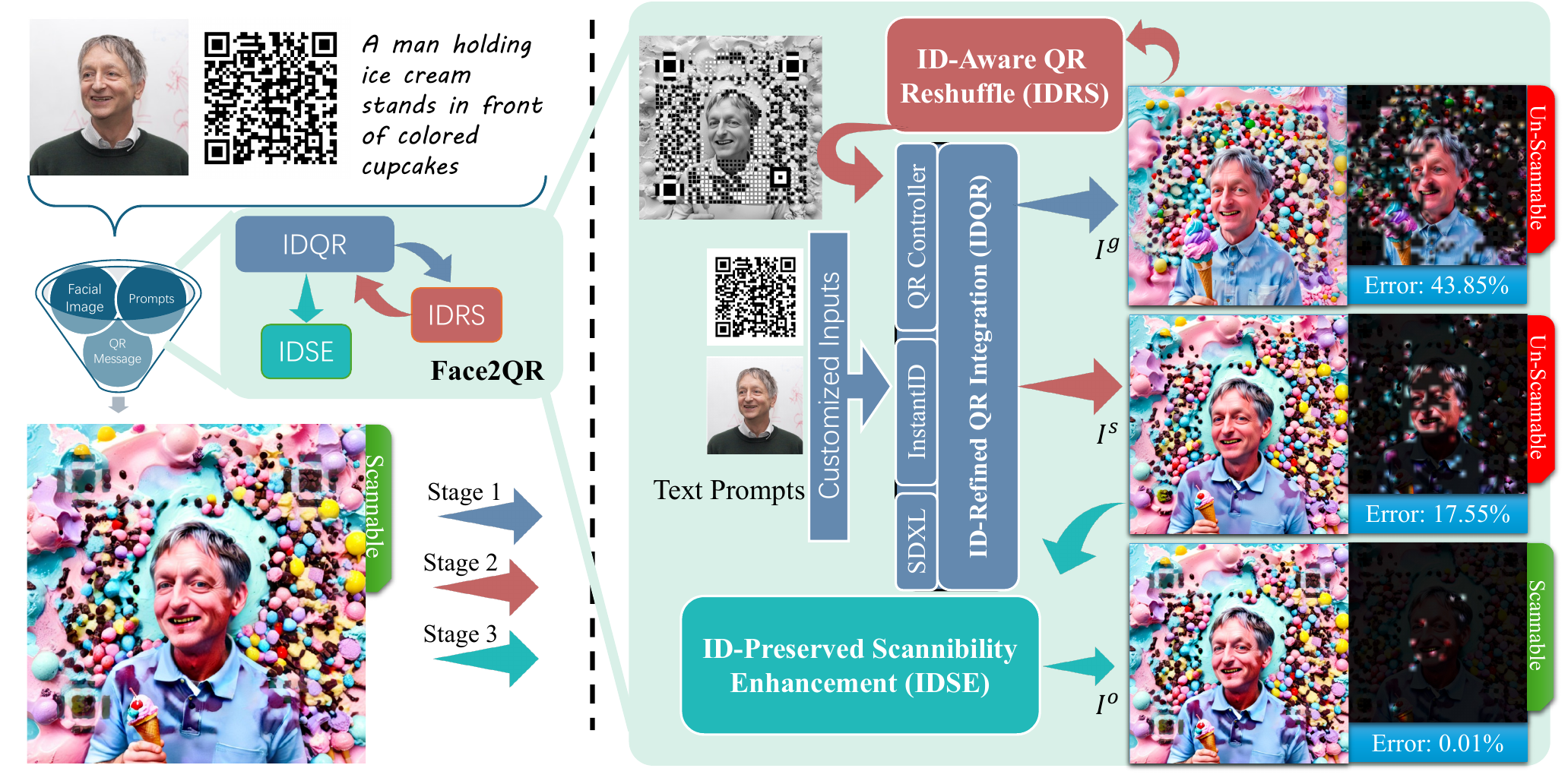}
    \caption{The pipeline of Face2QR is a training-free process for generating ID-consistent and scannable QR code images. Our pipeline has three stages, represented by blue, red, and green arrows. The IDRS module resolves conflicts between human identity and QR patterns during the control process, while the IDSE module reduces coding errors to ensure the output is scannable.
    }
    \label{fig:overall}
\vspace{-2mm}
\end{figure*}

The overall structure of Face2QR is shown in Figure~\ref{fig:overall}. The pipeline unfolds through three stages, represented by blue, red and green arrows.
Given a user-customized face image $f$, QR Code $m$, text prompts $c$ and random noise $z_0$, the first stage uses the ID-refined QR integration (IDQR) module to generate an initial QR image $I^g$. The IDQR module includes a pre-trained SDXL model (denoted as $\mathcal{SD}$), an InstantID~\cite{wang2024instantid} network (denoted as $C_{id}$) and a QR Controller~\cite{wu2024text2qr} (denoted as $C_{qr}$). Stage 1 can be expressed as:
\begin{equation}
\label{eq:first}
     I^g = \mathcal{SD}(c, z_0 | \mathcal{C}_{qr}(m, c, z_0), \mathcal{C}_{id}(f, c, z_0)).
\end{equation}
The InstantID network preserves the facial identity information in the generated images, while the QR Controller guides the luminance distribution of the images. 

However, as shown in Figure~\ref{fig:overall}, the initial output image from the first stage contains a significant error rate (over 43\%). This issue arises from the inherit conflict between two control signals: the foreground face information and the background QR patterns, which are incompatible in the center regions. These conflicts lead to unavoidable QR code errors, presenting a core challenge in our pipeline. To address this, we design the ID-aware QR ReShuffle (IDRS) module to harmonize these conflicts and regenerate the image using a fine-grained QR blueprint $I_b$. As illustrated in Figure~\ref{fig:overall}, this reduces the error rate by more than half. Finally, we use the ID-preserved Scannability Enhancement (IDSE) module to refine the result $I^s$ in latent space, further improving its scanning robustness without compromising the overall visual quality. In the following, we introduce the second and third stages in details.

\subsection{ID-Aware QR ReShuffle}
As revealed in~\cite{wu2024text2qr}, a fine-grained QR blueprint can effectively control the generator. To resolve control conflicts in the facial region, we design a novel blueprint that makes facial information and QR patterns compatible. By leveraging the dynamic characteristics of QR code encoding, we can adaptively rearrange the QR modules. Specifically, we maintain the brightness distribution of the facial region and reshuffle the remaining black and white modules accordingly.

First of all, we binarize $I^g\in \mathbb{R}^{H\times W \times 3}$ into module-wise binary information $\mathbf{E}\in \mathbb{R}^{n^2}$. By dividing $I^g$ into $n\times n$ modules each of size $a\times a$, and let $\theta_j$ be the set of pixel coordinates of the $j$-th module in $I^g$, the extracted information code $\mathbf{E}$ is given by:
\begin{gather}
    \mathbf{E}_j = 
    \begin{cases}
        0, & \text{if } \text{avg}(I^g(\theta_j)) < \tau, \\
        1, & \text{if } \text{avg}(I^g(\theta_j)) \geq \tau, 
    \end{cases}
\end{gather}
where $\text{avg}(\cdot)$ denotes the mean pixel value of the squared patch of size $a\times a$. The binarization uses a threshold $\tau$, typically set to 128 for a total of 256 grayscale levels. 

As shown in Figure~\ref{fig:IDSE} (left), the binarized QR code is un-scannable due to a significant error rate. To address this, we fix the facial and marker region within $\mathbf{E}$, then rearrange the remaining codes to align with the encoded information. To locate the facial region, we use a pre-trained face recognition model to obtain the binary facial mask $M_f\in\mathbb{R}^{H\times W}$. Let the set $\Delta_f=\{j\text{ }|\text{ }\text{avg}(M_f(\theta_j)) = 1\}$ represent the indices of information codes in $\mathbf{E}$ that correspond to the facial region, and let the $\Delta_m$ represent the indices of marker codes. Our goal is to obtain a new information code $\Tilde{\mathbf{E}}$ which is partially modified from $\mathbf{E}$ to make the QR decoder $\mathbb{D}$ extract lossless information:
\begin{align}
&\min | \mathbb{D}(\Tilde{\mathbf{E}}) -  \mathbb{D}(m) |,\\
&\text{s.t.}\quad  \Tilde{\mathbf{E}}_j = \mathbf{E}_j, \text{for } j\in \Delta_f \cup \Delta_m,
\end{align}

To ensure the resultant $\Tilde{\mathbf{E}}$ can be decoded to the target message, aligning with original QR code $m$, we re-generate the error correction code~\cite{reed1960polynomial} in $\Tilde{\mathbf{E}}$.

Afterwards, we expand the binary information of $\Tilde{\mathbf{E}}$ to image space. We use adaptive-halftone to combine the texture information of $I^g$ with binary code information of $\Tilde{\mathbf{E}}$ in an adaptive manner, resulting in the blueprint $I_b\in\mathbb{R}^{H\times W}$. Note that we leave the facial region unmodified to maintain rich facial features without compromising scanning robustness. The resultant blueprint $I_b$ is then fed into $\mathcal{SD}$ for the second generation:
\begin{equation}
     I^s = \mathcal{SD}(c, z_0 | \mathcal{C}_{qr}(I_b, c, z_0), \mathcal{C}_{id}(f, c, z_0)).
\end{equation}

Compared with the first generation in Equation~\ref{eq:first}, both controllers in stage 2 include facial information to mitigate conflicts. As shown in Figure~\ref{fig:overall}, the result of stage 2 reduces errors by more than half compared to stage 1, while consistently preserving face identity information.

\subsection{Scannability Enhancement}

\begin{figure}
    \centering
    \includegraphics[width=\linewidth]{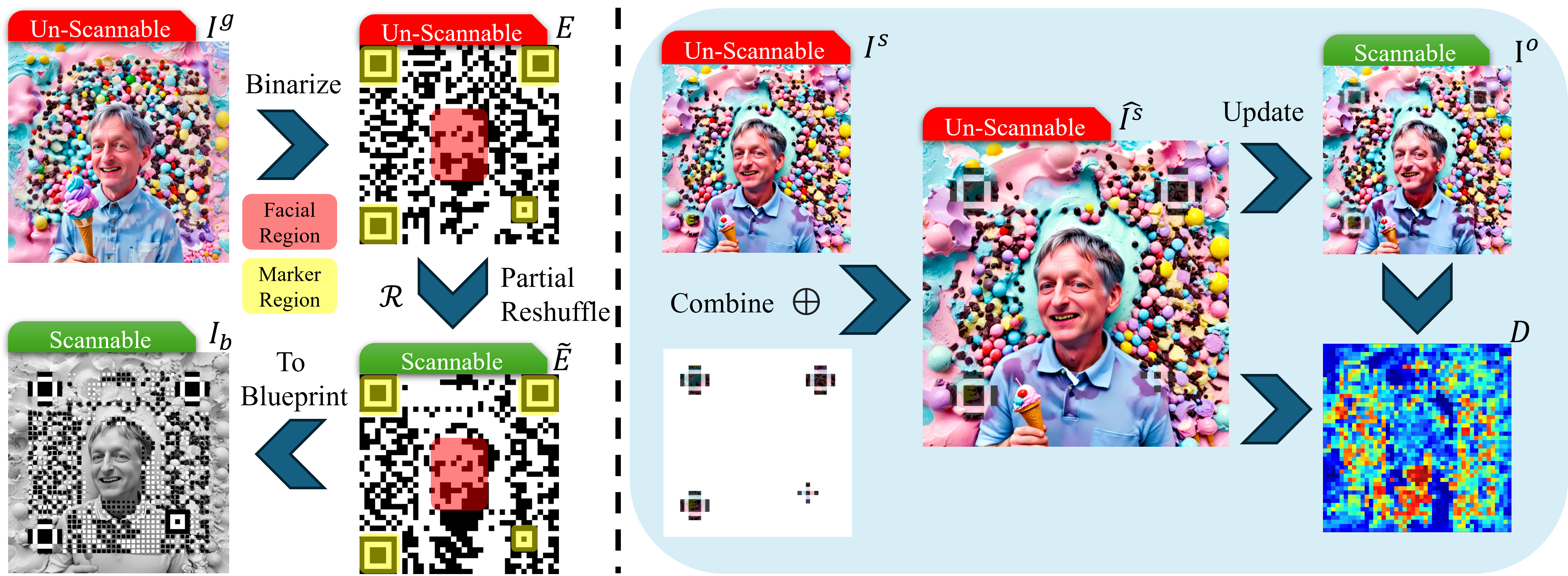}
    \caption{Illustration of IDRS (left) and IDSE (right). In IDRS, we maintain the information codes within the face and marker regions (red and yellow masks) and remap the remaining modules accordingly. In IDSE, we strengthen the finder and alignment pattern, and update in latent space using adaptive loss to enhance scannability. 
    Visualization $D$ shows the difference between $I^o$ and $\hat{I^s}$. Compared with uniform loss, adaptive loss modifies face region more gently.}\label{fig:IDSE}
    \vspace{-5pt}
\end{figure}

The resultant QR image $I^s$ from stage 2 contains a certain QR pattern and consistently reveals face identity, but it is still unscannable by common QR decoders.
In this part, we design the ID-Preserved Scannability Enhancement (IDSE) module to achieve the following two goals: 1) minimize modifications to the QR image (especially for facial region) to ensure its scannability; 2) enhance the marker region to better harmonize it without compromising scanning robustness. As illustrated in Figure~\ref{fig:IDSE} (right), we first strengthen the finder and alignment pattern of $I^s$, and then refine it using dynamic code loss to reach a harmonious balance between face ID, visual appeal and scannability.

\subsubsection{Marker Harmonziation}

The functional regions of a QR code, especially the finder and alignment patterns, are crucial for the decoder to locate the QR code. Therefore, these patterns on $I^s$ are strengthened to generate $\widehat{I^s}$. Specifically, for pixel $\mathbf{p}\in\theta_k$ where $k\in \Delta_m$, we have:
\begin{gather}
    \widehat{I^s}(\mathbf{p}) = 
    \begin{cases}
        I^s(\mathbf{p}) - \min(I^s(\mathbf{p})-\tau (1+\lambda), 0), & \text{if } \mathbf{E}_k=1,\\
        I^s(\mathbf{p}) - \max(I^s(\mathbf{p})-\tau (1-\lambda), 0), & \text{if } \mathbf{E}_k=0,
    \end{cases}
\end{gather}
where $\lambda\in (0,1)$ is a hyper-parameter, typically set to 0.8 by default. This threshold-based enhancement helps ensure that the functional regions of the output QR image are easily located.

\subsubsection{Spatially Dynamic Loss Function}
Inspired by~\cite{wu2024text2qr}, we use gradient descent to update the latent code of $\widehat{I^s}$ to optimize certain loss function. However, instead of using a fixed loss function with constant coefficients, we propose to leverage a spatially dynamic loss function. 

Given a pretrained VQ-VAE~\cite{van2017neural} with the encoder $\mathcal{E}$ and the decoder $\mathcal{D}$, the optimization is given by:
\begin{equation}
    \hat{z} = \operatorname*{argmin}_{z} \mathcal{L}(\mathcal{D}(z), I_b, \widehat{I^s}),
\end{equation}
where $z\in\mathbb{R}^{\frac{H}{8}\times\frac{W}{8}\times 4}$ is the latent code. The loss function $\mathcal{L}$ consists of an aesthetic content loss $\mathcal{L}_{a}$ and a spatially dynamic code loss $\mathcal{L}_c$:
\begin{equation} 
\hat{z} = \operatorname*{argmin}_{z}\{ \mathcal{L}_{c}(\mathcal{D}(z), I_b)  + \mathcal{L}_{a}(\mathcal{D}(z), I^s)\}.
\end{equation}
We initialize $z$ to $\mathcal{E}(\widehat{I^s})$, and use Adam~\cite{kingma2014adam} as the optimizer with a learning rate of 0.002 to iteratively update $z$ until convergence. Finally, the output $I^o=\mathcal{D}(\hat{z})$ achieves robust scannability and high visual quality.

\paragraph{Adaptive Code Loss.} 
A simulated decoder~\cite{su2021artcoder} using a 2D Gaussian kernel can extract module-wise information consistent with common QR decoders. The variance $\sigma$ of the Gaussian kernel is a key factor in balancing visual quality and scanning robustness. However, in our scenario, we want the facial region to be smooth and the background region to be lossless. Therefore, we propose a spatially dynamic code loss. Let  $Z=\mathcal{D}(z)$, the loss of $j$-th module is calculated by: 
\begin{equation}
s_j = w(j) \times \text{avg}\{ [Z(\theta_j)-I_b(\theta_j)] \odot G(j)\},
\end{equation}
where $\odot$ denotes the Hadamard product. $G(j)\in \mathbb{R}^{a\times a}$ is a weighting kernel, and $w(j)$ is a weighting factor defined by:
\begin{gather}
    G(j) = 
    \begin{cases}
        G_{\sigma_f}, & \text{if } j \in \Delta_f,\\
        G_{\sigma_b}, & \text{otherwise} 
    \end{cases}; \text{ } w(j) = 
    \begin{cases}
        w_f, & \text{if } j \in \Delta_f,\\
        w_b, & \text{otherwise}, 
    \end{cases}
\end{gather}
where $G_{\sigma}$ is a 2D Gaussian kernel with variance $\sigma$. The specific settings for the hyper-parameters $w_f$, $w_b$, $\sigma_f$, and $\sigma_b$ can be found in the experiments section.
Finally, the adaptive code loss is computed by:
\begin{equation}
    \mathcal{L}_c(Z, I_b)  = \sum_{j=1}^{n^2} w(j) \times \text{avg}\{ [Z(\theta_j)-I_b(\theta_j)] \odot G(j)\}.
\end{equation}
Gaussian distribution with bigger $\sigma$ is flatter, which helps equalize the color within the module when updating the latent code. Although this makes modules easier to decode after iterations, bigger $\sigma$ might create unnatural shadow in the face region. On the other hand, Gaussian distribution with smaller $\sigma$ effectively regulates only the central region of a module, making the modules remain unscannable even after updates.

This problem is addressed by utilizing adaptive loss for different regions, \textit{i.e.}, applying smaller weight $w_f$ and $\sigma_f$ in the face region to prevent distortion on face, and relatively larger $w_f$ and $\sigma_f$ in remaining region to maintain balance between scannability and aesthetic quality.

\paragraph{Aesthetic Content Loss.} To ensure the retention of aesthetic qualities while preserving face ID and enhancing scannability, we use the aesthetic content loss to retain essential visual characteristics. It is quantified by calculating $L^2$-Wasserstein distance~\cite{Wasserstein} (denoted as $D_{W2}$) of feature representations between $Z$ and $\widehat{I^s}$ as follows:
\begin{equation}
    \mathcal{L}_{a}(Z, \widehat{I^s}) = \sum_i D_{W2}(g_i(Z), g_i(\widehat{I^s})),
\end{equation}
where $g_i$ is feature representations from a pre-trained VGG-19~\cite{vgg} network at layer $i$. The aesthetic content loss reflects the global aesthetic quality of the image. By optimizing both code loss and content loss, IDSE module adeptly balances the aesthetic quality, face-preserving, and scannability and creates optimal customized QR code images.
\section{Experiments}\label{sec:exp}

\subsection{Experimental Setup and Configuration}\label{sec:exp4-1}

We implemented our pipeline in Python using the PyTorch framework and conducted experiments on an NVIDIA GeForce 4090 GPU. The scannability of QR images is tested using a 27-inch IPS display monitor with a refresh rate of 144Hz. In our experiments, we set control strengths for the InstantID network~\cite{wang2024instantid} and QR Controller at 0.8 and 1.4, respectively. The parameter $\lambda$ in the marker harmonization process defaults to 0.8. The VAE configuration is consistent with the SD model. The face recognition model AntelopeV2 from InsightFace~\cite{insightface} assists the generation of face mask $M_f$ in IDRE. The VGG-19 architecture, pre-trained on the MS-COCO dataset, facilitates the feature map extraction in IDSE. The Adam optimizer powers the optimization within IDSE, performing 300 iterations at a learning rate of 0.002. 
Default settings for $\sigma_f$, $\sigma_b$, $w_f$, and $w_b$ are 1.5, 3.0, 1.0, and 15.0 respectively. We produce QR code in version 5, each with $37\times 37$ modules. For clarity, we define $e$ as the number of error modules in QR image $I^o$ (excluding finder and alignment pattern areas), and $e_f$ as the number of error modules within the face region. Our dataset for comparative analysis contains 200 uniquely stylized QR images, each $1024\times 1024$ pixels in size, with diverse visual content and artistic styles. To more accurately assess the preservation of face identity, we define the face feature distance $d$ as the cosine similarity between the facial features (extracted using ArcFace~\cite{deng2019arcface}) of the generated QR image $I^o$ and the original face image $f$.

\subsection{Qualitative Comparison}
\paragraph{Aesthetic Quality.}
\newlength{\mfigwidth}
\setlength{\mfigwidth}{0.16\textwidth} 
\begin{table*}[t]
    \vspace{-5pt}
    \caption{Visual comparison of different methods.
    }
    \label{tab:comparison}
    \centering
    \begin{tabularx}{\linewidth}{>{\centering\arraybackslash}X *{5}{>{\centering\arraybackslash}X}}

    \toprule
    Input 
    & QArt~\cite{qart}
    & {Halftone~\cite{chu2013halftone}} 
    & ArtCoder~\cite{su2021artcoder} & Text2QR~\cite{wu2024text2qr} & Face2QR \\
    \midrule

    \includegraphics[width=\mfigwidth, height=\mfigwidth]{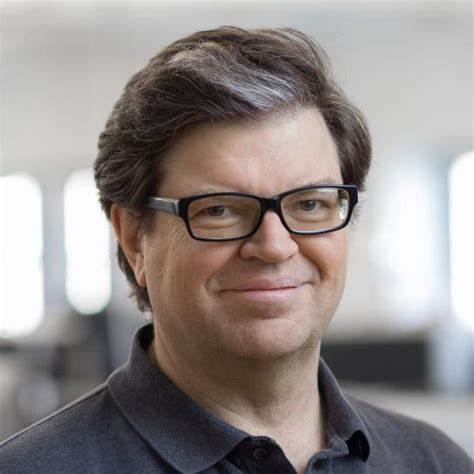} & 
    \includegraphics[width=\mfigwidth, height=\mfigwidth]{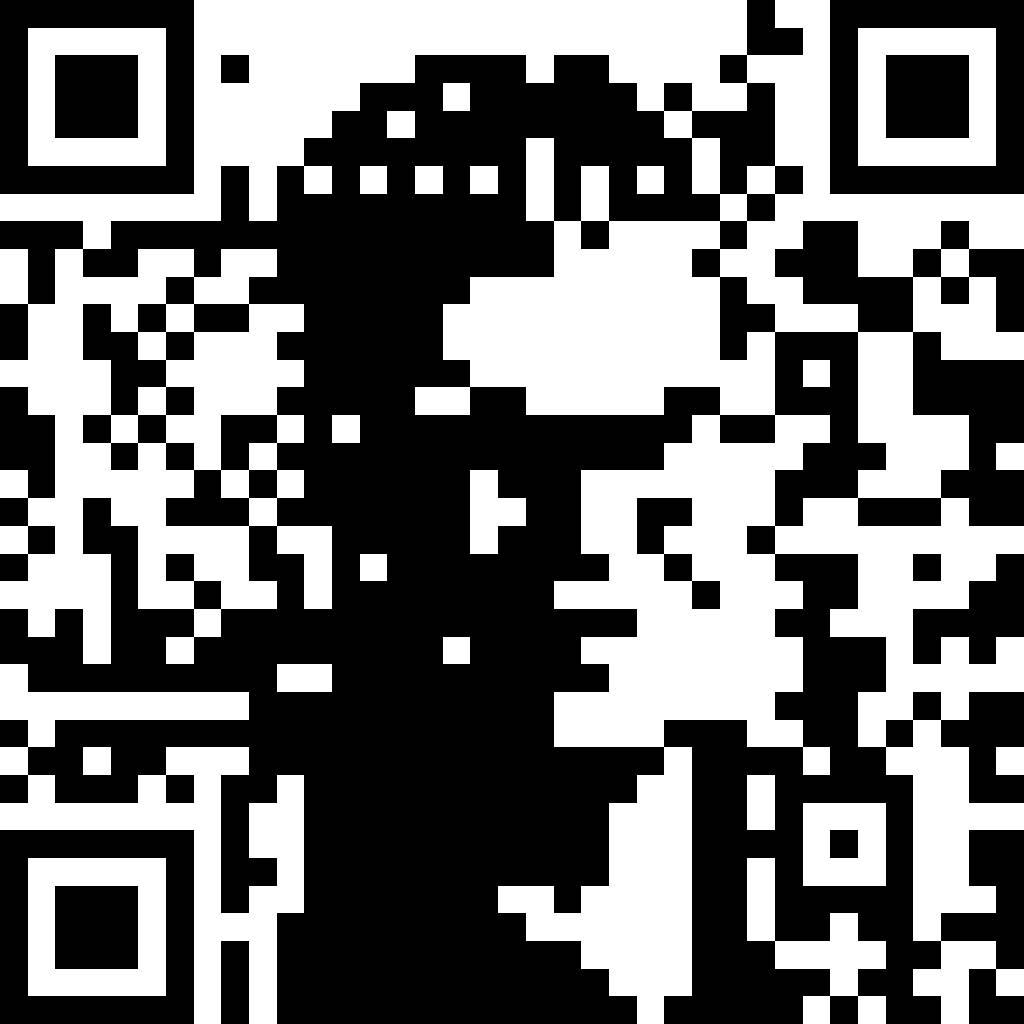} &
    \includegraphics[width=\mfigwidth, height=\mfigwidth]{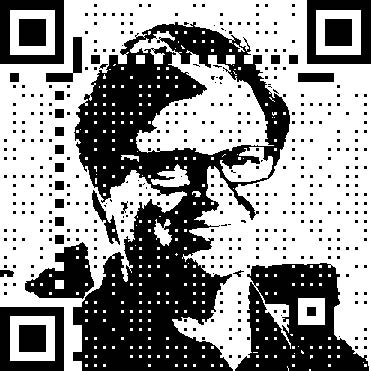} & 
    \includegraphics[width=\mfigwidth, height=\mfigwidth]{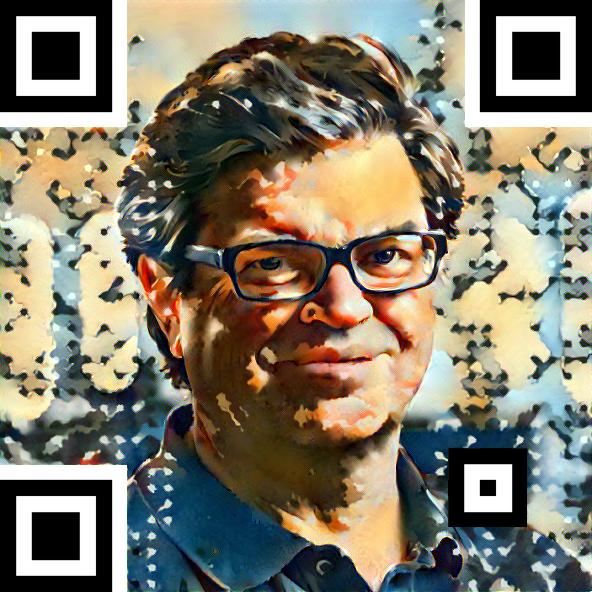} & 
    \includegraphics[width=\mfigwidth, height=\mfigwidth]{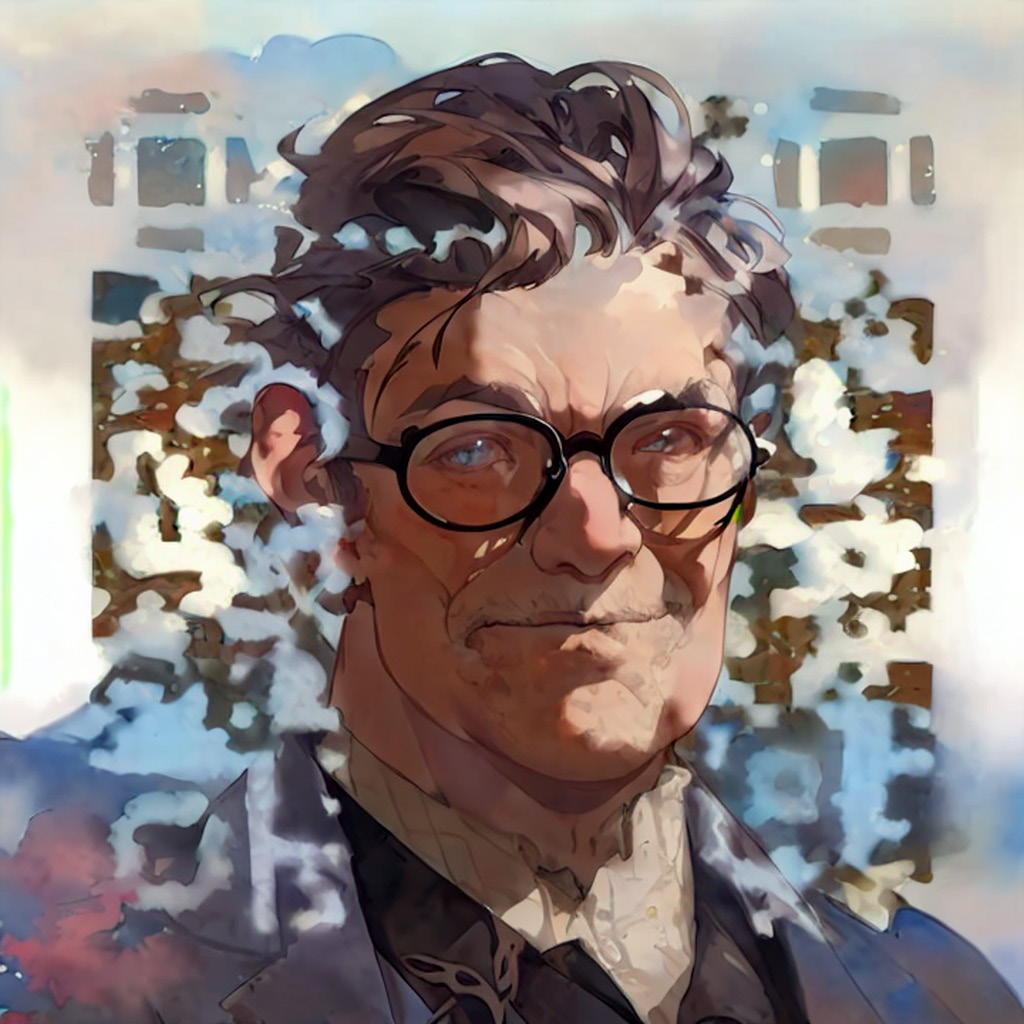} & 
    \includegraphics[width=\mfigwidth, height=\mfigwidth]{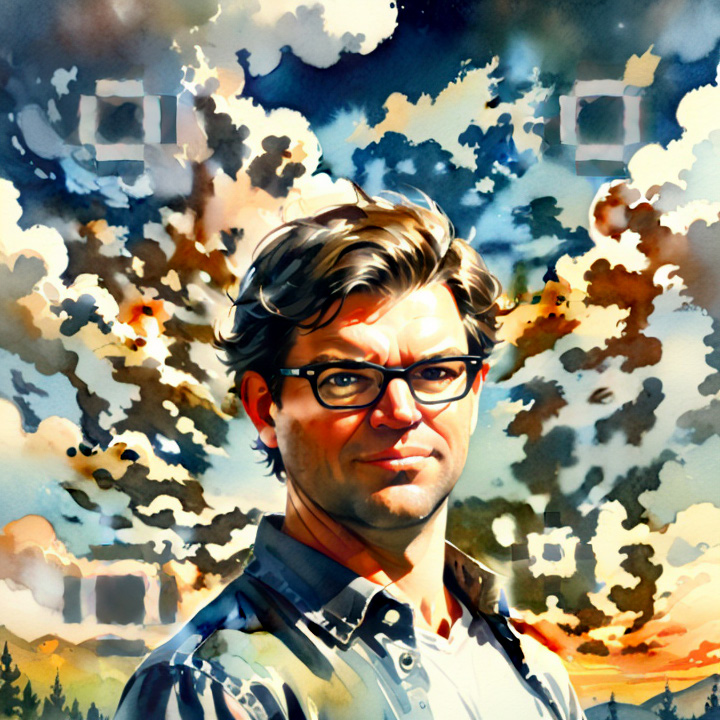} \\

    \includegraphics[width=\mfigwidth, height=\mfigwidth]{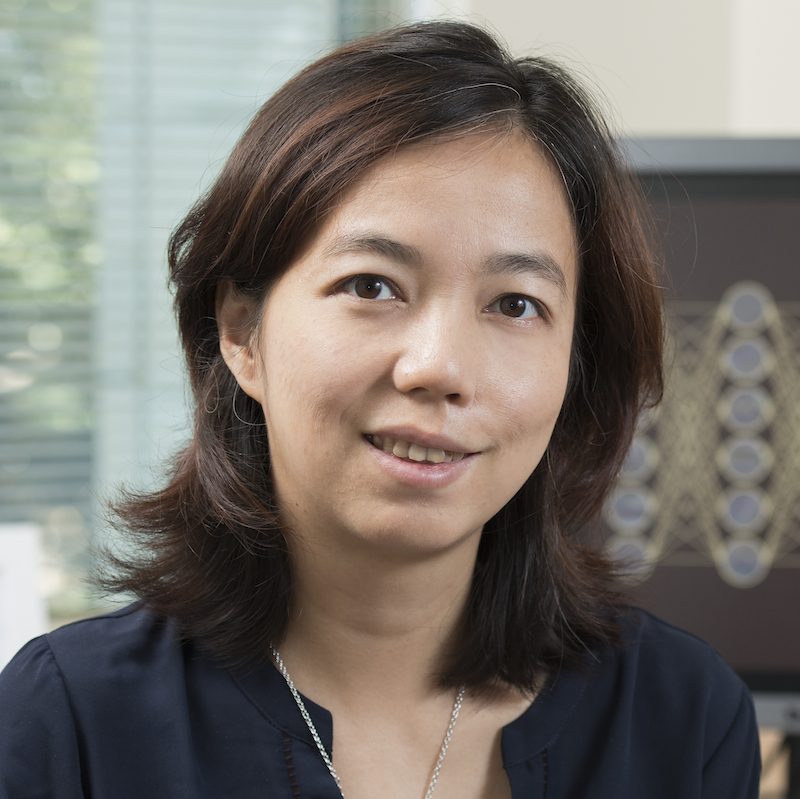} & 
    \includegraphics[width=\mfigwidth, height=\mfigwidth]{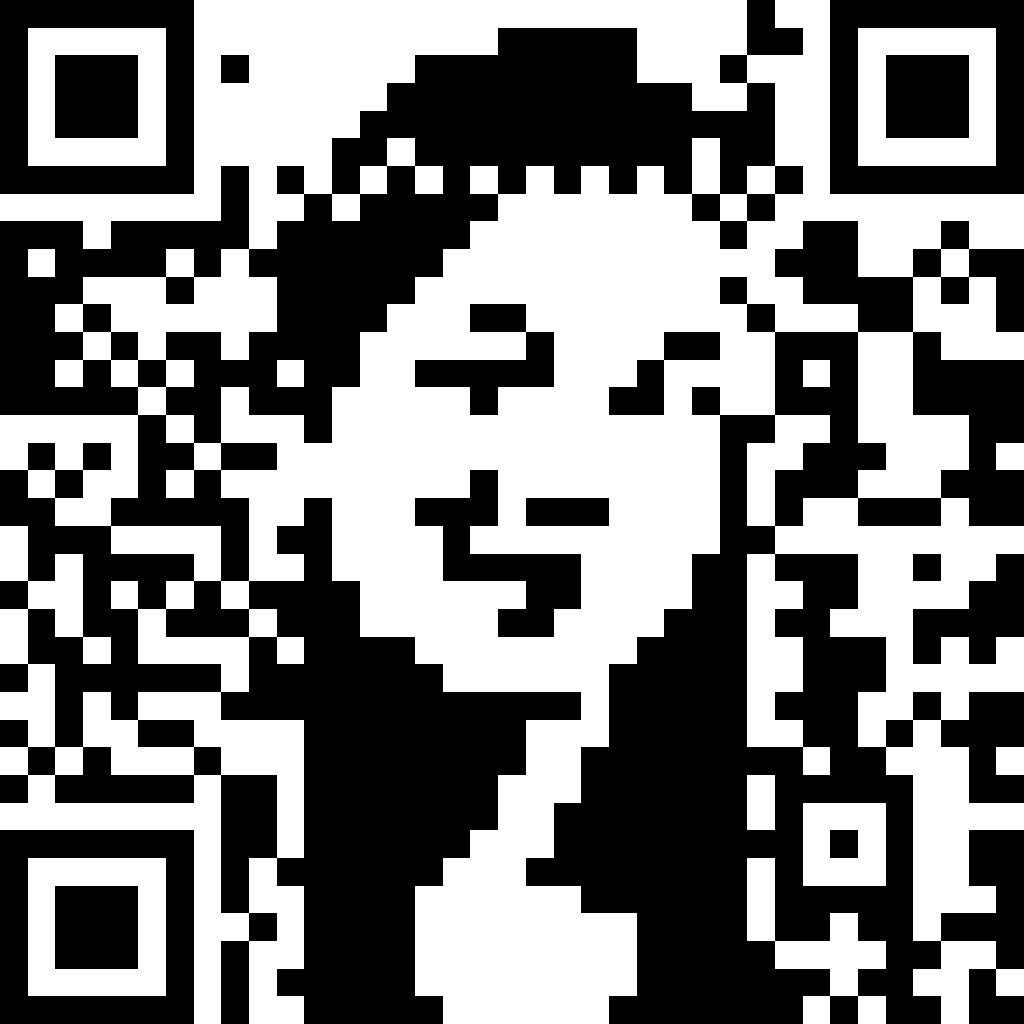} &
    \includegraphics[width=\mfigwidth, height=\mfigwidth]{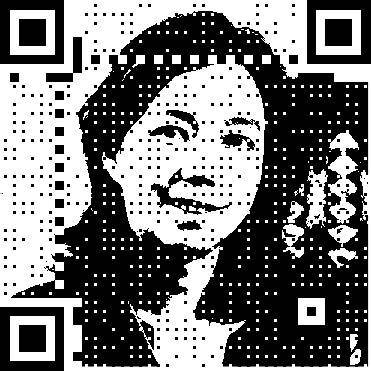} & 
    \includegraphics[width=\mfigwidth, height=\mfigwidth]{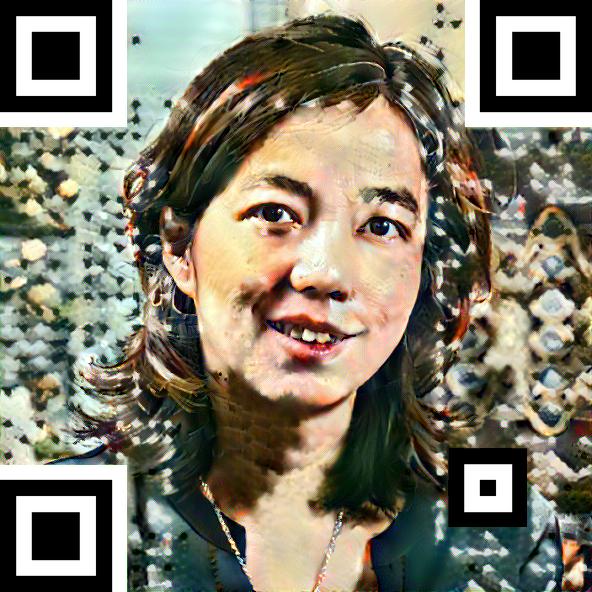} & 
    \includegraphics[width=\mfigwidth, height=\mfigwidth]{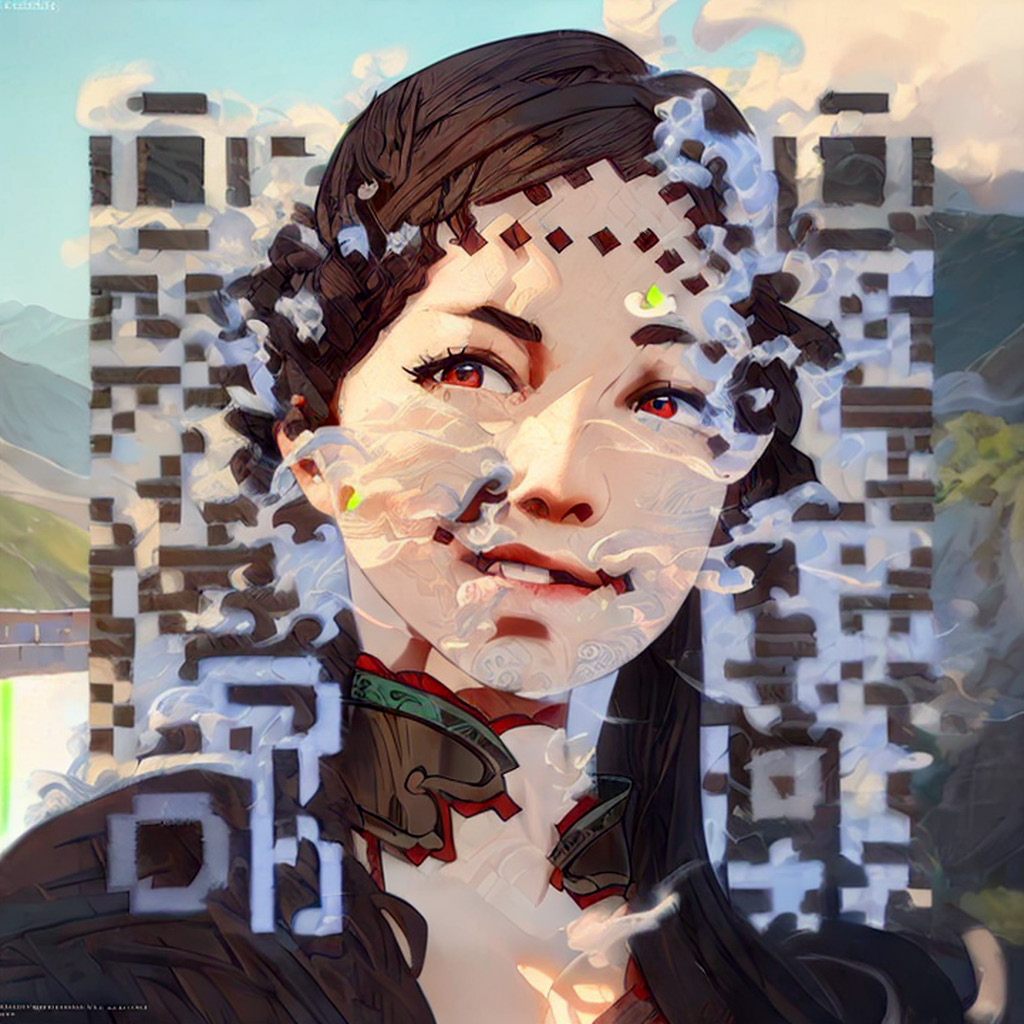} & 
    \includegraphics[width=\mfigwidth, height=\mfigwidth]{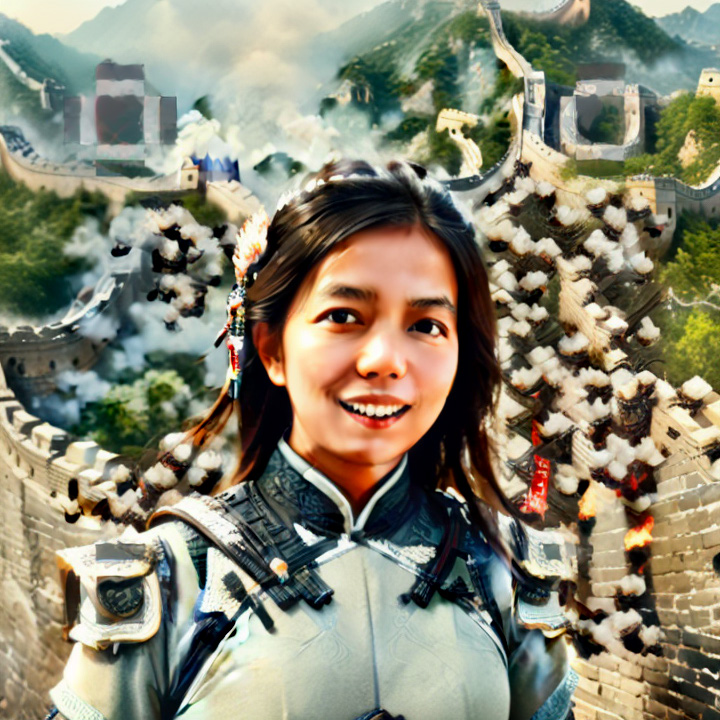} \\

    \includegraphics[width=\mfigwidth, height=\mfigwidth]{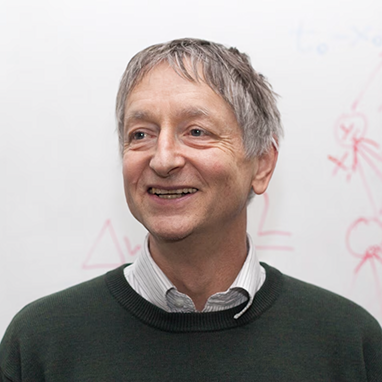} & 
    \includegraphics[width=\mfigwidth, height=\mfigwidth]{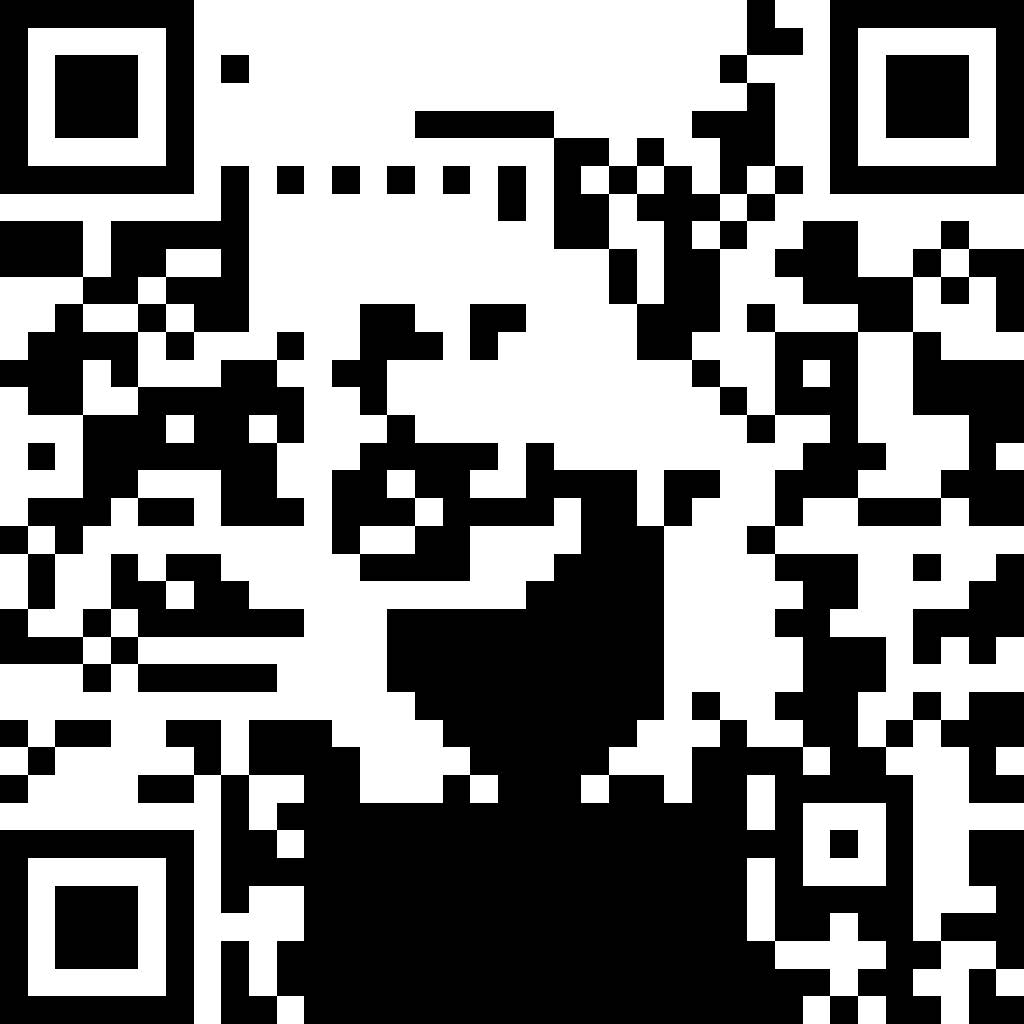} &
    \includegraphics[width=\mfigwidth, height=\mfigwidth]{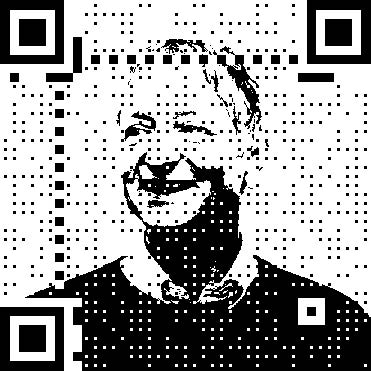} & 
    \includegraphics[width=\mfigwidth, height=\mfigwidth]{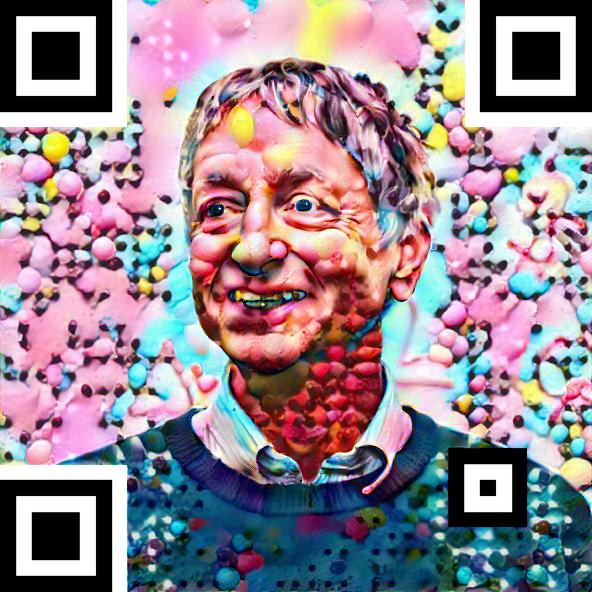} & 
    \includegraphics[width=\mfigwidth, height=\mfigwidth]{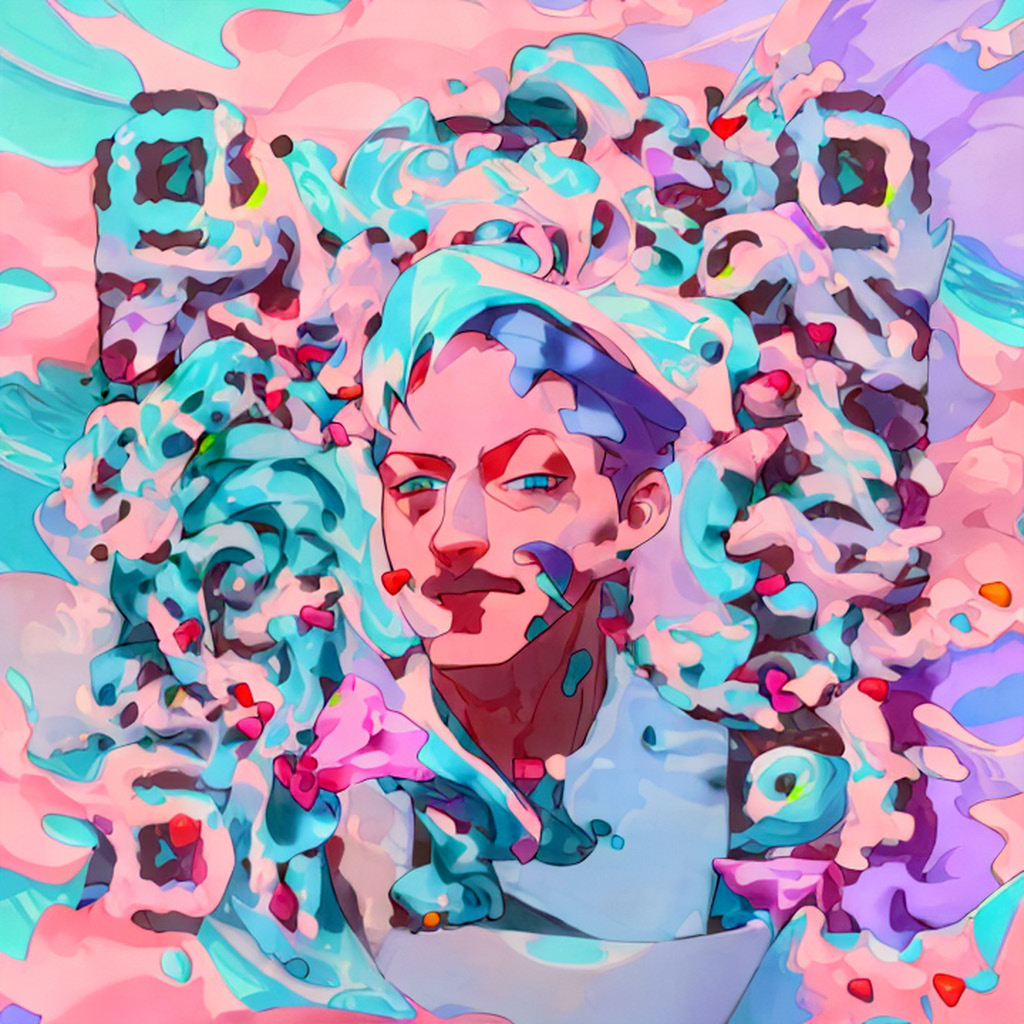} & 
    \includegraphics[width=\mfigwidth, height=\mfigwidth]{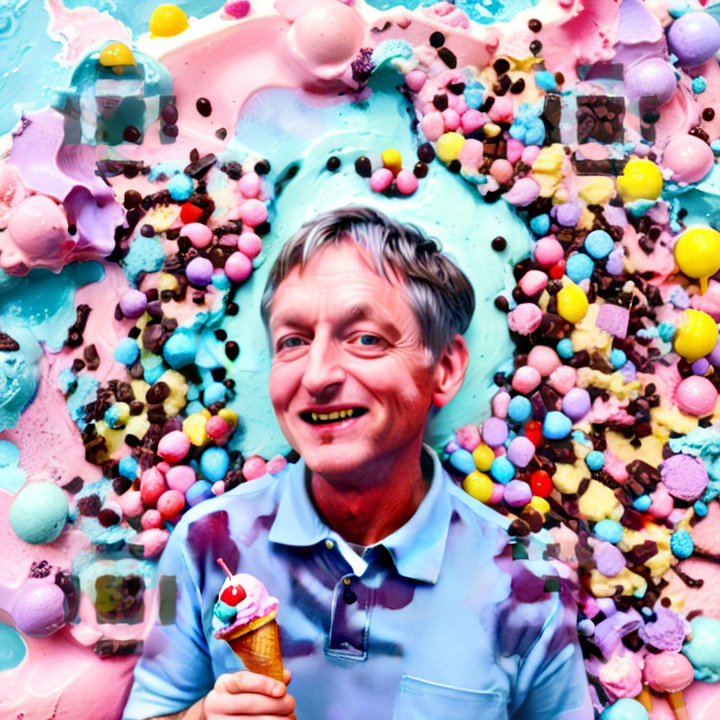} \\

    \bottomrule
    \end{tabularx}
    \vspace{-5pt}
\end{table*}

In our comparative study, we evaluate our approach against several state-of-the-art aesthetic QR code generation techniques, including QArt~\cite{qart}, Halftone QR code~\cite{chu2013halftone}, ArtCoder~\cite{su2021artcoder} and Text2QR~\cite{wu2024text2qr}, as detailed in Table~\ref{tab:comparison}. QArt, Halftone QR and Text2QR take the original face image $f$ as the primary input, except that Text2QR takes in additional prompt input $c$. As ArtCoder is based on neural-style transfer technique, we employ $f$ and $I^g$ to serve as the content reference and the style reference respectively. The results show that Artcoder tends to render the texture of style image to face region, causing unwanted distortion on the face. Text2QR, on the other hand, cannot preserve face ID due to lack of specific control mechanisms for the face region.
In contrast, our QR codes are adept at harmoniously integrating face ID, background and QR pattern, thereby achieving superior visual quality as well as scannability.

\paragraph{Identity Preservation.}
\begin{table}[ht]
    \centering
    \caption{Visual comparison of face ID preservation in face image $f$ and generated QR image $I^o$. $I^o_1$ and $I^o_2$ are generated from $f_1$, and $I^o_3$ and $I^o_4$ are generated from $f_2$. Face feature distance $d$ is measured between pairs of $I^o$ and $f$.}
    \vspace{5pt}
    \begin{tabular}{c}
        \toprule
        \begin{picture}(0,0)
            \put(-165,2){\makebox(0,0)[c]{\footnotesize $f_1$}}
            \put(-100,2){\makebox(0,0)[c]{\footnotesize $I^o_1$}}
            \put(-32,2){\makebox(0,0)[c]{\footnotesize $I^o_2$}}
            \put(35,2){\makebox(0,0)[c]{\footnotesize $f_2$}}
            \put(100,2){\makebox(0,0)[c]{\footnotesize $I^o_3$}}
            \put(165,2){\makebox(0,0)[c]{\footnotesize $I^o_4$}}
        \end{picture} \\ \midrule
        \includegraphics[width=0.98\linewidth]{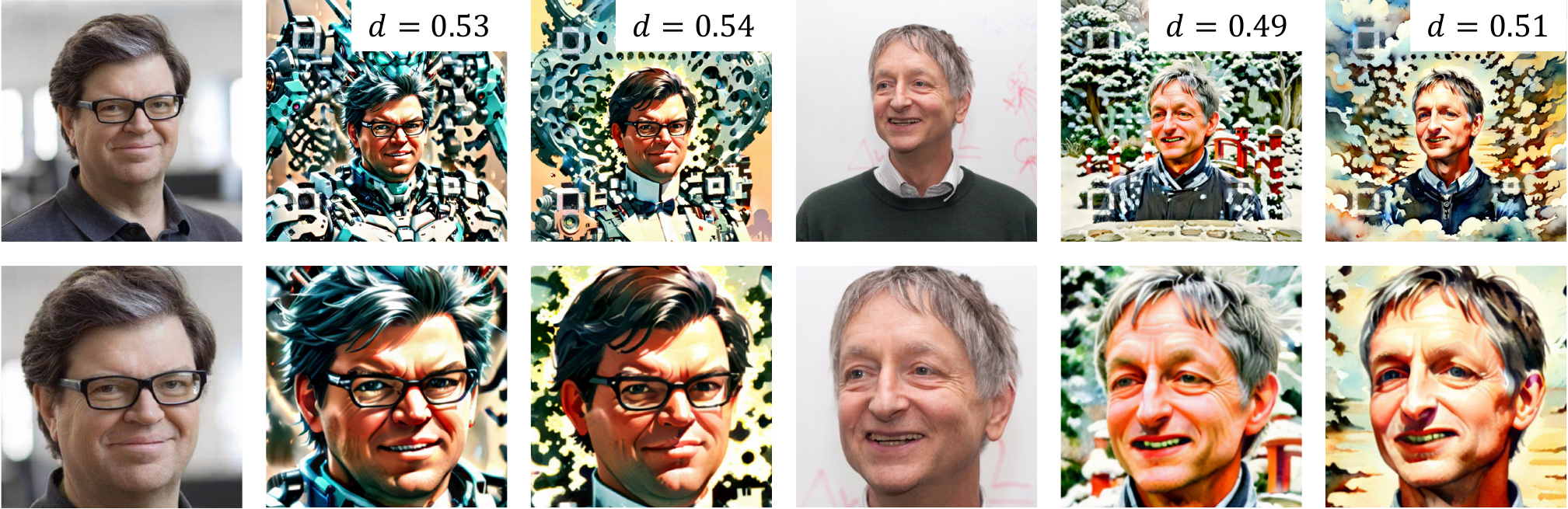}
        \\ \bottomrule
    \end{tabular}
    \vspace{-2mm}
    \label{fig:face}
\end{table}

The comparison between original face image $f$ and the generated image $I^o$ is shown in Table~\ref{fig:face}. The face ID is well preserved in the final generated QR image $I^o$, with minimal change in haircut or facial expression, which can be further customized by users by adding prompt. The facial region is consistent with the background in style, and the QR pattern is blended seamlessly into the picture. We also compare the generated image $I^o$ with output of the baseline pipeline InstantID~\cite{wang2024instantid} in Table~\ref{tab:com_instantid}, which shows that our pipeline achieves a similar level of identity preservation as the baseline. The outcomes displayed in Table~\ref{tab:angle} demonstrate that Face2QR consistently generates high-quality images across various poses.

\subsection{Quantitative Comparsion}
\paragraph{Scanning Robustness.} 

In this study, we assess the scanning robustness of our QR images using different scanning applications. We first generate a batch of 20 aesthetically pleasing QR codes, each with a dimension of 1,024 $\times$ 1,024 pixels. These QR images are then displayed on a high-definition monitor in three standard sizes: 3cm $\times$ 3cm, 5cm $\times$ 5cm, and 7cm $\times$ 7cm. During our controlled test, smartphones are held at a fixed distance of 25cm from the display, and each code is scanned for 3 seconds from different angles. Over a total of 50 trials, the percentage of successful scans is recorded in Table~\ref{table:scannability}. The results reveal an average success rate exceeding 94\%, showcasing high reliability of the generated QR images in diverse practical settings. It is also noted that QR images that fail the test in 3s can eventually be scanned if given more time. The scanning success rate is similar to that of Text2QR~\cite{wu2024text2qr}, as presented in our comparative analysis.

\begin{table*}[ht]
    \centering
    \newlength{\rebl}
    \setlength{\rebl}{0.156\textwidth}
    \newlength{\angles}
    \setlength{\angles}{0.156\textwidth}
    \begin{minipage}[b]{.49\linewidth}
        \caption{Visual comparison of identity preservation with InstantID~\cite{wang2024instantid}.}
        \label{tab:com_instantid}
        \centering
        \begin{tabularx}{\linewidth}{>{\centering\arraybackslash}X *{3}{>{\centering\arraybackslash}X}}
    
        \toprule
        Input 
        &  InstantID~\cite{wang2024instantid} & Face2QR \\
        \midrule

        \includegraphics[width=\rebl, height=\rebl]{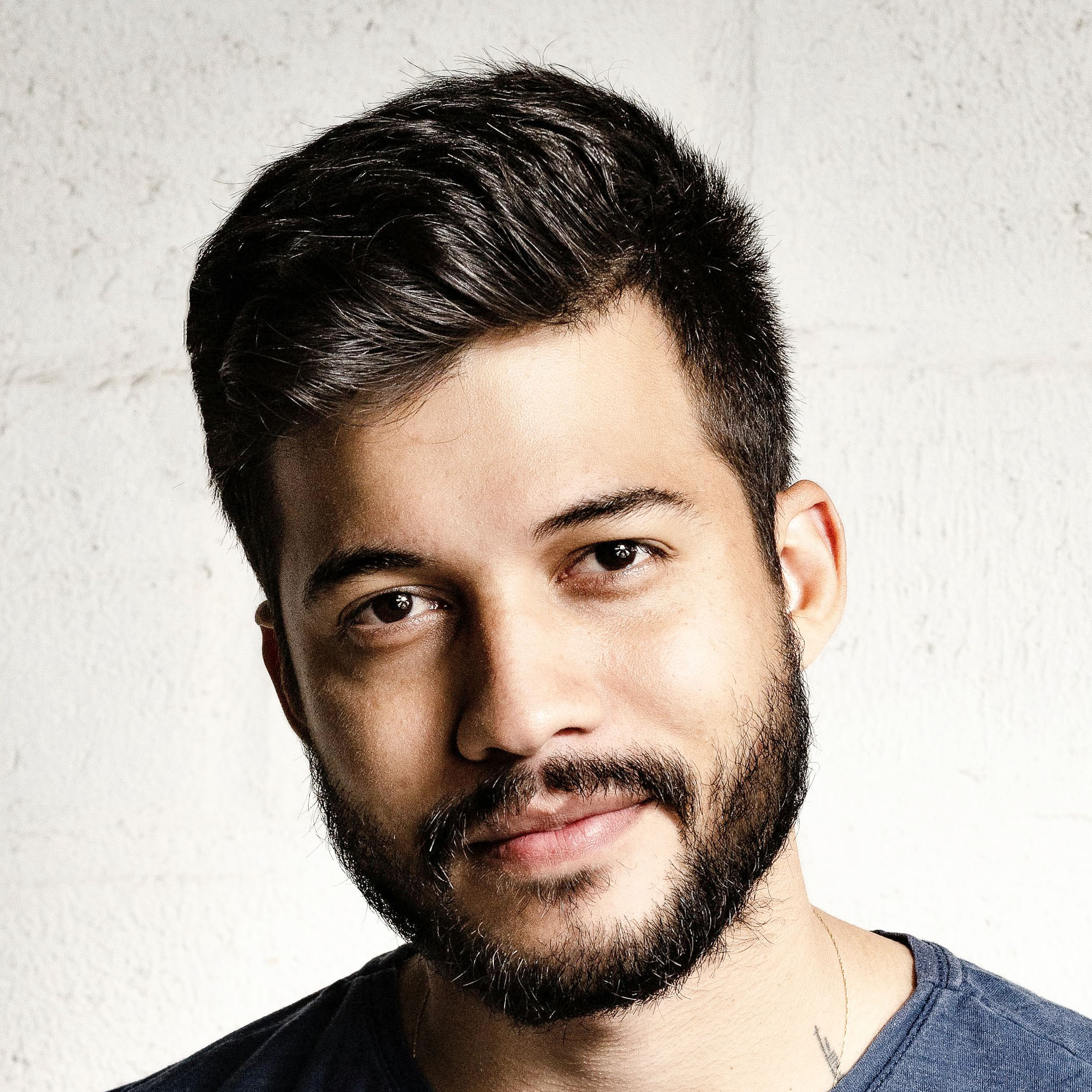} & 
        \includegraphics[width=\rebl, height=\rebl]{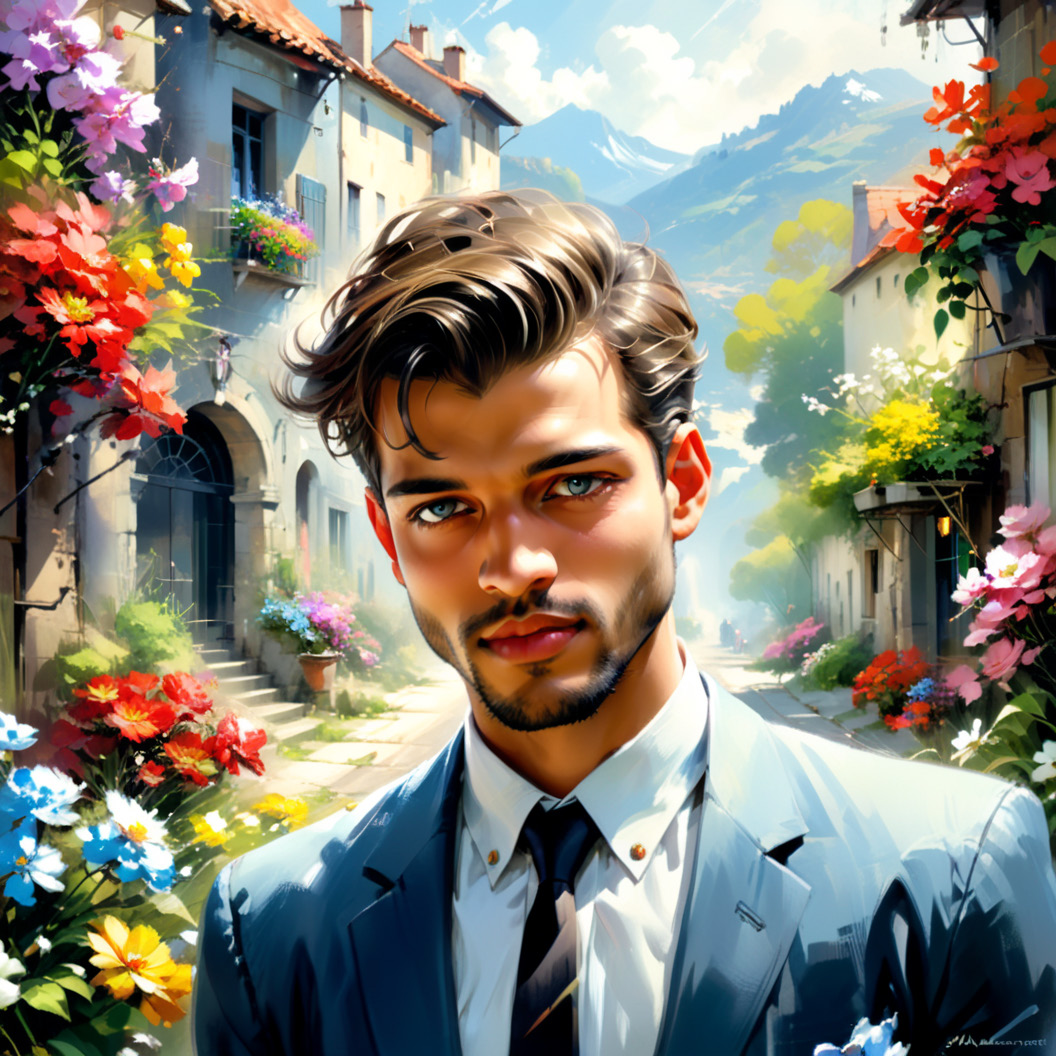} & 
        \includegraphics[width=\rebl, height=\rebl]{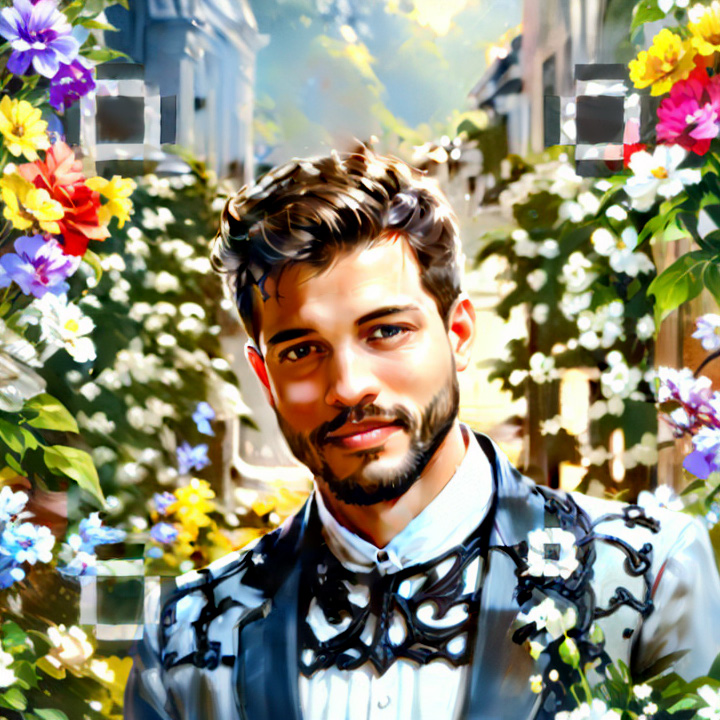} \\
    
        \includegraphics[width=\rebl, height=\rebl]{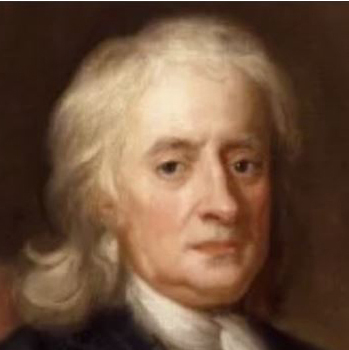} & 
        \includegraphics[width=\rebl, height=\rebl]{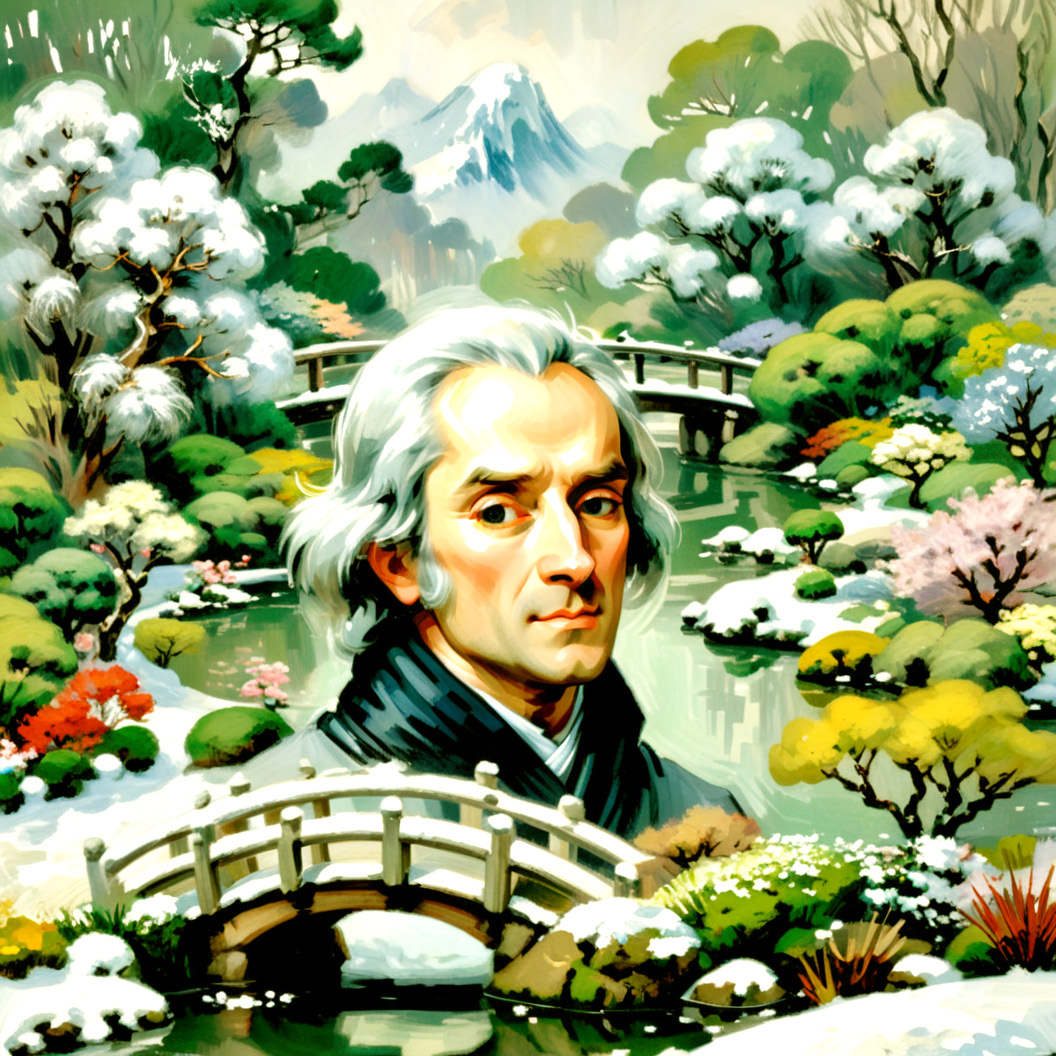} & 
        \includegraphics[width=\rebl, height=\rebl]{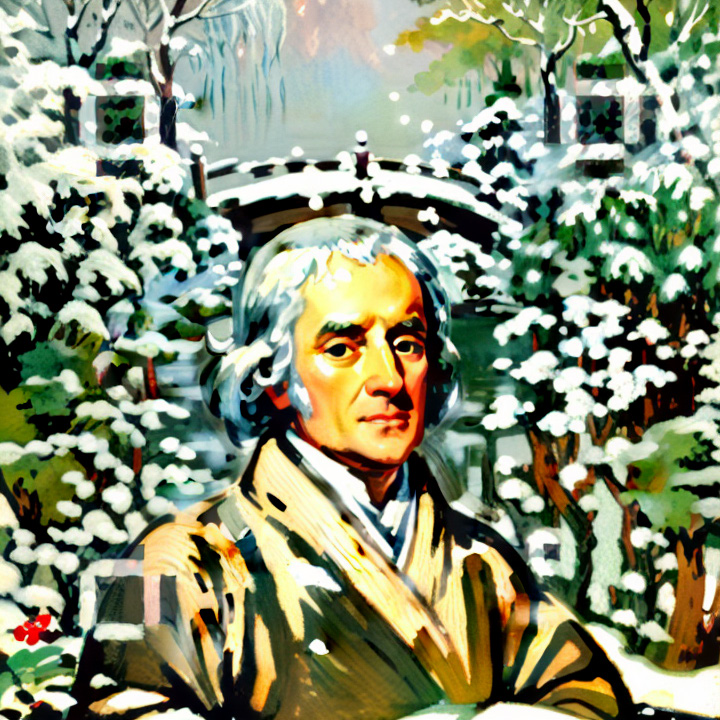} \\
    
        \bottomrule
        \end{tabularx}
    \end{minipage}\hfill
    \begin{minipage}[b]{.49\linewidth}
        
        \caption{Generated QR images (second row) using face images (first row) with different poses.
        }
        \label{tab:angle}
        \centering
        \begin{tabularx}{\linewidth}{>{\centering\arraybackslash}X *{3}{>{\centering\arraybackslash}X}}
    
        \toprule
        $\sim90^{\circ}$ 
        &  $\sim45^{\circ}$ & $0^{\circ}$ \\
        \midrule
    
        \includegraphics[width=\angles, height=\angles]{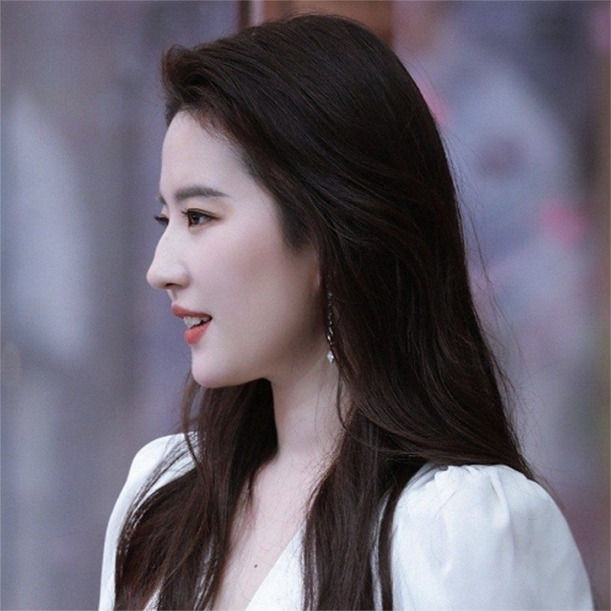} & 
        \includegraphics[width=\angles, height=\angles]{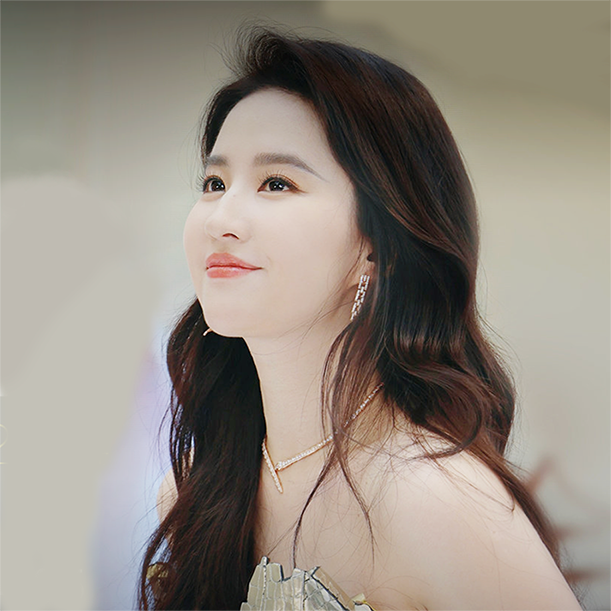} & 
        \includegraphics[width=\angles, height=\angles]{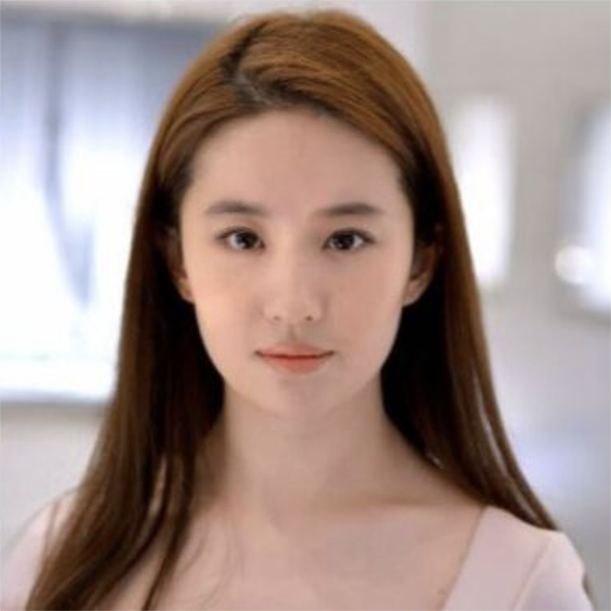} \\
    
        \includegraphics[width=\angles, height=\angles]{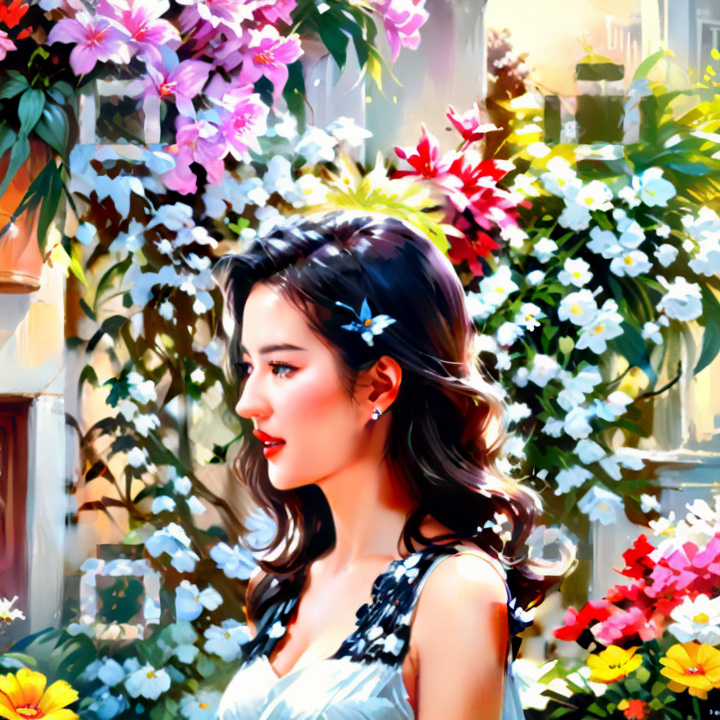} & 
        \includegraphics[width=\angles, height=\angles]{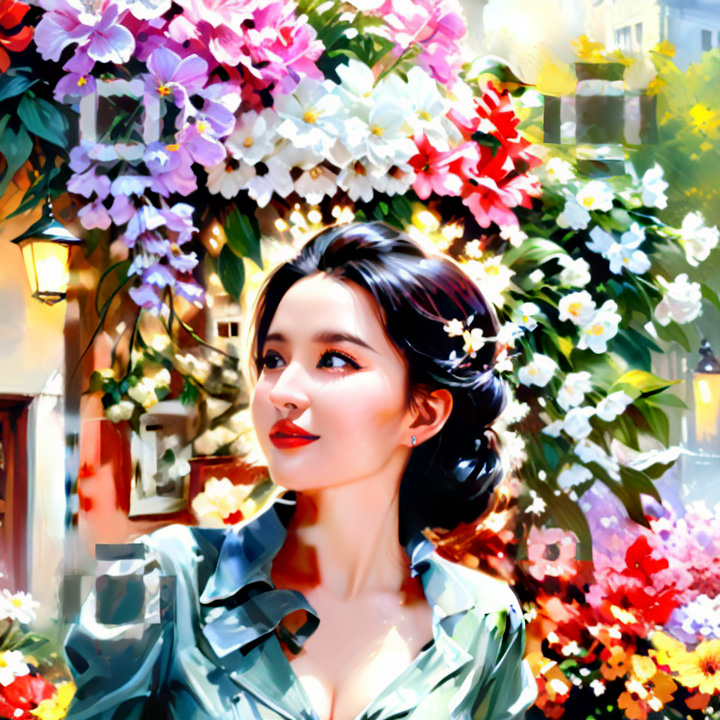} & 
        \includegraphics[width=\angles, height=\angles]{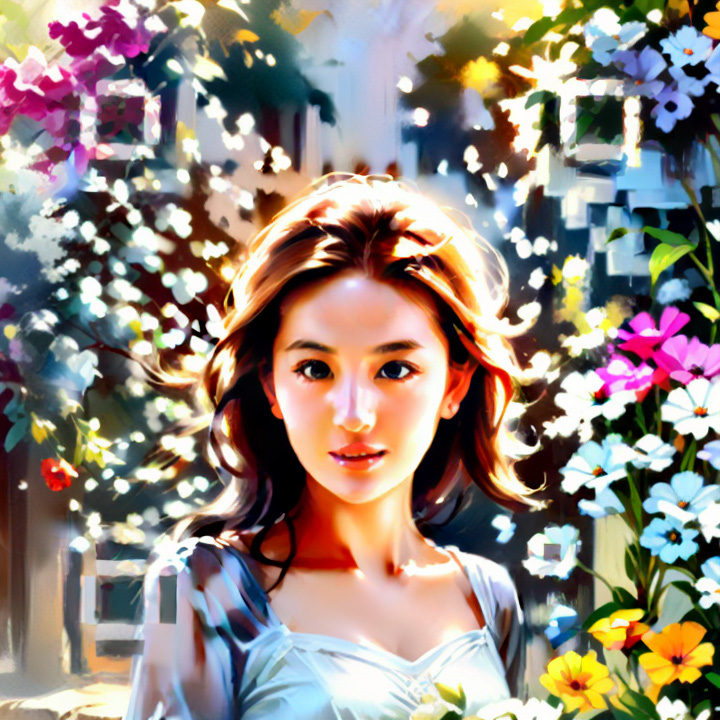} \\
    
        \bottomrule
        \end{tabularx}
    \end{minipage}
\end{table*}

\paragraph{Subjective Study.} Figure~\ref{fig:users} presents a user study consisting of 30 participants to evaluate 150 QR images (50 for each methods) generated by different methods (the approval from Institutional Review Board is obtained). Participants are asked to choose the better one from a pair of pictures in the aspect of face ID preservation and aesthetic quality. Each pair of QR images are generated by different methods using the same face image as input. The percentages represent how many times users prefer the results of a method over the other. Our results are preferred by most users.

\begin{table*}[ht]
    \centering
    \begin{minipage}[t]{.45\linewidth}
        \centering
        \caption{Scannability success rates of QR codes across various decoders at different sizes and angles.}
        \label{table:scannability}
        \begin{tabularx}{\linewidth}{c *{6}{>{\centering\arraybackslash}X}}
        \toprule
        \multirow{3}{*}{\textbf{Decoders}} & \multicolumn{6}{c}{\textbf{Success Rate (\%)}}  \\ 
        \cmidrule(lr){2-7}
        & \multicolumn{2}{c}{(3cm)$^2$} & \multicolumn{2}{c}{(5cm)$^2$} & \multicolumn{2}{c}{(7cm)$^2$}  \\ 
        
        \cmidrule(lr){2-3} \cmidrule(lr){4-5} \cmidrule(lr){6-7}
        & {$45^{\circ}$} & {$90^{\circ}$} & {$45^{\circ}$} & {$90^{\circ}$}   & {$45^{\circ}$} & {$90^{\circ}$}   \\ 
        
        \midrule
        & \multicolumn{6}{c}{Face2QR (ours)}  \\
        \cmidrule(lr){2-7}
        Scanner & 100 & \underline{94} & 100 & 100 & 100 & 100 \\ 
        TikTok &100 & 100 & 100 & 100 & 100 & 100 \\ 
        WeChat &96 & 100 & 100 & 100 & \underline{94} & 100  \\ 
        \midrule
        & \multicolumn{6}{c}{Text2QR~\cite{wu2024text2qr}}  \\
        \cmidrule(lr){2-7}
        Scanner & 96 & 96 & 100 & 100 & 100 & 100 \\ 
        TikTok &100 & 100 & 100 & 100 & 100 & 100 \\ 
        WeChat &100 & \underline{94} & 100 & 100 & \underline{94} & 100  \\ 
        \bottomrule
        \end{tabularx}
    \end{minipage}\hfill
    \begin{minipage}[t]{.53\linewidth}
        \centering

        \captionof{table}{Comparison of average face feature distance $d$ and average Aesbench scores $B_a$. [Key: \textbf{Best}]}
        \label{tab:stat}
        \begin{tabularx}{\linewidth}{c >{\centering\arraybackslash}X *{3}{>{\centering\arraybackslash}X}}
            \toprule
            & ArtCoder~\cite{su2021artcoder} & Text2QR~\cite{wu2024text2qr} & Face2QR \\ \cmidrule{2-4}
            $d$  & 0.50 & 0.43 & $\mathbf{0.51}$ \\ \midrule
            $B_a$ & 62.0 & 87.5 & $\mathbf{90.1}$ \\ \bottomrule
        \end{tabularx}
        \vspace{10pt}
        
        \includegraphics[width=0.92\linewidth]{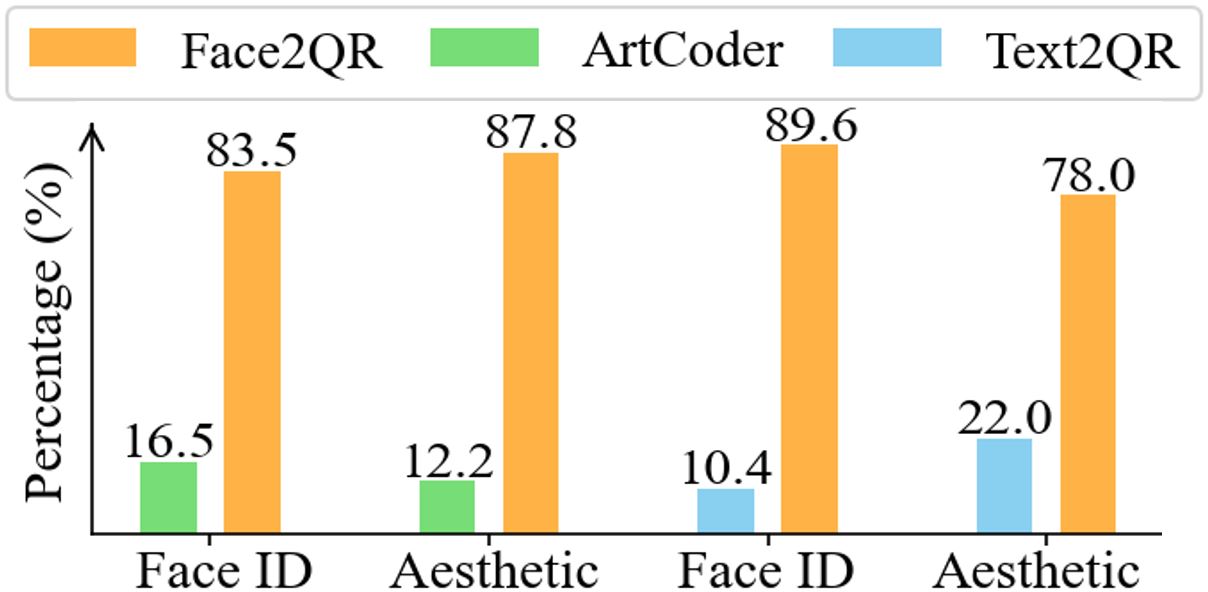}

        \captionof{figure}{User study of different methods.}
        \label{fig:users}
    \end{minipage}
\end{table*}

\paragraph{Objective Study.} Table~\ref{tab:stat} shows the statistical performance measured by taking the average of 100 samples. We use the feature distance $d$, varying from -1 to 1, as a quantifiable measure for the preservation of face ID. The higher the distance $d$, the generated face is more consistent with the original face image. We also use AesBench tool~\cite{aesbench}, which assigns aesthetic scores ranging from $0$ to $100$ (with higher scores denoting better aesthetics), to objectively evaluate the aesthetic quality of generated pictures. The results indicate that our approach exceeds competing methods in all evaluated metrics, confirming its capability to produce QR iamges with faithfully preserved face ID and high aesthetic quality.

\subsection{Ablation Study}
\begin{table*}[ht]
    \centering
    \newlength{\aww}
    \setlength{\aww}{0.13\linewidth}
    \begin{minipage}[b]{.45\linewidth}
    \caption{IDRE Ablation Study: Compared with result obtained by completing the entire stage 2 (rightmost column), skipping stage 2 (leftmost column) or conducting stage 2 without IDRE (middle column) will result in more error modules $e_f$ in the face region. The distribution of error modules, highlighted as bright areas, is depicted in the second row.}
        \label{tab:IDRE}
        \centering
        
        \begin{tabularx}{\linewidth}{>{\centering\arraybackslash}X >{\centering\arraybackslash}X *{3}{>{\centering\arraybackslash}X}}
    
        \toprule
        $I^g$ & w/o. IDRE & with IDRE \\ \midrule
    
        \includegraphics[width=\aww, height=\aww]{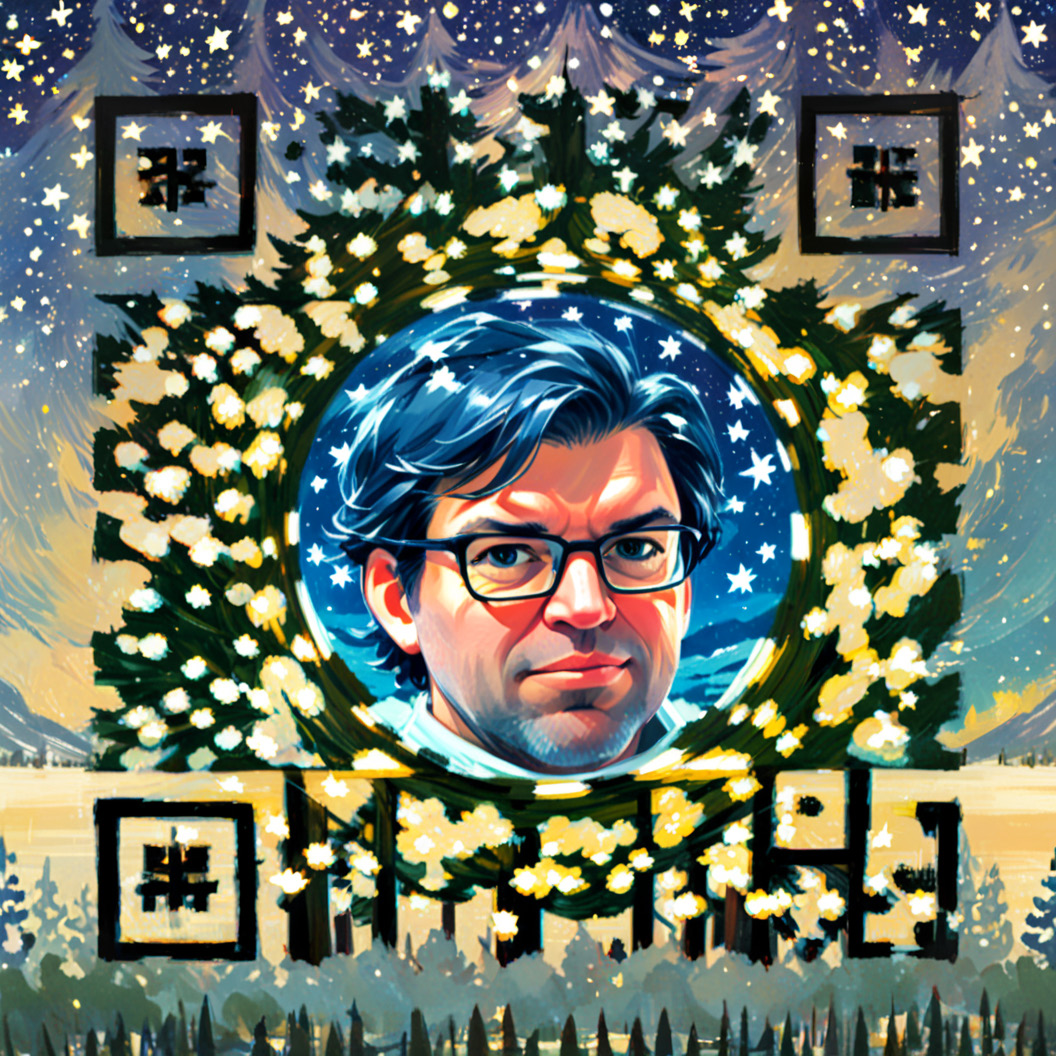} & 
        \includegraphics[width=\aww, height=\aww]{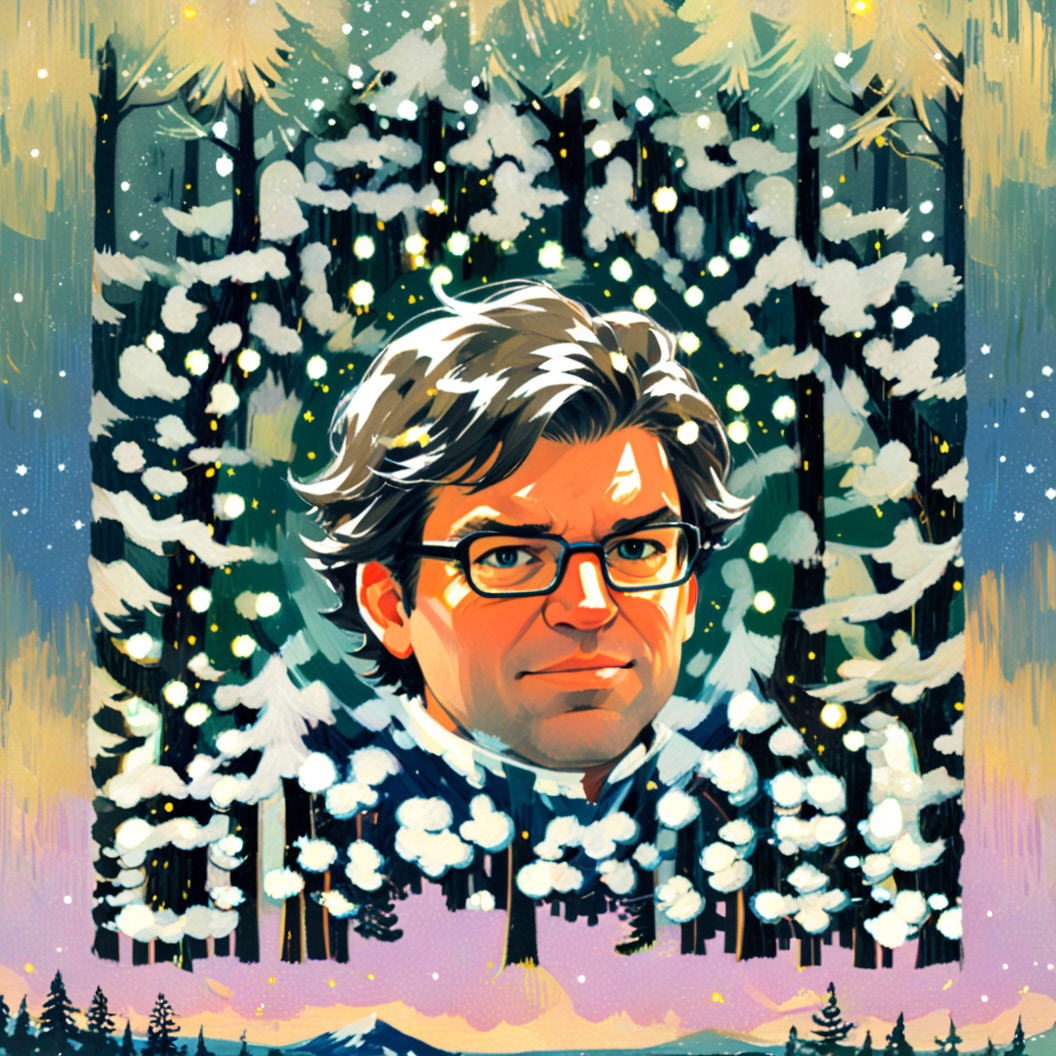} & 
        \includegraphics[width=\aww, height=\aww]{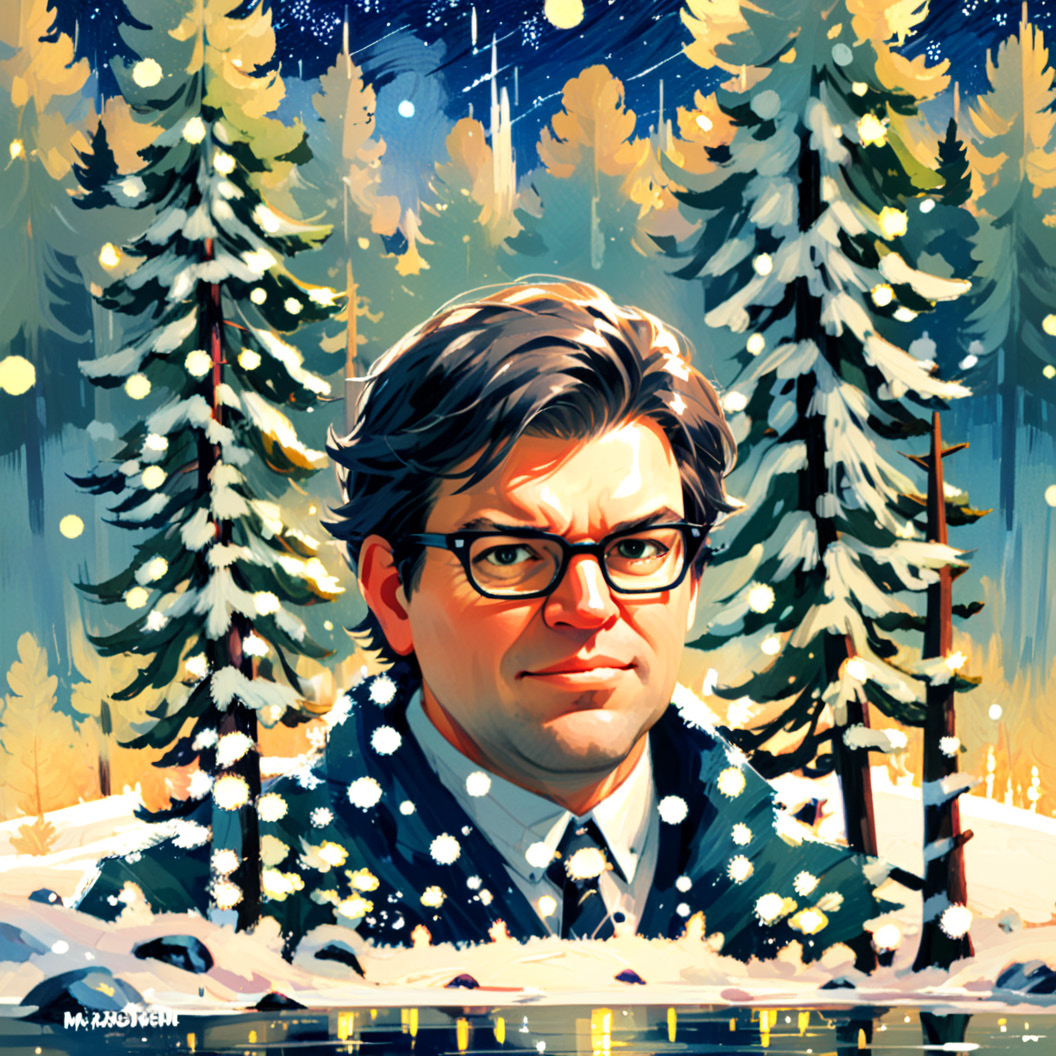} \\

        \includegraphics[width=\aww, height=\aww]{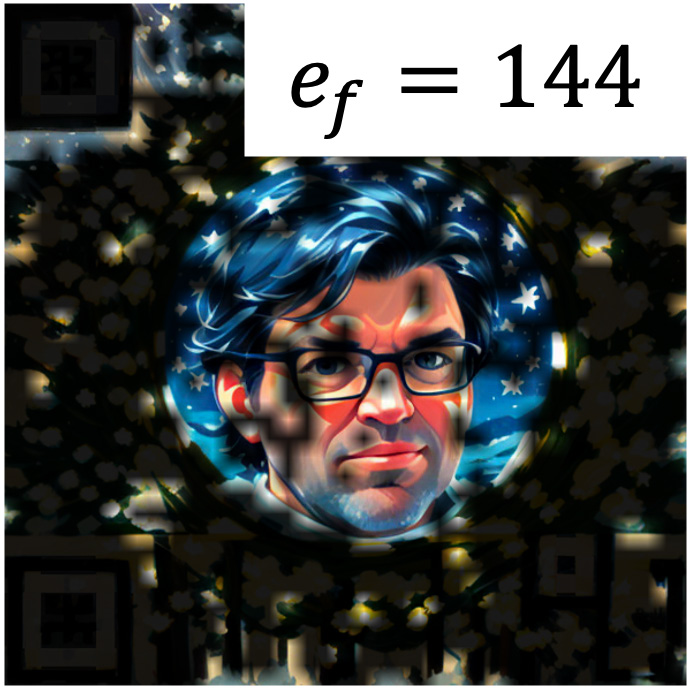} & 
        \includegraphics[width=\aww, height=\aww]{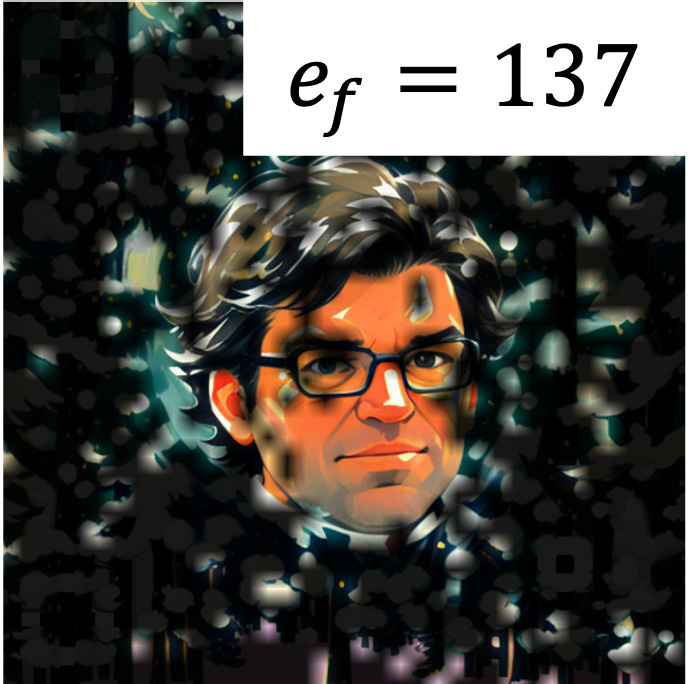} & 
        \includegraphics[width=\aww, height=\aww]{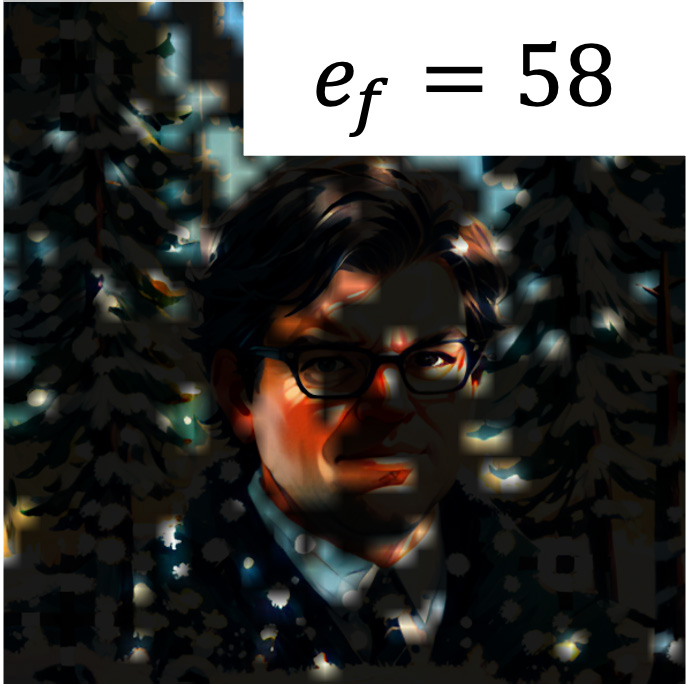} \\
    
        \bottomrule
        \end{tabularx}
    \end{minipage}\hfill
    \begin{minipage}[b]{.53\linewidth}
        \caption{IDSE Ablation Study: We examine the influence of $w_f, w_b, \sigma_f, \sigma_b$ to the generated QR image $I^o$. Model 1 $(w_f=w_b, \sigma_f=\sigma_b)$ tends to create undesirable shadow in the face region. Model 2 $(w_f=w_b, \sigma_f<\sigma_b)$ leads to an increased number of error modules $e$. Model 3 $(w_f<w_b, \sigma_f<\sigma_b)$ reaches a balance between face ID and scannability. Second row zooms in on the face region.}
        \label{tab:IDSE}
        \centering
        \begin{tabularx}{\linewidth}{>{\centering\arraybackslash}X *{5}{>{\centering\arraybackslash}X}}
    
        \toprule
        $I^s$ & Model 1 
        & Model 2 
        & Model 3 
        \\ \midrule
    
        \includegraphics[width=\aww, height=\aww]{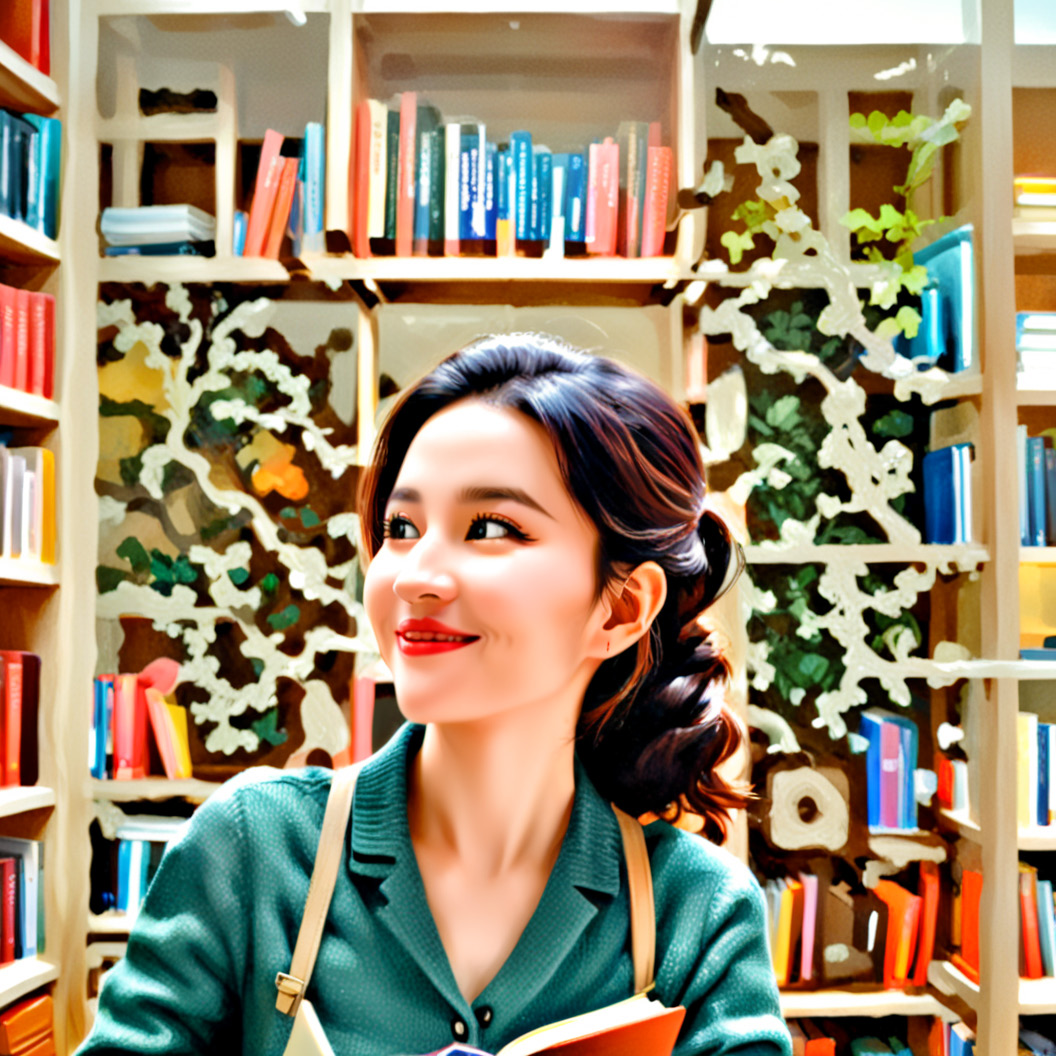} & 
        \includegraphics[width=\aww, height=\aww]{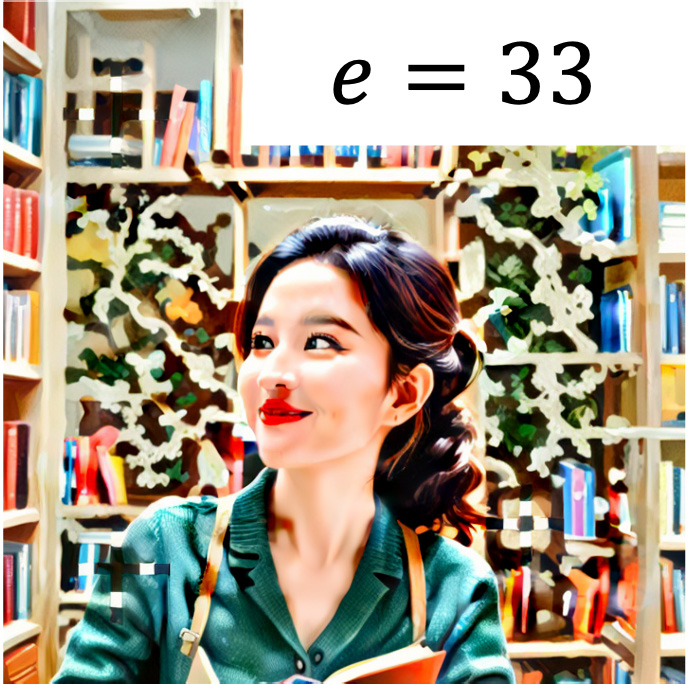} & 
        \includegraphics[width=\aww, height=\aww]{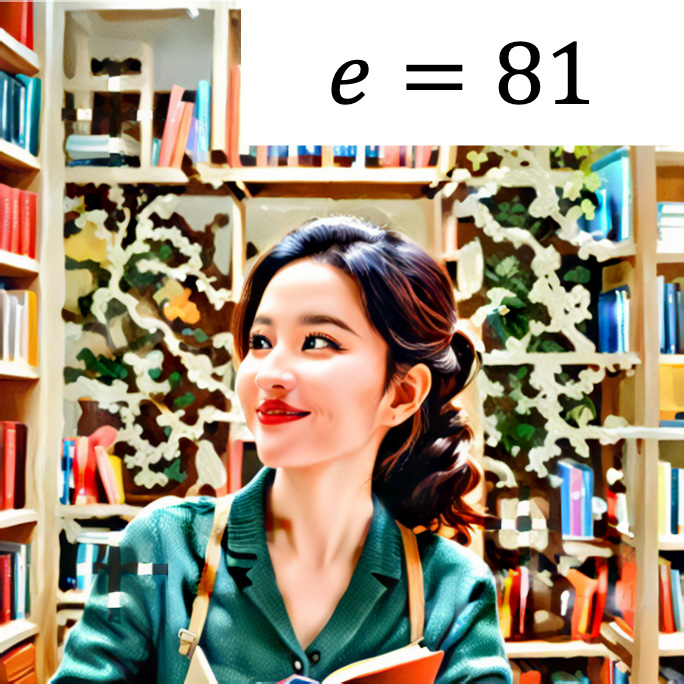} & 
        \includegraphics[width=\aww, height=\aww]{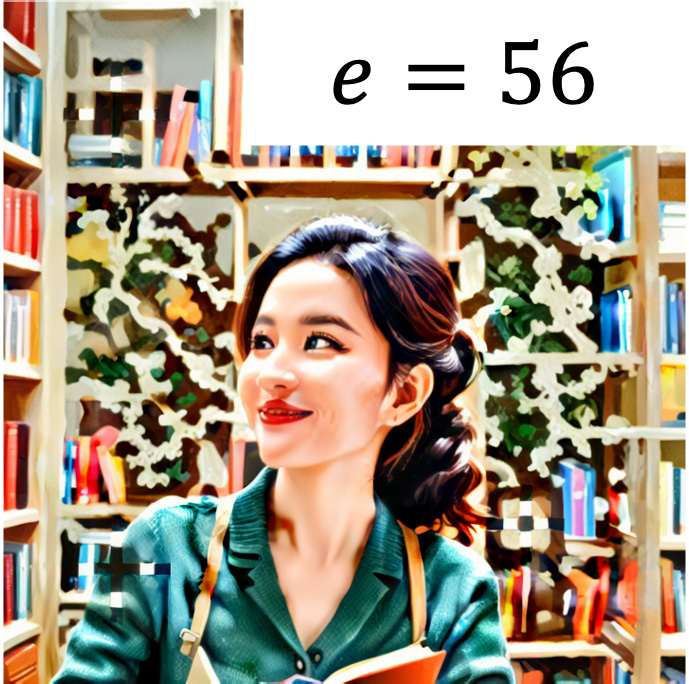} \\

        \includegraphics[width=\aww, height=\aww]{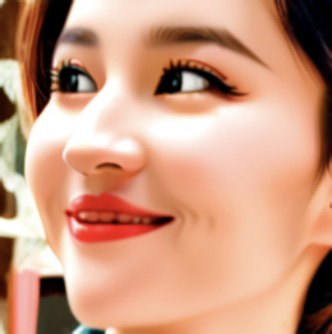} & 
        \includegraphics[width=\aww, height=\aww]{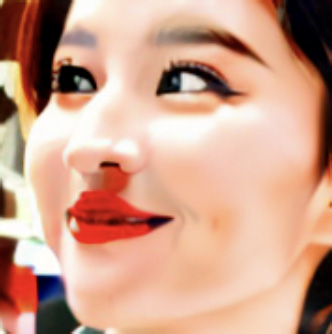} & 
        \includegraphics[width=\aww, height=\aww]{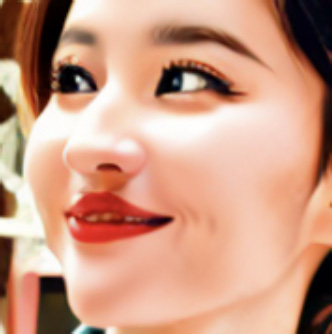} & 
        \includegraphics[width=\aww, height=\aww]{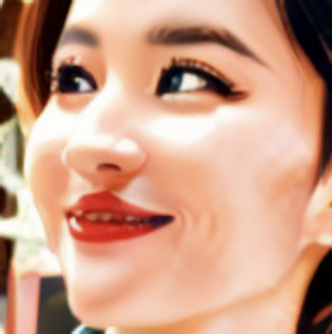} \\
    
        \bottomrule
        \end{tabularx}
    \end{minipage}
    \vspace{-10pt}
\end{table*}

\paragraph{IDRE Module.}
Table~\ref{tab:IDRE} illustrates how skipping stage 2 or the IDRE module affects the scannability of resultant QR image $I^s$. In stage 2, IDRE first rearranges the QR modules to construct a blueprint $I_b$ in which the QR pattern is compatible with face ID, and then SD model generates $I^s$ guided by $I_b$. If stage 2 is bypassed, the output $I^g$ from stage 1 is used as the input for stage 3 directly. If IDRE is omitted, a normal QR code is used to guide the QR pattern generation in stage 2. The result shows that skipping stage 2 or IDRE results in a substantial increase in the number of error modules $e_f$ within the facial area, which significantly reduces the scannability of the QR code.

\paragraph{IDSE Module.} 
In stage 3, the IDSE module leverages adaptive code loss, which is key to maintaining face identity and simultaneously decreasing number of error modules. This loss function is determined by two parameter pairs ($\sigma_f$ and $\sigma_b$ for Gaussian kernel; $w_f$ and $w_b$ for loss strength). Table~\ref{tab:IDSE} presents images produced by the IDSE with varying parameter configurations. The comparison shows that a uniform $\sigma$ leads to distortions in the facial area, and a universal loss weight $w$ will increase the number of error modules and compromise the generated image's scannability. A clearer comparison is given by the normalized image difference visualization $D$ shown in Table~\ref{tab:Ddetail}. The visualization $D$ demonstrates the module-wise difference between the QR images before and after IDSE. The results demonstrate that the adaptive loss makes the modification in the face region gentler than uniform loss, reaching a better balance between face identity and scannability.

\begin{table*}[h!]

    \centering
    \newlength{\Blen}
    \setlength{\Blen}{0.156\linewidth}
    \newlength{\Dlen}
    \setlength{\Dlen}{0.156\linewidth}
    
    \begin{minipage}[b]{.54\linewidth}
        \caption{Image difference visualization $D$ of uniform loss (first row) and adaptive loss (second row).}

        \label{tab:Ddetail}
        \centering
        \begin{tabularx}{\linewidth}{>{\centering\arraybackslash}X *{2}{>{\centering\arraybackslash}X}}
    
        \toprule
        $I^o$ & $D$
        & face region
        \\ \midrule

        \includegraphics[width=\Blen, height=\Blen]{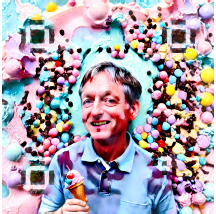} & 
        \includegraphics[width=\Blen, height=\Blen]{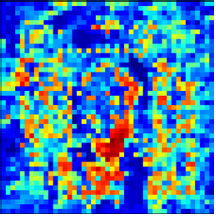} & 
        \includegraphics[width=\Blen, height=\Blen]{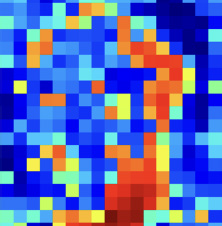}  \\

        \includegraphics[width=\Blen, height=\Blen]{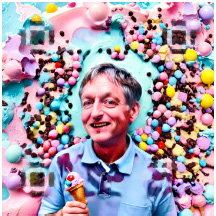} & 
        \includegraphics[width=\Blen, height=\Blen]{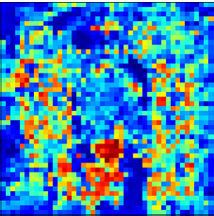} & 
        \includegraphics[width=\Blen, height=\Blen]{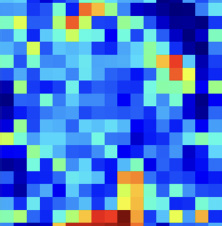}  \\
        
        \bottomrule
        \end{tabularx}
    \end{minipage}\hfill
    \begin{minipage}[b]{.40\linewidth}
    \caption{Bad cases caused by failure of generative model.}

        \label{tab:BadDiff}
        \centering
        
        \begin{tabularx}{\linewidth}{>{\centering\arraybackslash}X >{\centering\arraybackslash}X *{1}{>{\centering\arraybackslash}X}}
    
        \toprule
        Input & Output \\ \midrule
    
        \includegraphics[width=\Dlen, height=\Dlen]{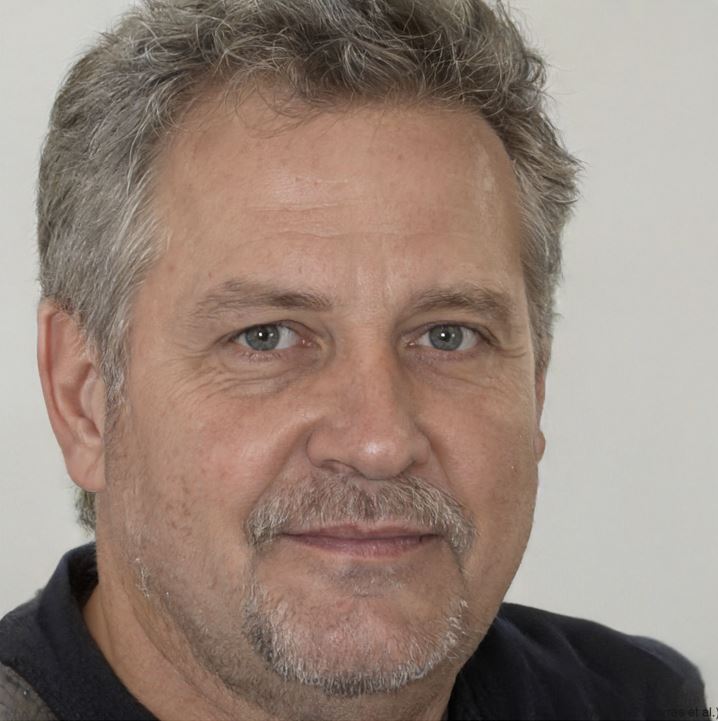} & 
        \includegraphics[width=\Dlen, height=\Dlen]{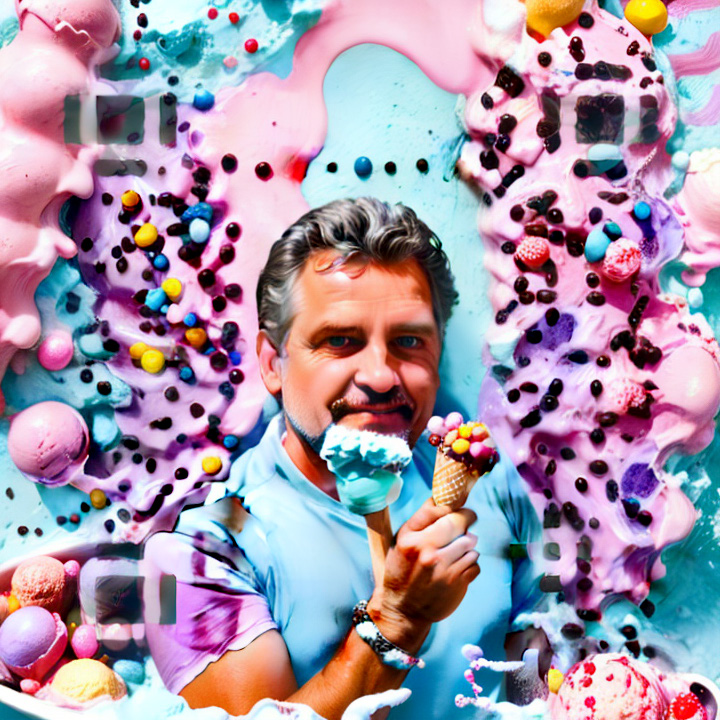} \\

        \includegraphics[width=\Dlen, height=\Dlen]{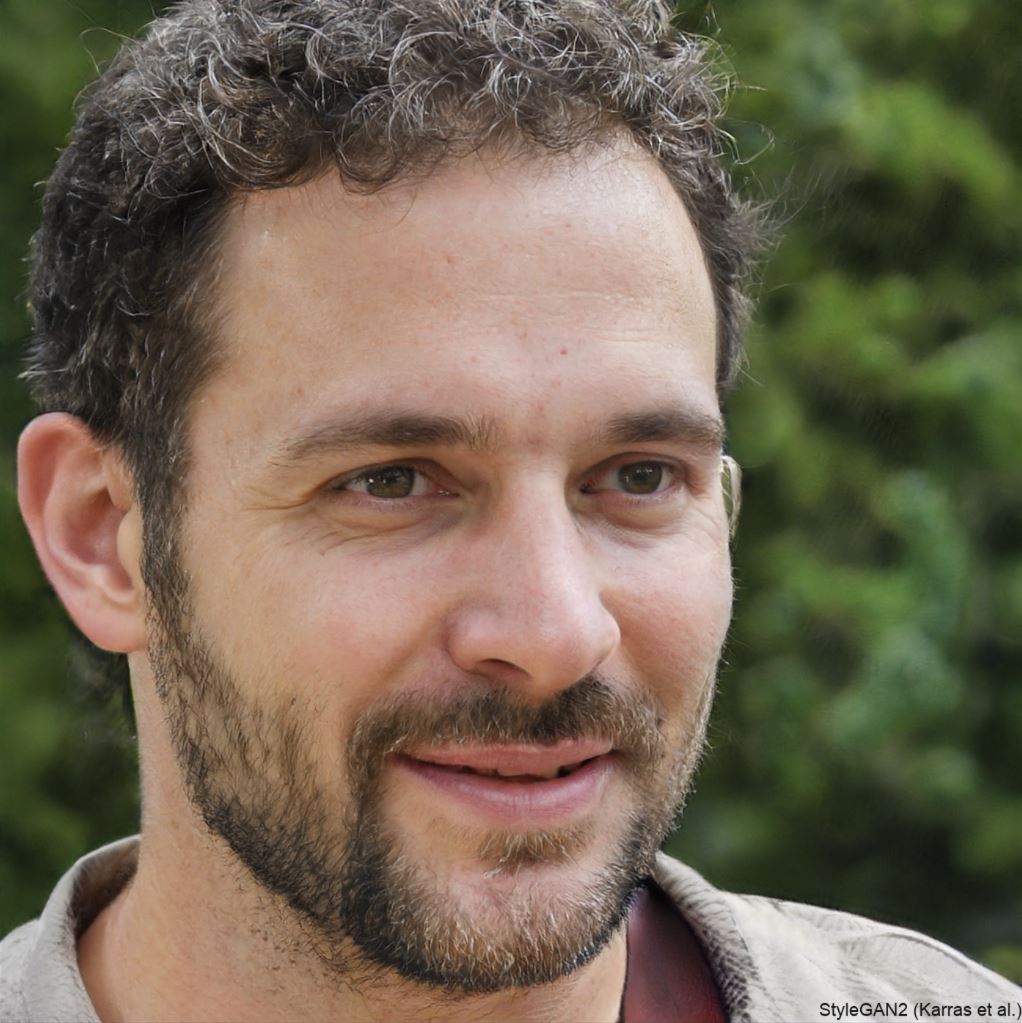} & 
        \includegraphics[width=\Dlen, height=\Dlen]{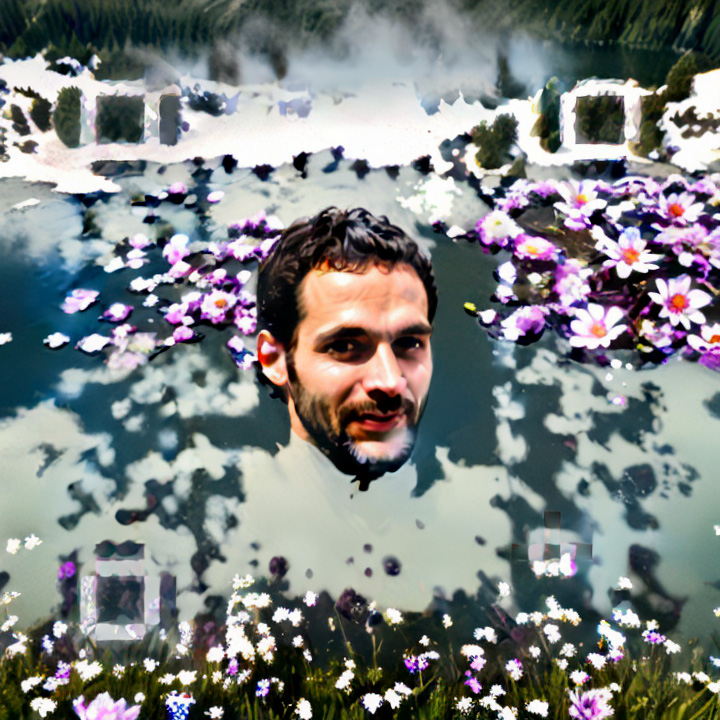} \\
    
        \bottomrule
        \end{tabularx}
    \end{minipage}
\end{table*}

\section{Conclusion}\label{sec:conclusion}

In summary, we introduce Face2QR, an innovative pipeline that seamlessly integrates face ID, aesthetic design and scannability in the generation of QR codes. By introducing three key modules, \textit{i.e.} IDQR for integrating face ID with aesthetic background, IDRS for resolving conflict between face ID and QR pattern, and IDSE for enhancing scannability while preserving face ID and aesthetic quality, our pipeline is able to balance between three inherently conflicting control signals and generate QR codes that preserve face ID, aesthetic quality and scannability at the same time. Extensive experiments demonstrate that Face2QR significantly outperforms previous methods, establishing a new benchmark for generating ID-preserved aesthetic QR codes.

\paragraph{Limitations.} Our method is still constrained by some limitations of generative models. Although generative models are powerful, they can produce inconsistent results and often require substantial computing power to generate detailed, high-resolution images. Some typical bad cases caused by failure of generative models are shown in Table~\ref{tab:BadDiff}. As these computational models become more advanced, we can anticipate further improvements in the accuracy, speed, and overall aesthetic quality of the generated QR codes.

\paragraph{Broader Impact.} By enhancing the visual appeal and personal connection of QR codes, our work has the potential to revolutionize their use in entertainment, social media, marketing, and personal memorabilia, transforming them from mere tools for information transfer into objects of personal expression and aesthetic value. Looking forward, we anticipate that future work will not only refine these methods but also explore their integration into various technological ecosystems, consistently enriching the social and functional aspects of QR codes.

\paragraph{Acknowledgement.} The work was supported in part by the National Natural Science Foundation of China under Grant 62301310 and 62225112, and in part by Sichuan Science and Technology Program under Grant 2024NSFSC1426.

\clearpage\newpage
{
    \small
    \bibliographystyle{ieee_fullname}
    \bibliography{main}
}

\clearpage
\setcounter{page}{1}
\appendix

\section{{Appendix}}

\subsection{Additional Experiments}
\subsubsection{Additional Results}
Our Face2QR pipeline is generalizable to real faces, generated realistic faces, and cartoon faces. The experimental results in Table~\ref{tab:more_results} demonstrate that facial identities are well preserved and seamlessly blended into the background in all generated QR images, showcasing the effectiveness of Face2QR across these three face types.

\subsubsection{Visualization of Intermediate Results}
In Table~\ref{tab:comparison1}, we show the intermediate results of Face2QR. Here, $I^{g}$ represents the output of stage 1, $I_b$ signifies blueprint image generated by IDRE, and $I^s$ denotes the results of regeneration results from stage 2. The blueprint $I_b$ guides both the generation of $I^s$ in IDQR module within stage 2, and the IDSE process in stage 3 to generate QR image $I^o$ with a harmonious balance between face ID, aesthetic quality and scannability. Table~\ref{tab:selr} presents results from different iterations in the IDSE process. Additionally, Table~\ref{tab:comparison3} presents the prompts and models used in the generation of the aforementioned QR image samples.

\subsubsection{Loss Curve \& Running Time}

\begin{figure}[h]
    \vspace{-5pt}
    \centering
    \includegraphics[width=\linewidth]{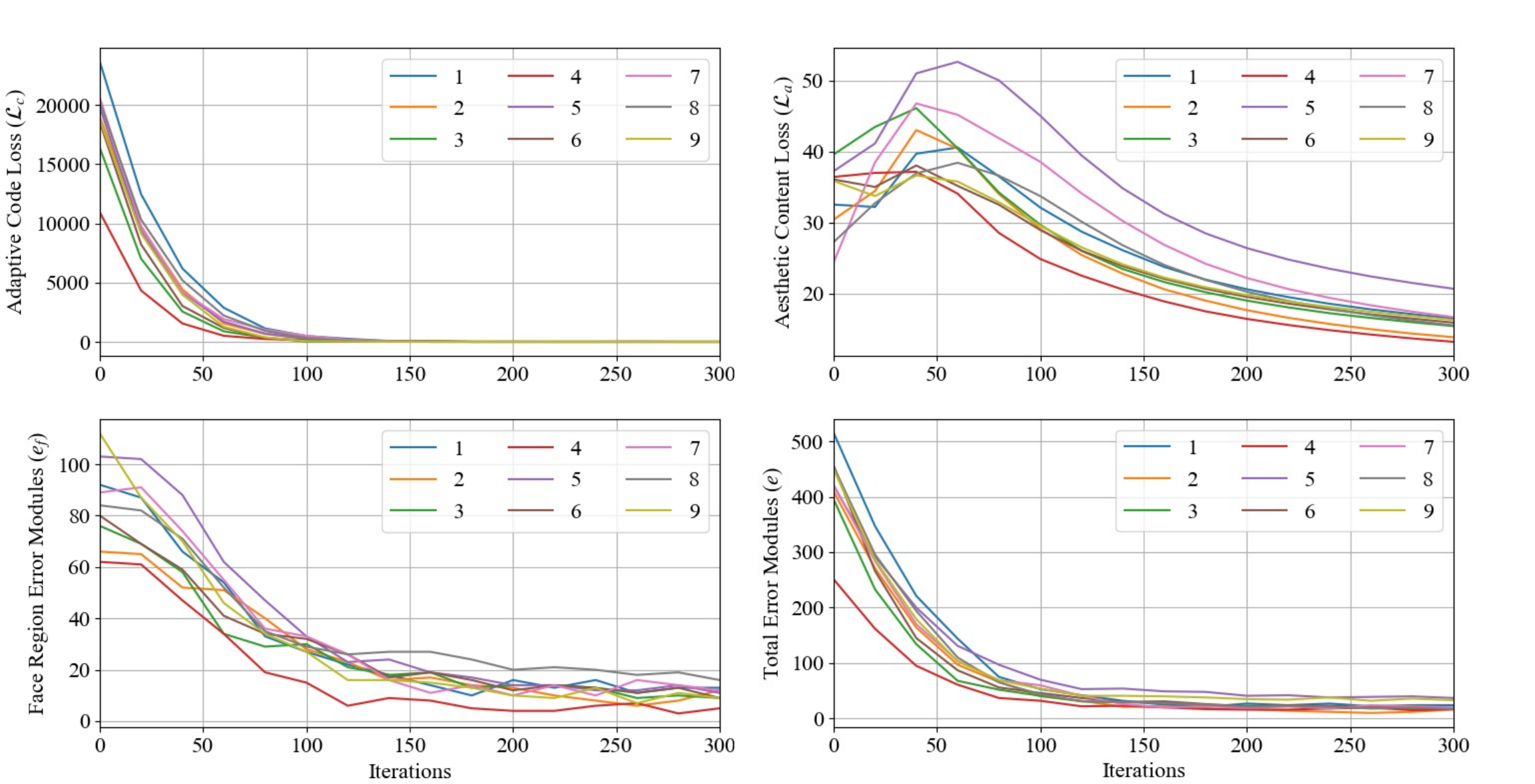}
    \caption{Curve of different metrics during IDSE. We show metric curves for diverse samples, each represented by distinct colors. These curves illustrate metric variations over 300 iterations.}
    \label{fig:curve}
    \vspace{-5pt}
\end{figure}

In stage 3, we use IDSE to enhance the scannability of $I^s$ by updating the its latent code. The dynamic loss function consists of aesthetic content loss $\mathcal{L}_a$ and adaptive code loss $\mathcal{L}_c$. Both losses apply at the same time and helps the update process to converge sooner. The total number of error module $e$ acts as a indicator of scannability, and the error module number in the face region $e_f$ helps visualize the modification process in the face region. Figure~\ref{fig:curve} illustrates the above four metrics. The IDSE process converges in about 120 seconds when executed on an NVIDIA 4090 GPU to enhance images of size 1024$\times$1024 pixels.

\subsection{Bad Cases}
In addition to bad cases shown in Table~\ref{tab:BadDiff}, we present suboptimal cases when the face image and the prompt are conflict with each other in Table~\ref{fig:bad}. For example, the first row in Table~\ref{fig:bad} shows the case when a face image of a woman and the prompt "A male man" are given at the same time.

\subsection{The Interface for User Study}
\begin{figure*}[ht]
    \centering
    \includegraphics[width=0.9\linewidth]{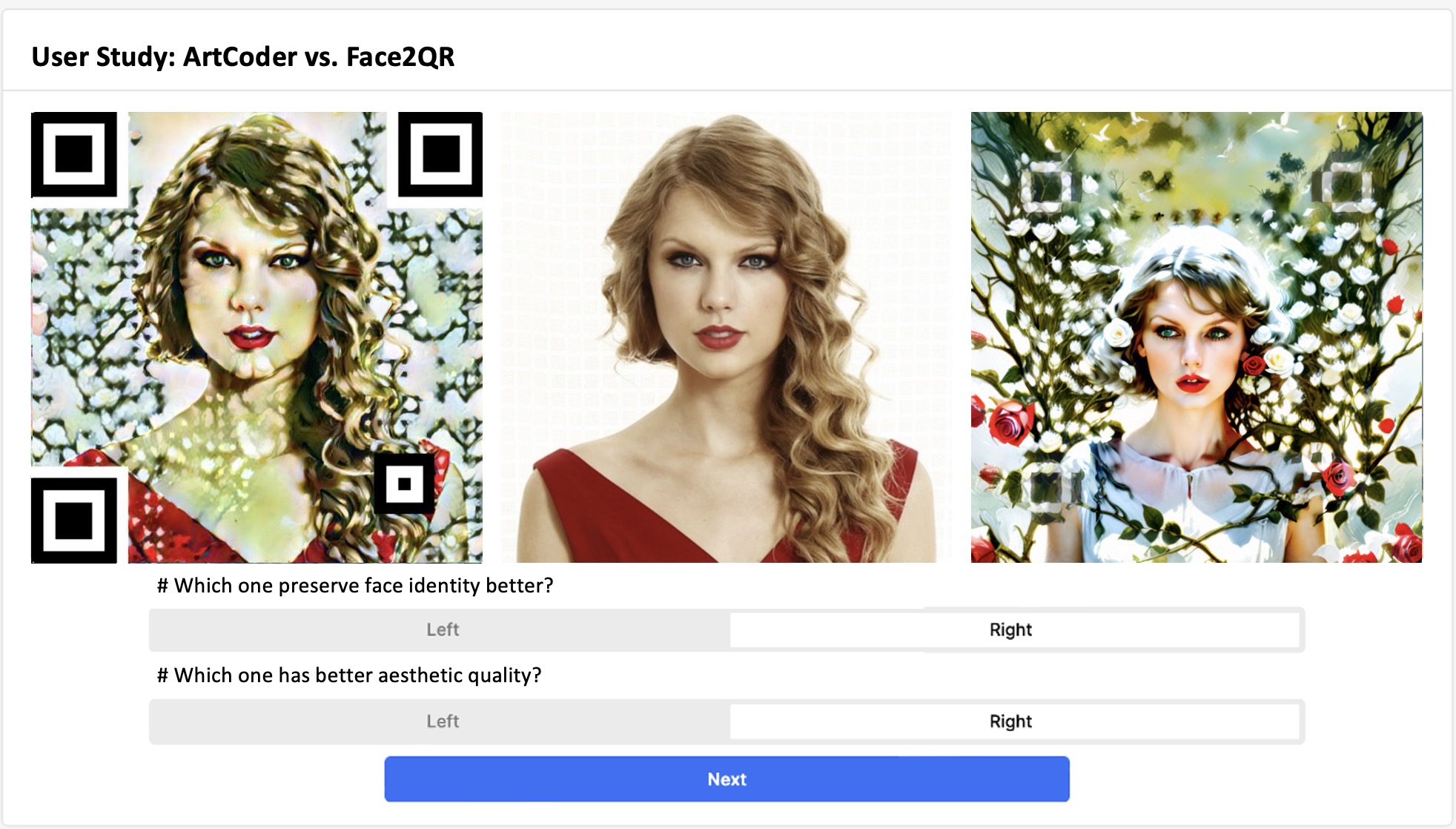}
    \caption{The user interface.
    }
    \label{fig:interface}
\vspace{-2mm}
\end{figure*}

The scoring interface of user study is shown in Figure~\ref{fig:interface}. We adopt the pair-wise comparisons for subjective study rather than absolute ratings since the former is relatively more accurate in general.

\vspace{30pt}
\begin{table*}[h]
    \newlength{\morewidth}
    \setlength{\morewidth}{0.156\textwidth}

    \caption{Real face images of ordinary people (Row 1) collected from~\cite{pexels}, generated realistic face (left three on Row 3) using StyleGAN2~\cite{Karras2019stylegan2} and cartoon faces (right three on Row 3) with corresponding QR images $I^o$ (Rows 2 and 4).
    }

    \label{tab:more_results}
    \centering
    \begin{tabularx}{\linewidth}{>{\centering\arraybackslash}X *{5}{>{\centering\arraybackslash}X}}

    \toprule

    \includegraphics[width=\morewidth, height=\morewidth]{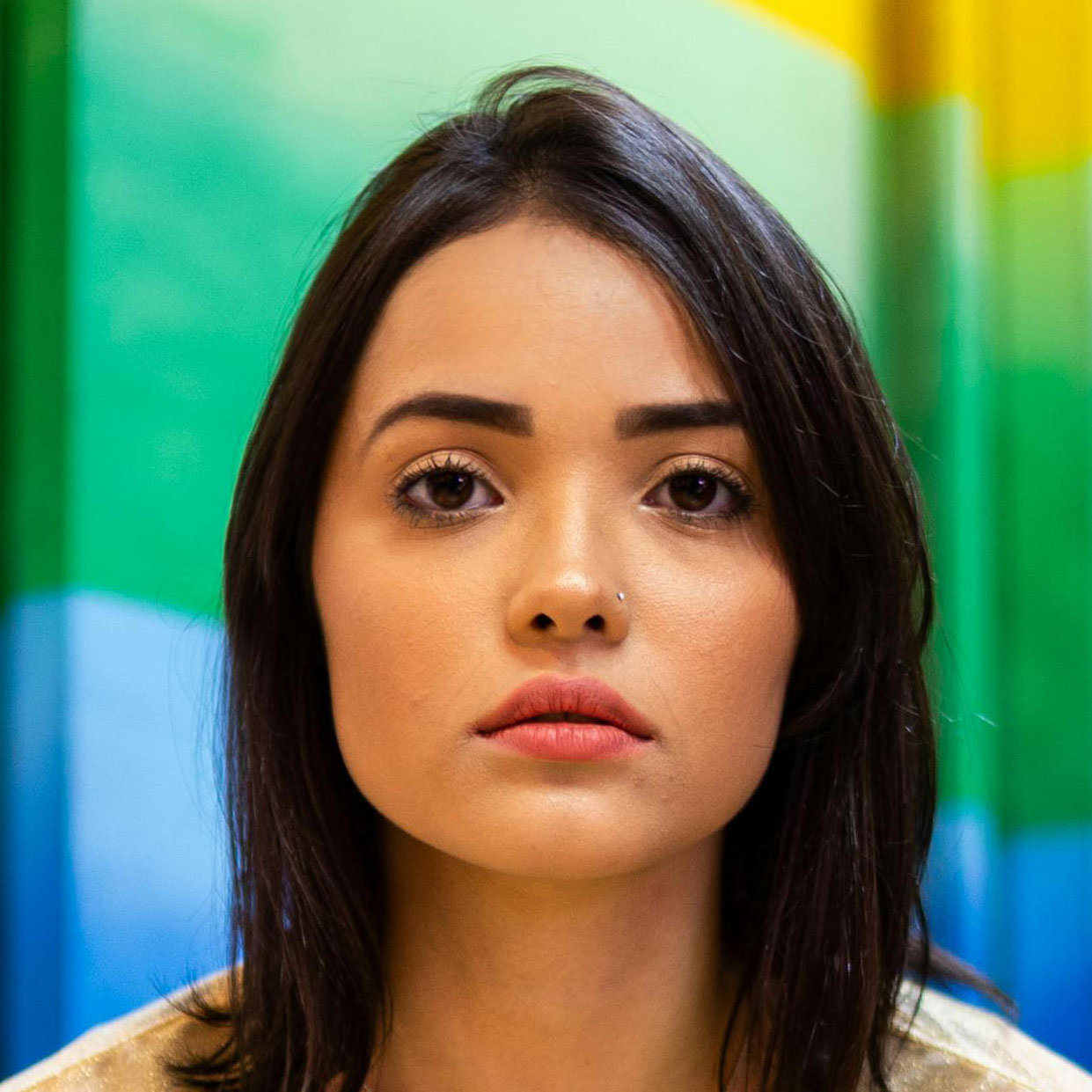} & 
    \includegraphics[width=\morewidth, height=\morewidth]{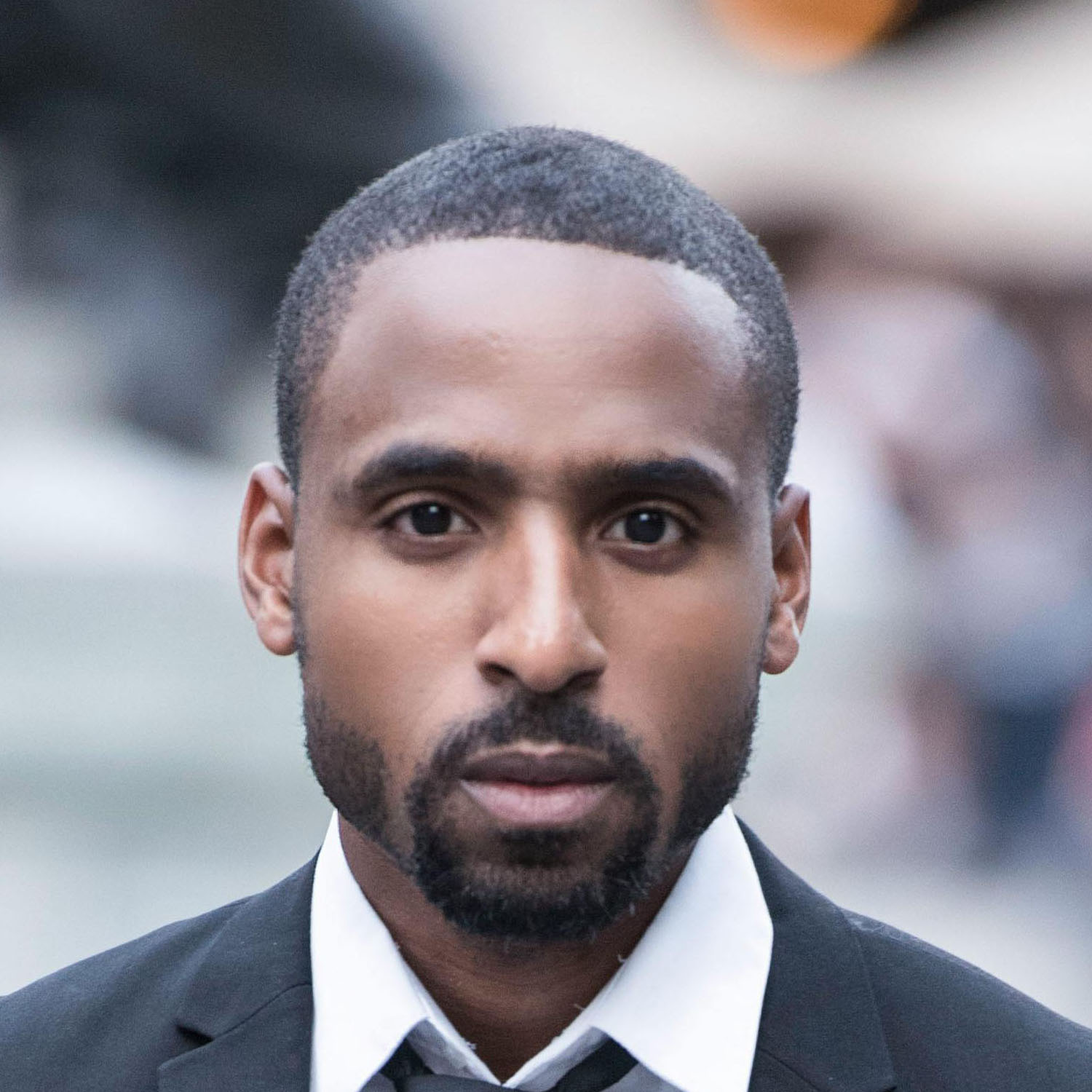} &
    \includegraphics[width=\morewidth, height=\morewidth]{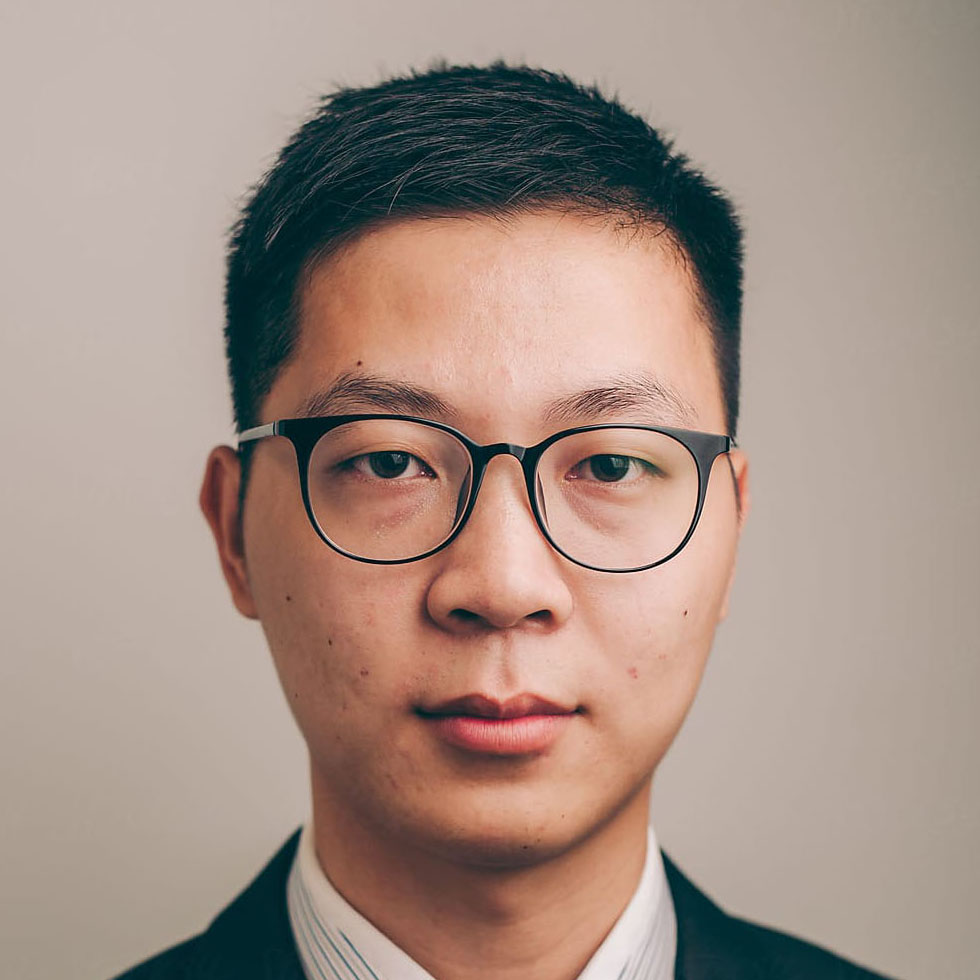} & 
    \includegraphics[width=\morewidth, height=\morewidth]{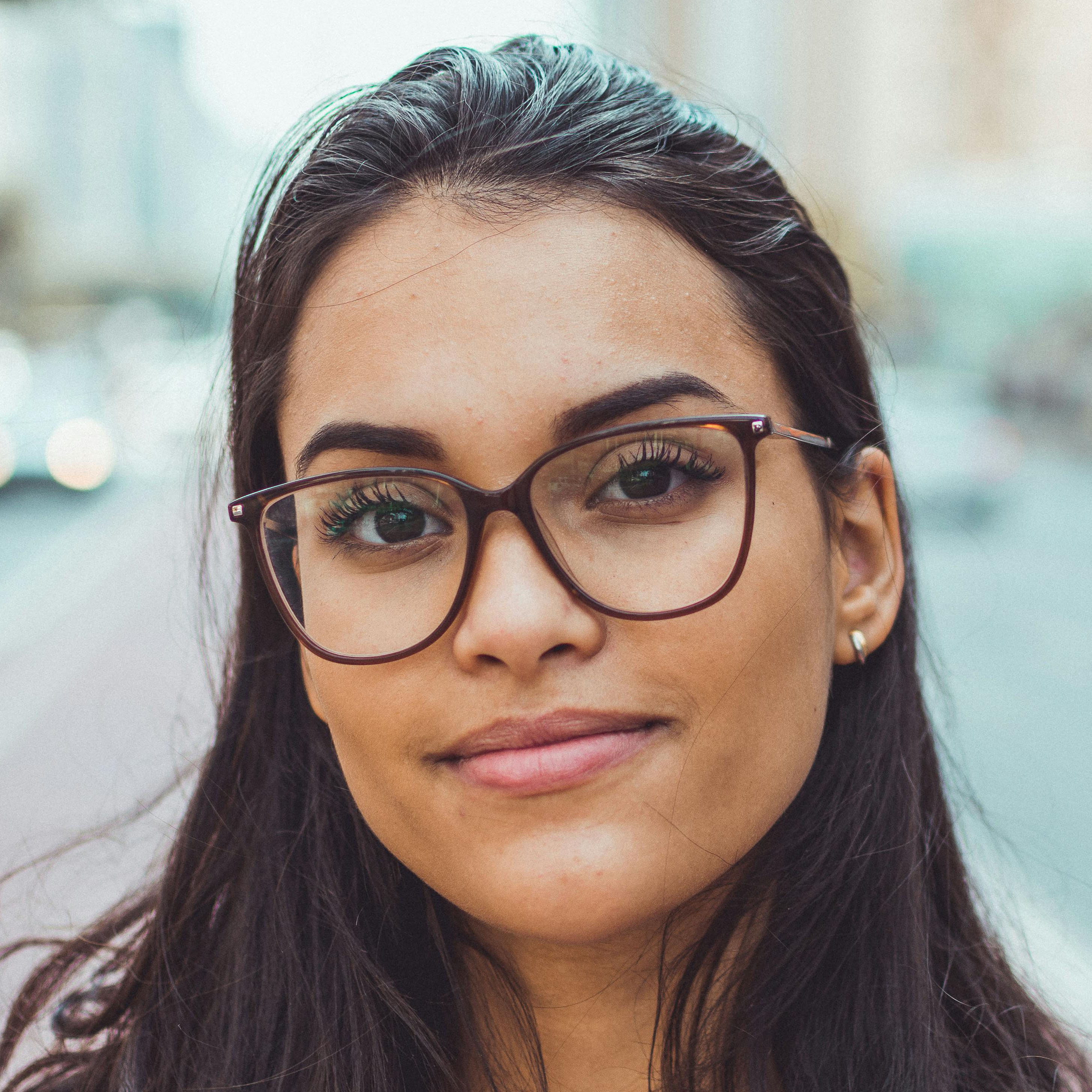} & 
    \includegraphics[width=\morewidth, height=\morewidth]{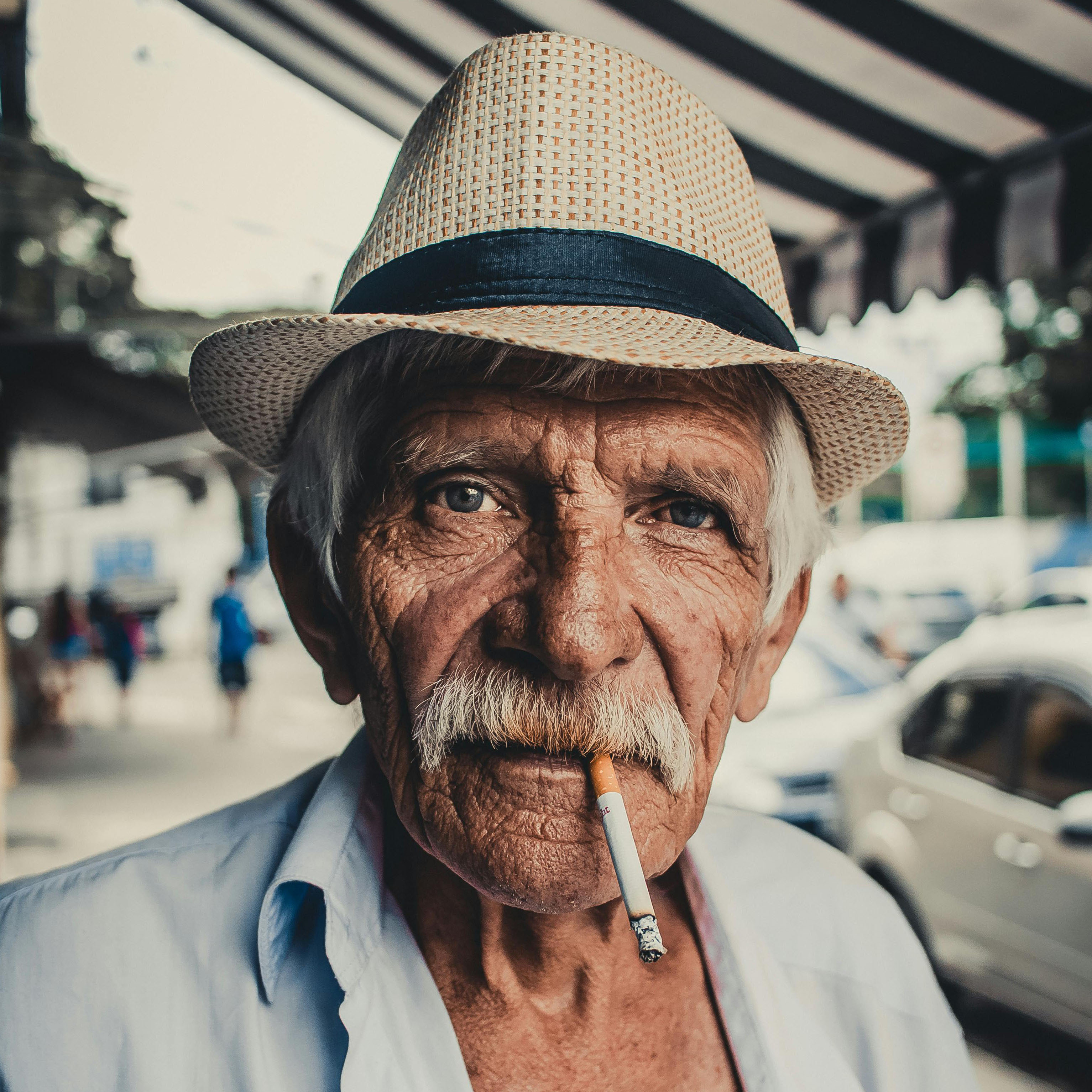} & 
    \includegraphics[width=\morewidth, height=\morewidth]{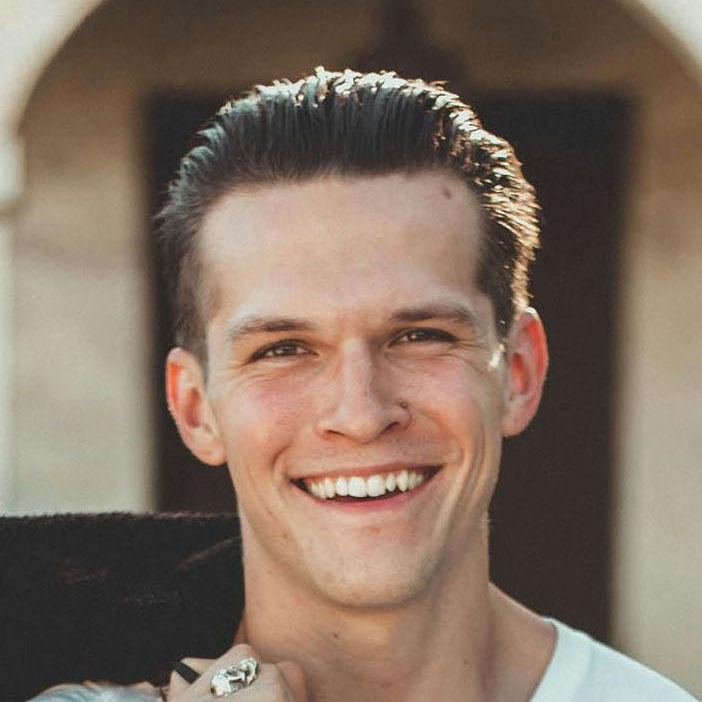} \\

    \includegraphics[width=\morewidth, height=\morewidth]{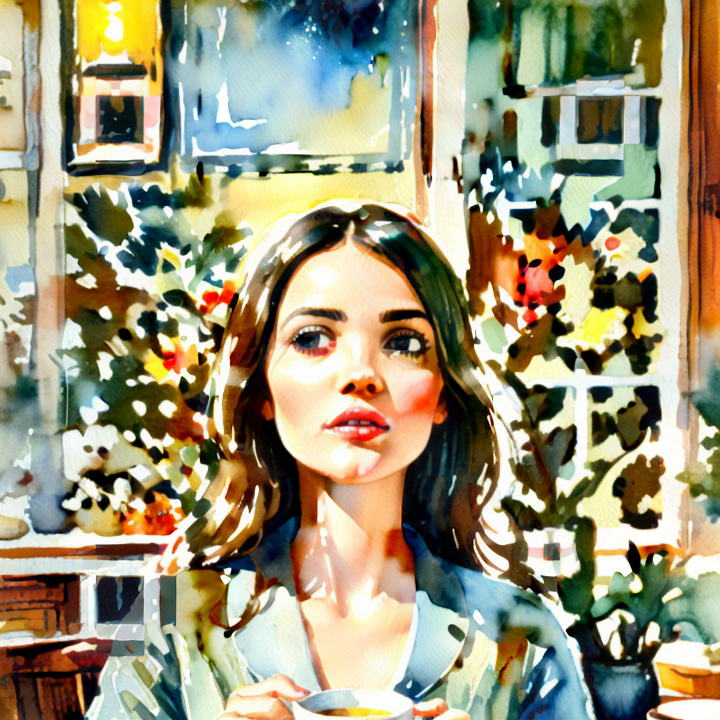} & 
    \includegraphics[width=\morewidth, height=\morewidth]{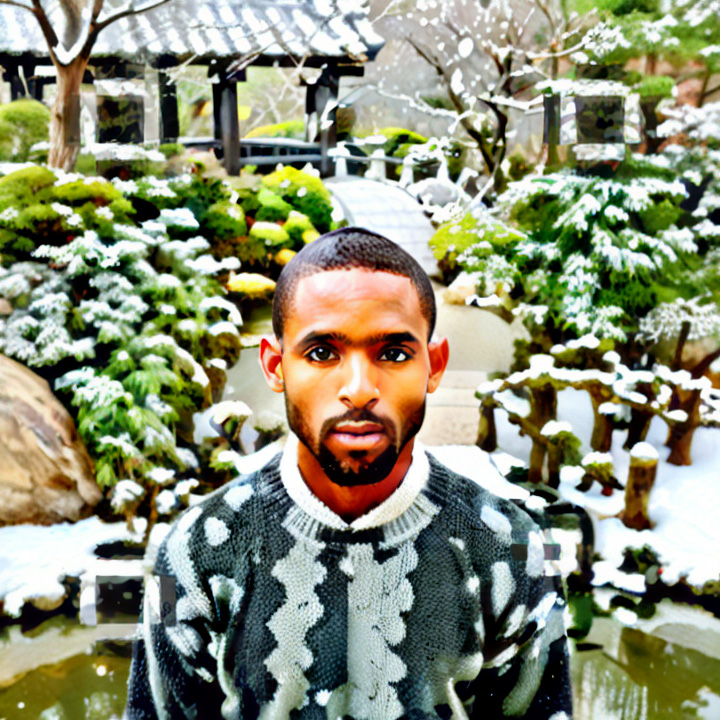} &
    \includegraphics[width=\morewidth, height=\morewidth]{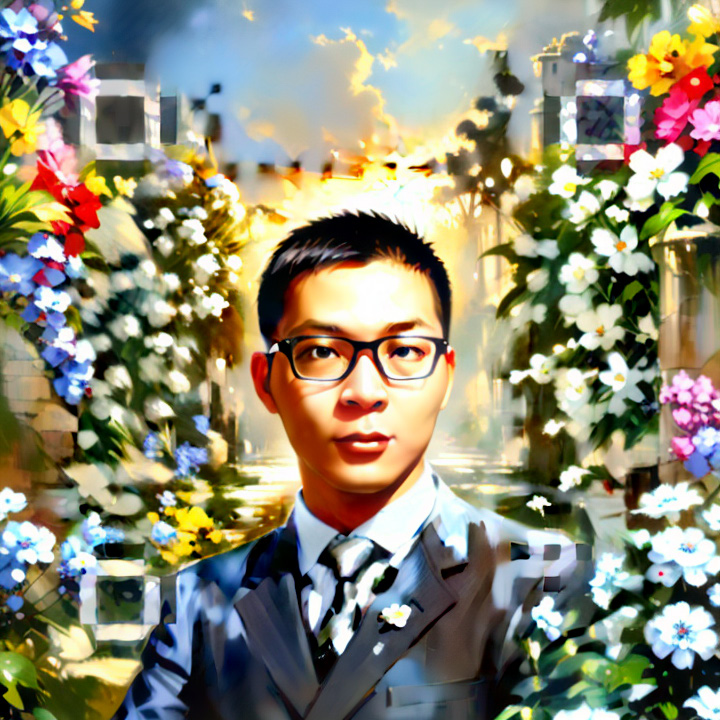} & 
    \includegraphics[width=\morewidth, height=\morewidth]{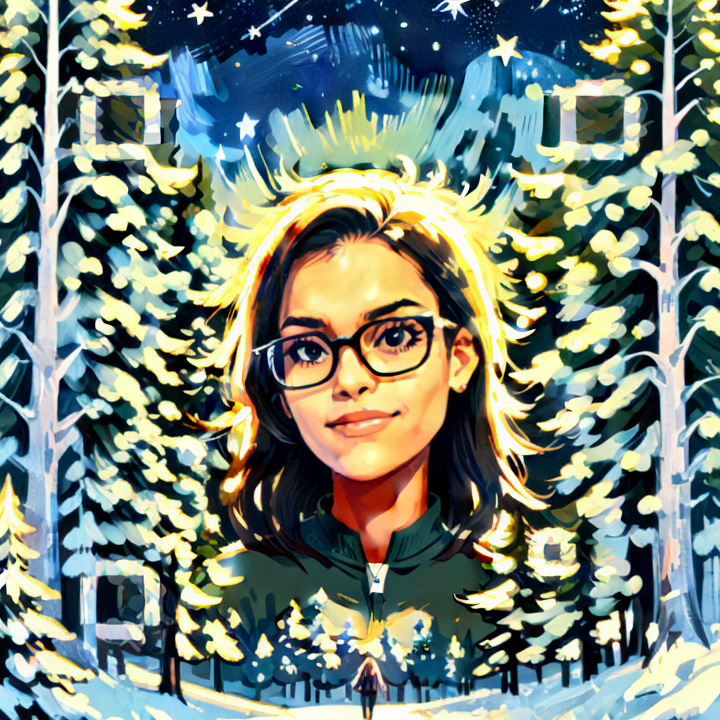} & 
    \includegraphics[width=\morewidth, height=\morewidth]{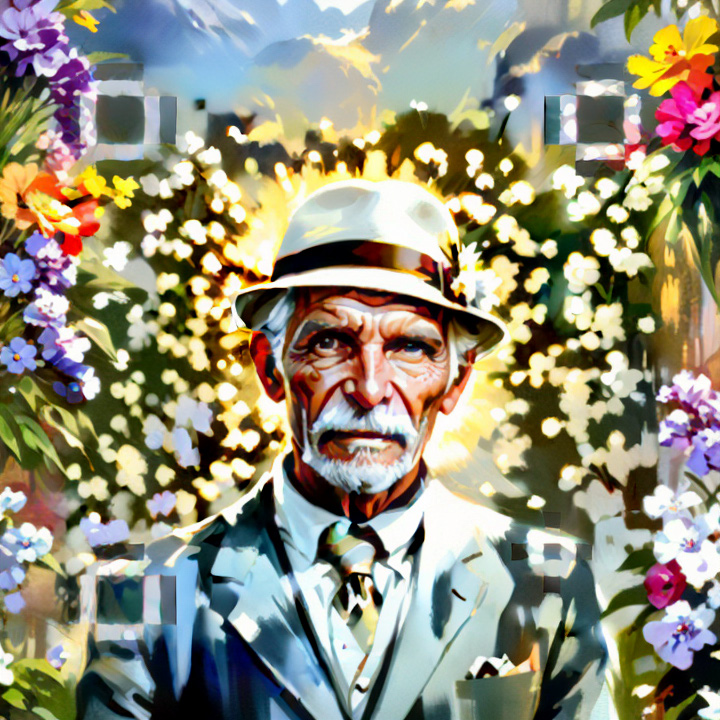} & 
    \includegraphics[width=\morewidth, height=\morewidth]{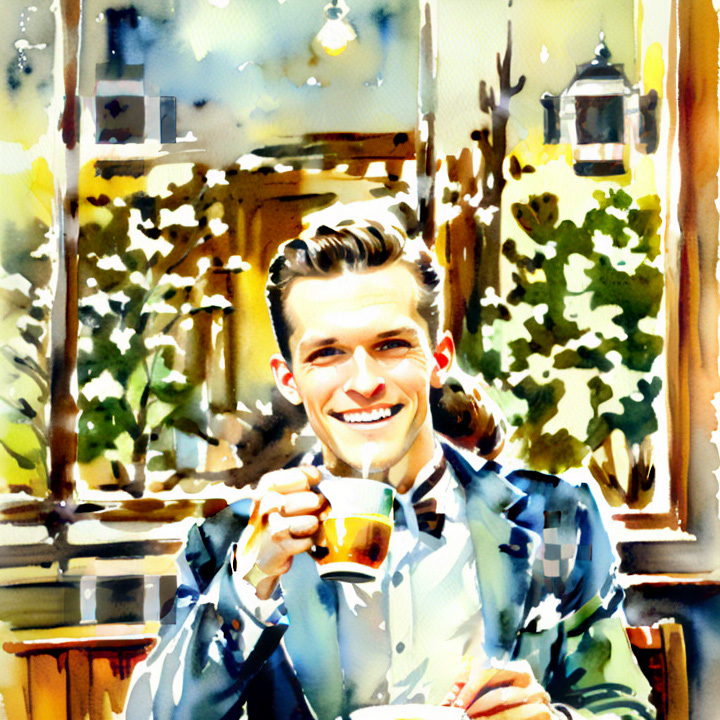} \\

    \midrule

    \includegraphics[width=\morewidth, height=\morewidth]{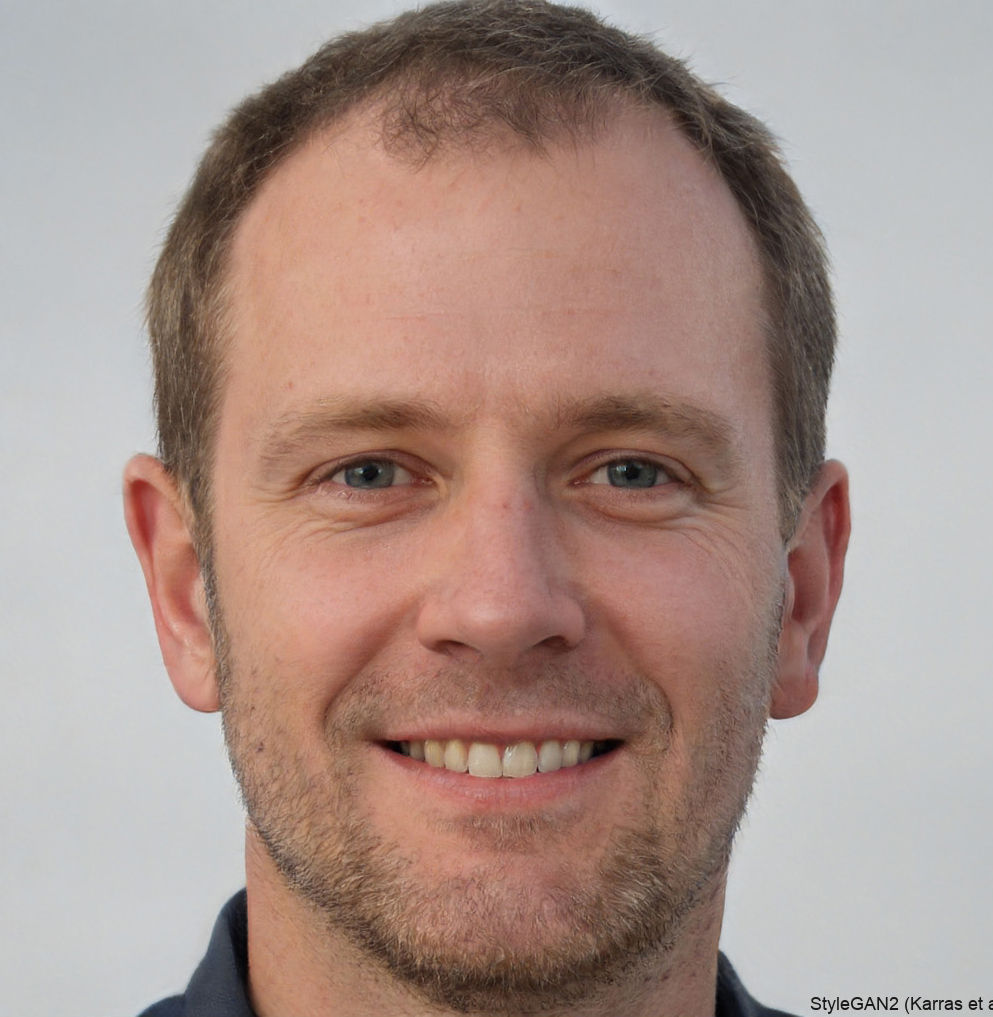} & 
    \includegraphics[width=\morewidth, height=\morewidth]{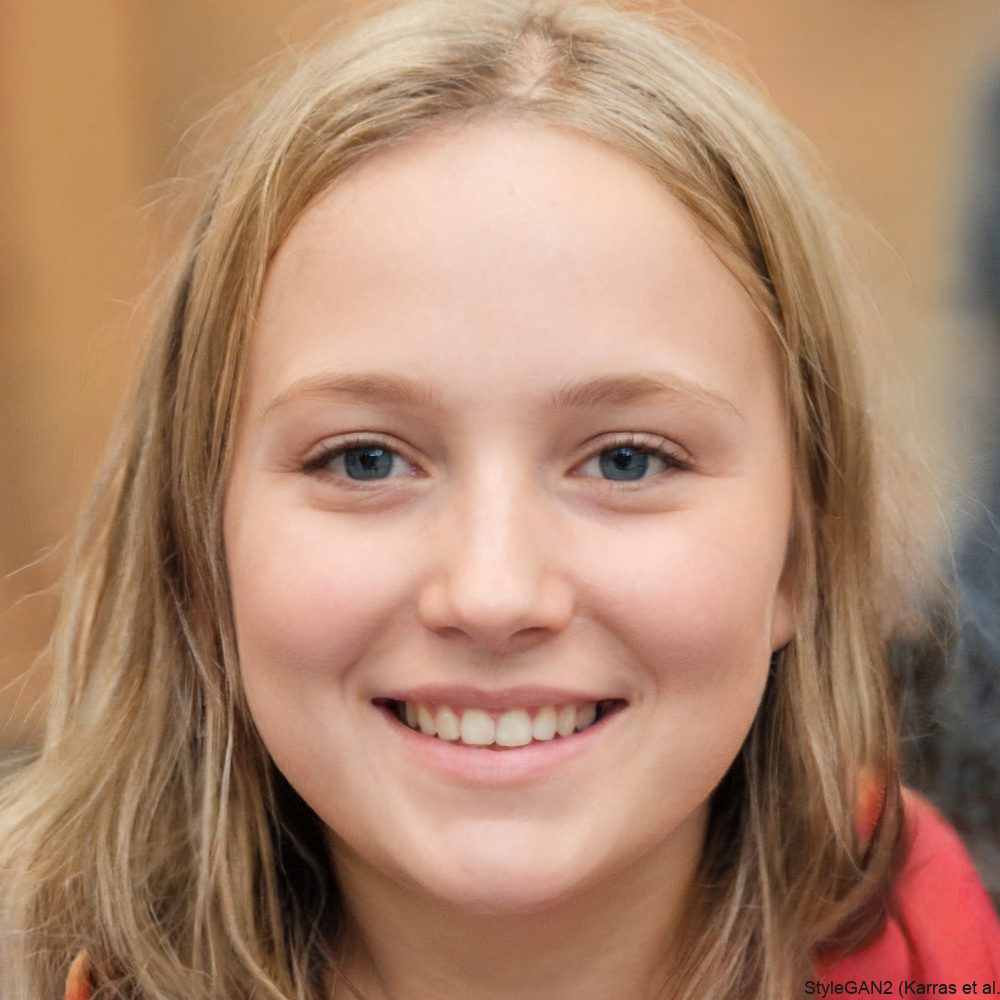} &
    \includegraphics[width=\morewidth, height=\morewidth]{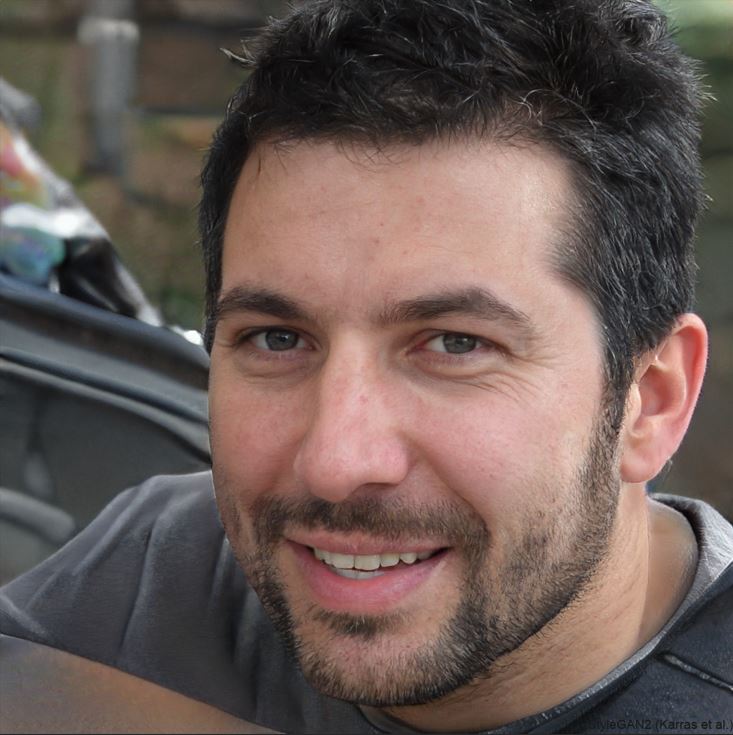} &
    \includegraphics[width=\morewidth, height=\morewidth]{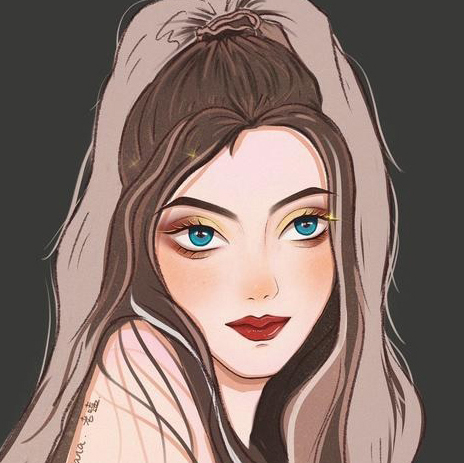} &
    \includegraphics[width=\morewidth, height=\morewidth]{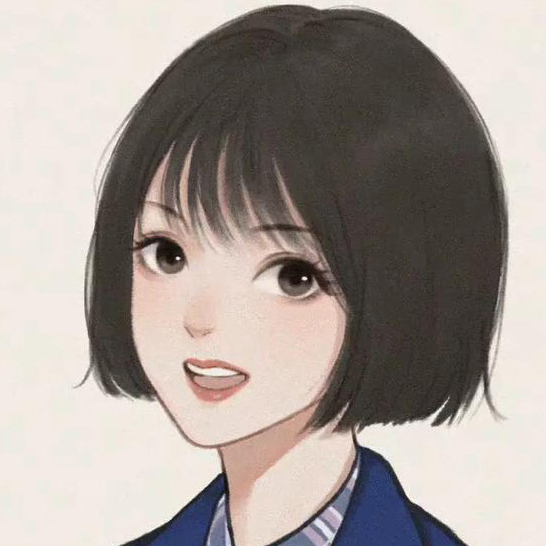} &
    \includegraphics[width=\morewidth, height=\morewidth]{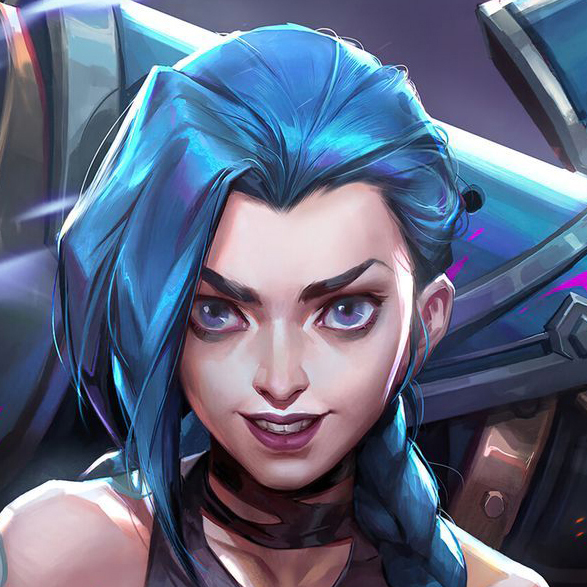} \\

    \includegraphics[width=\morewidth, height=\morewidth]{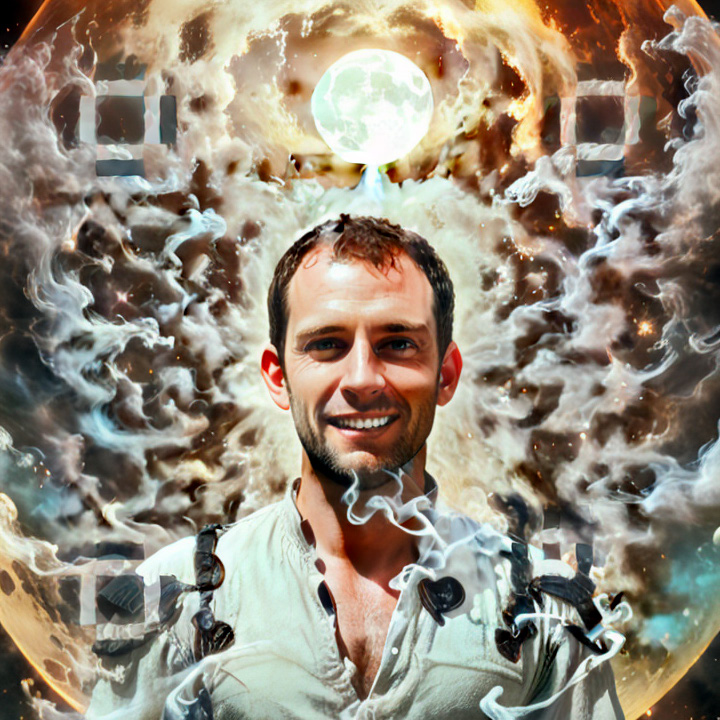} & 
    \includegraphics[width=\morewidth, height=\morewidth]{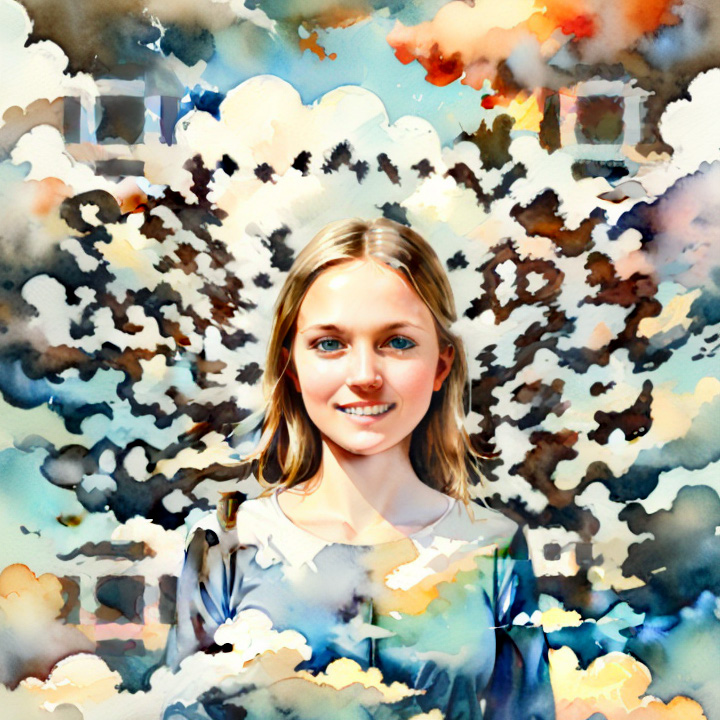} & 
    \includegraphics[width=\morewidth, height=\morewidth]{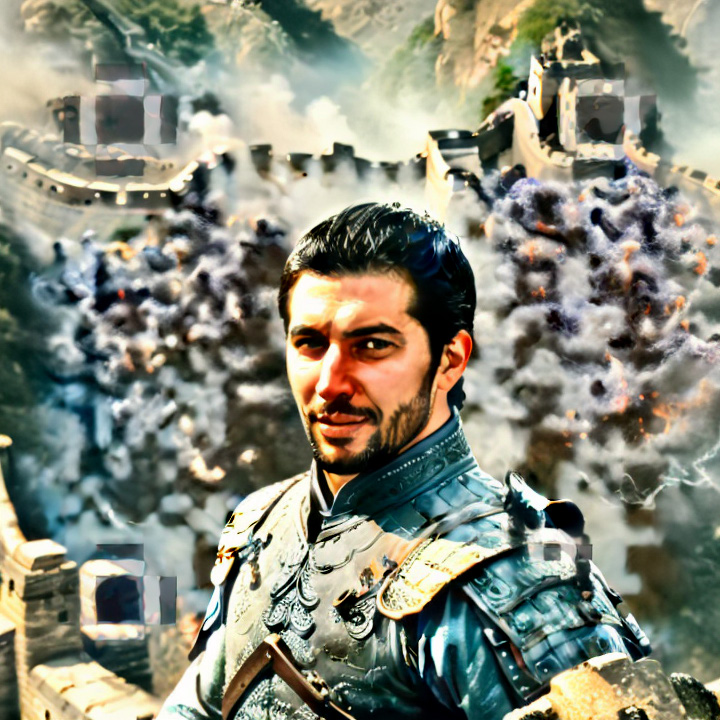} &
    \includegraphics[width=\morewidth, height=\morewidth]{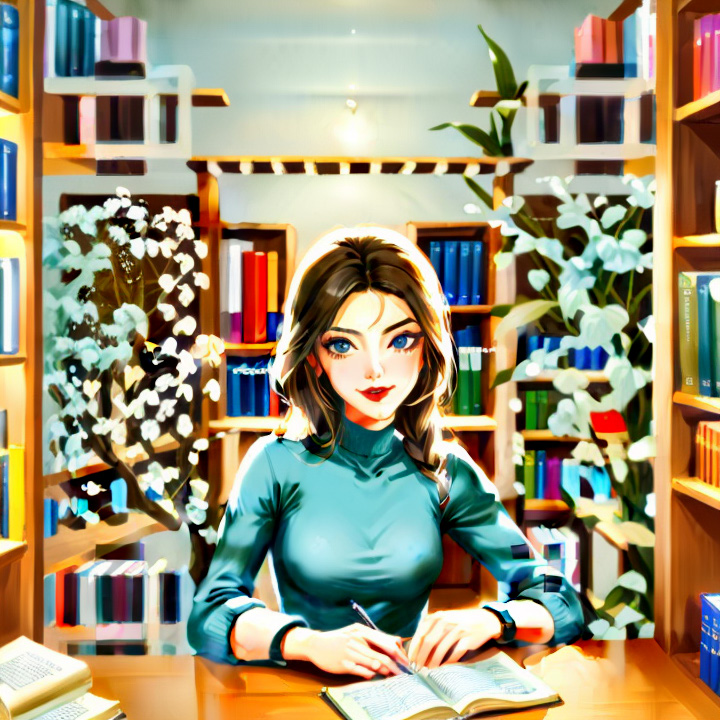} &
    \includegraphics[width=\morewidth, height=\morewidth]{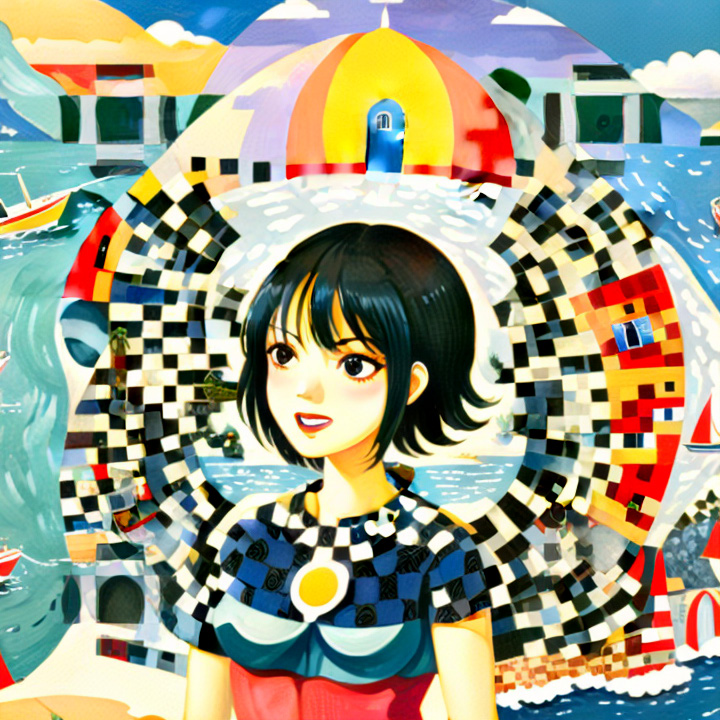} &
    \includegraphics[width=\morewidth, height=\morewidth]{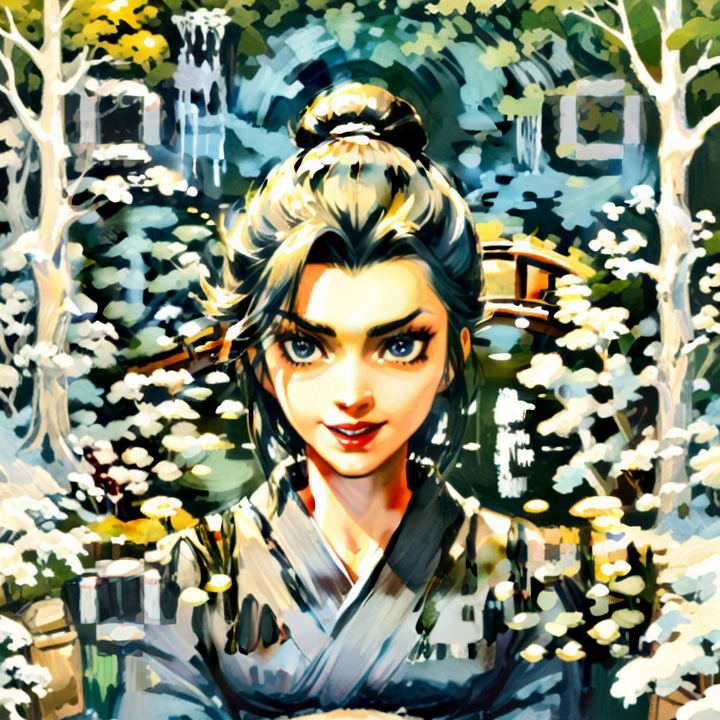}
    \\

    \bottomrule
    \end{tabularx}

\end{table*}

\begin{table*}[t]
    \centering
    \newlength{\picw}
    \setlength{\picw}{0.14\textwidth}
    \begin{minipage}[btp]{.6513\linewidth}
        
        \caption{Visualization of IDSE process at different iteration steps.
        }
        \label{tab:selr}
        \centering
        \begin{tabularx}{\linewidth}{>{\centering\arraybackslash}X *{4}{>{\centering\arraybackslash}X}}
    
        \toprule
        Input & Iteration 100 & Iteration 200 & Iteration 300  \\
        \midrule
    
        \includegraphics[width=\picw, height=\picw]{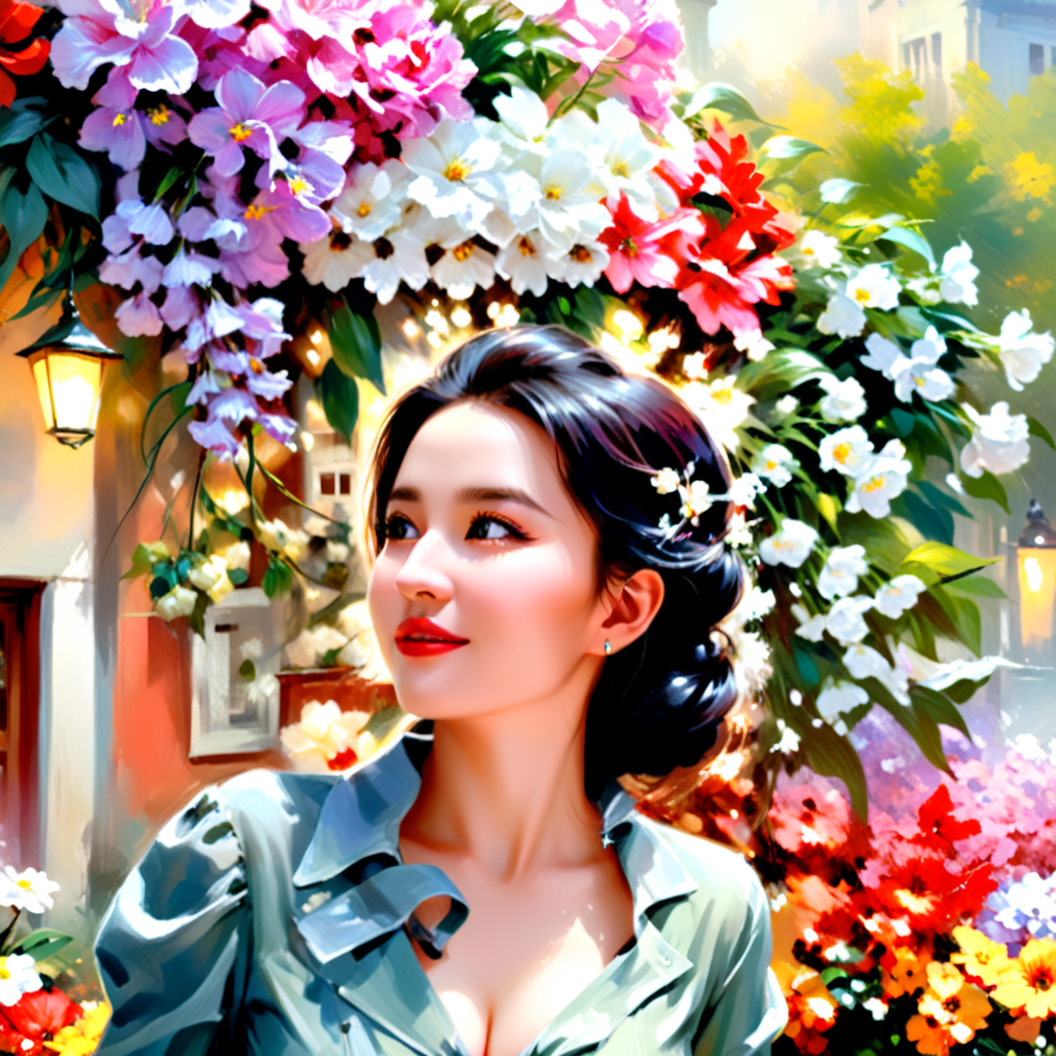} & 
        \includegraphics[width=\picw, height=\picw]{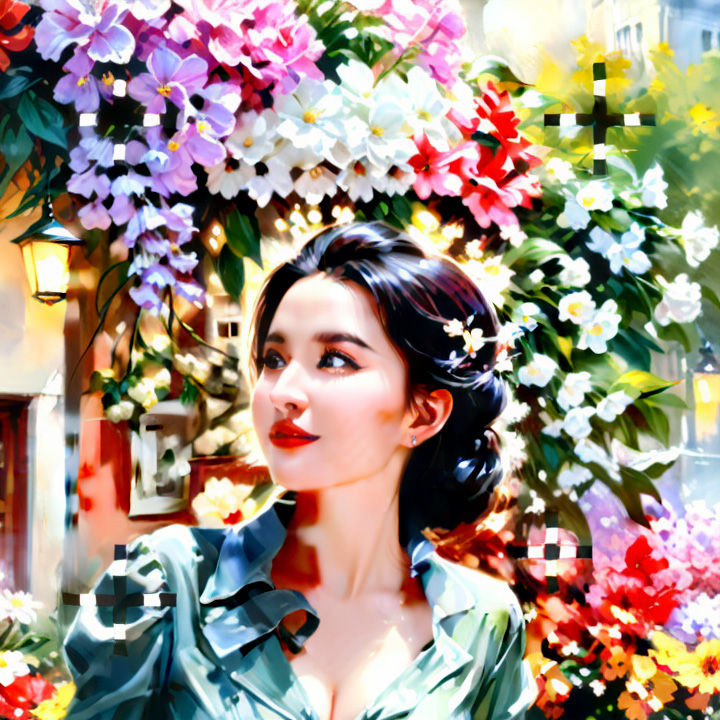} & 
        \includegraphics[width=\picw, height=\picw]{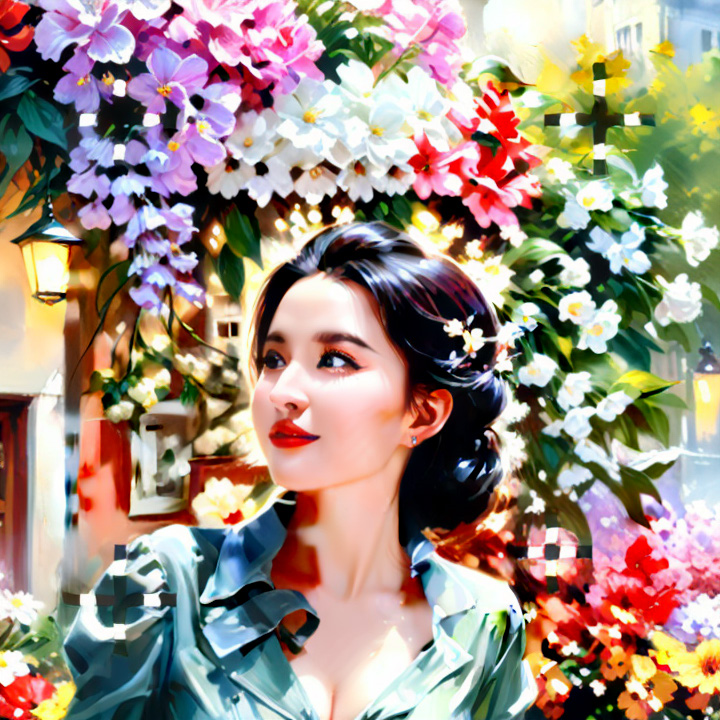} & 
        \includegraphics[width=\picw, height=\picw]{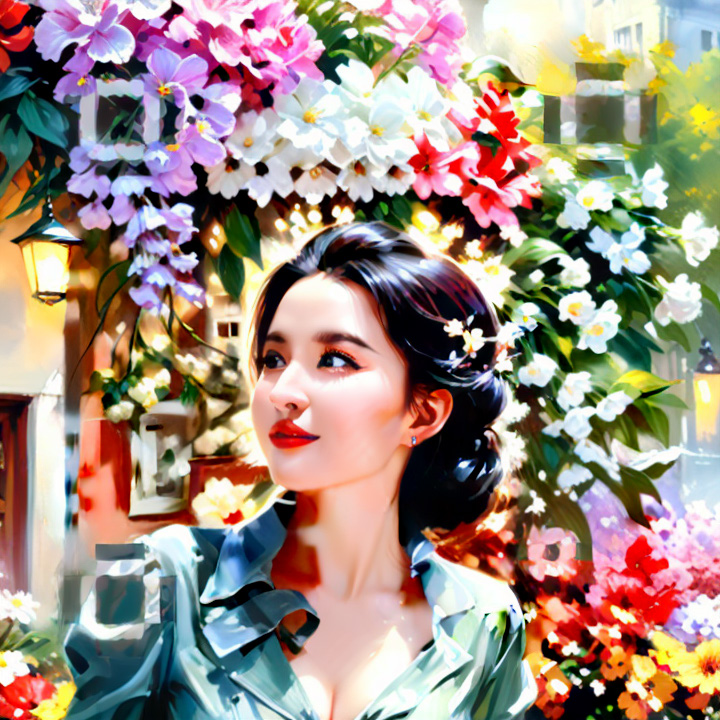}\\

        \includegraphics[width=\picw, height=\picw]{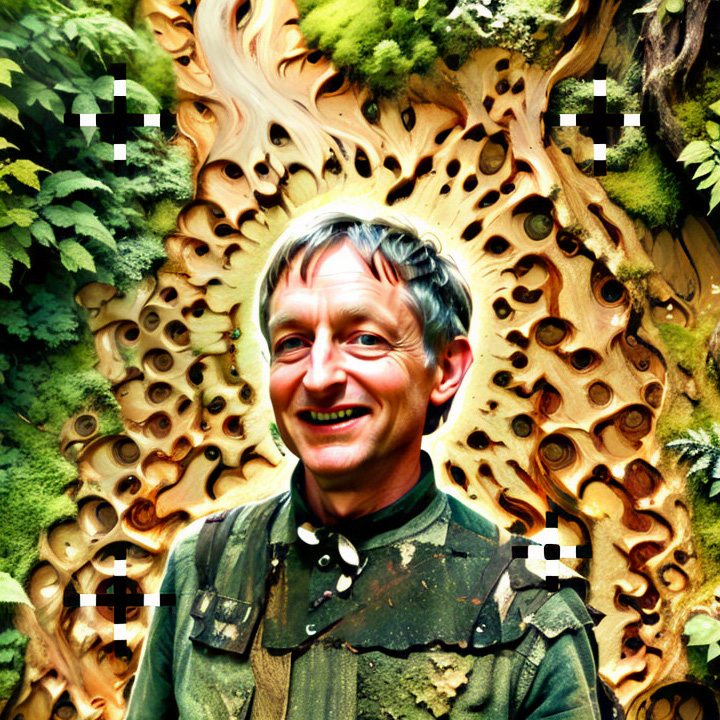} & 
        \includegraphics[width=\picw, height=\picw]{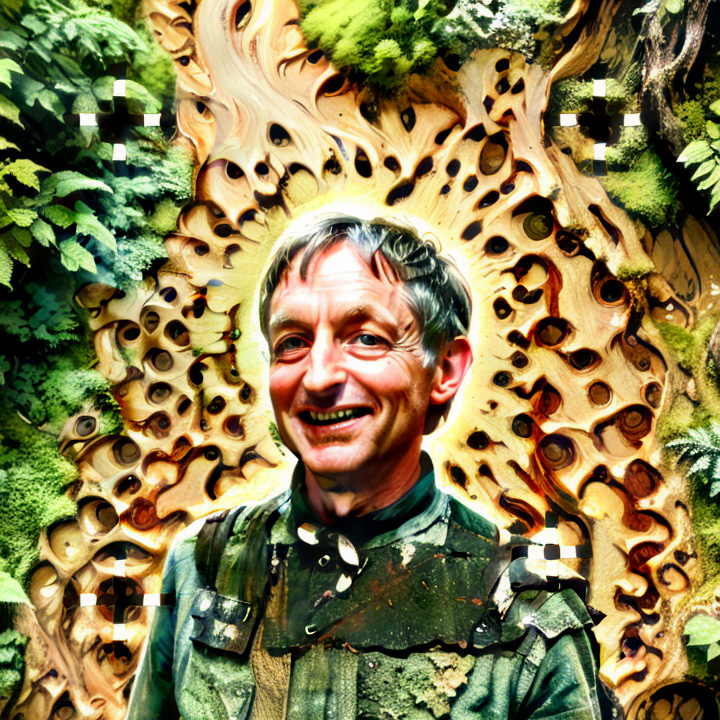} & 
        \includegraphics[width=\picw, height=\picw]{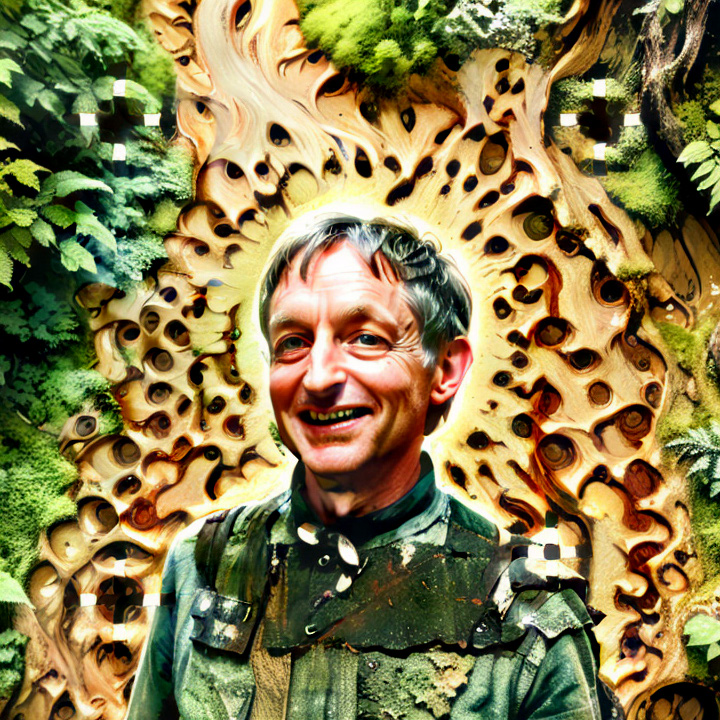} & 
        \includegraphics[width=\picw, height=\picw]{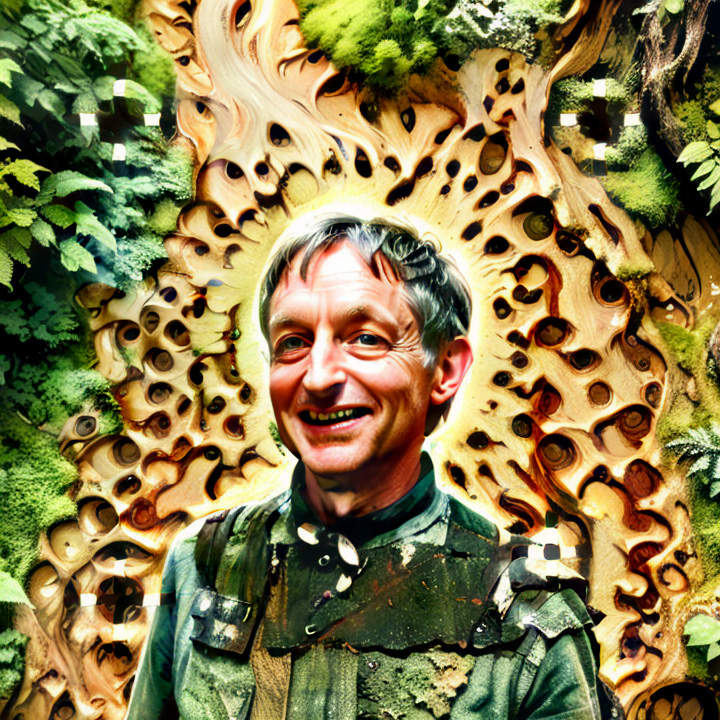}\\

        \includegraphics[width=\picw, height=\picw]{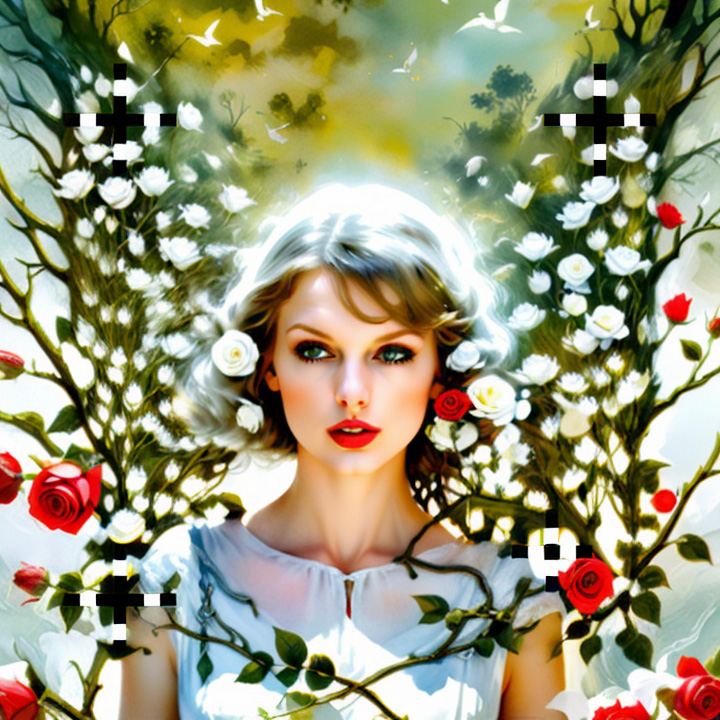} & 
        \includegraphics[width=\picw, height=\picw]{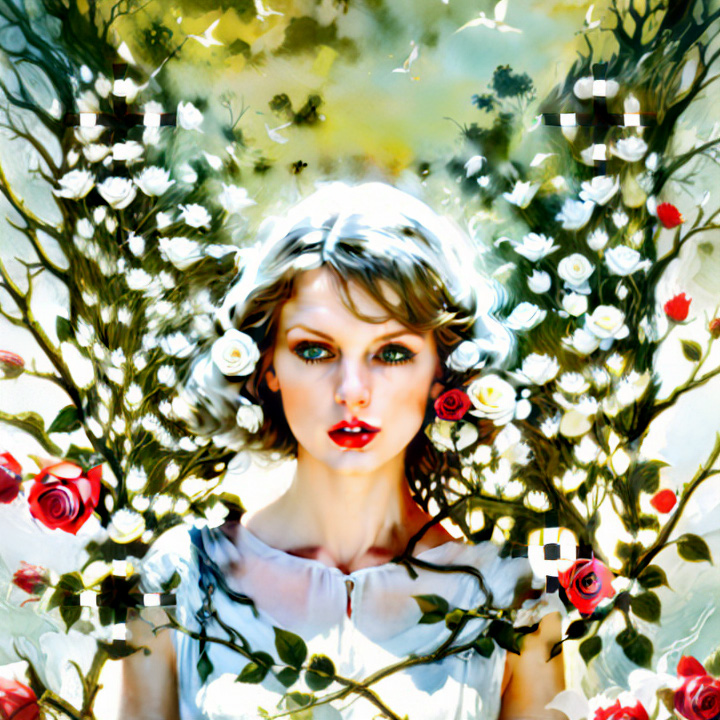} & 
        \includegraphics[width=\picw, height=\picw]{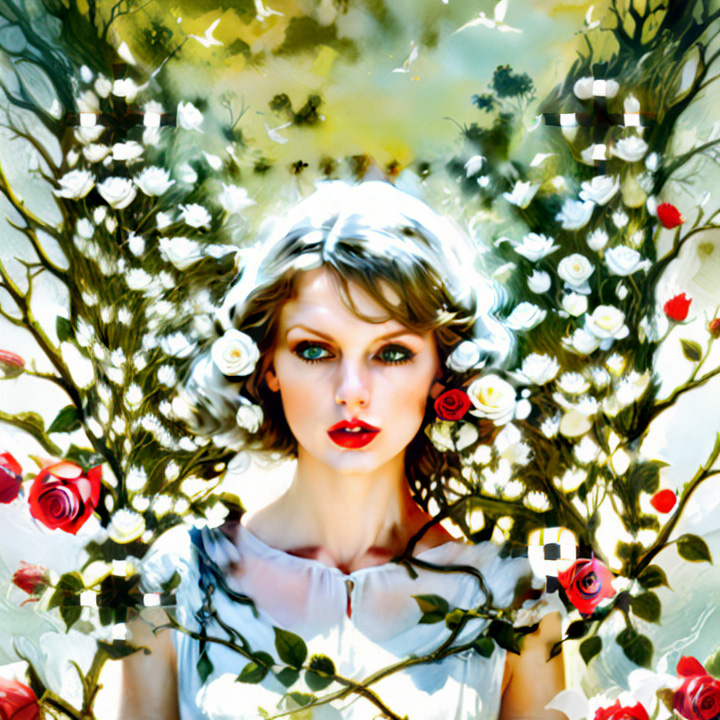} & 
        \includegraphics[width=\picw, height=\picw]{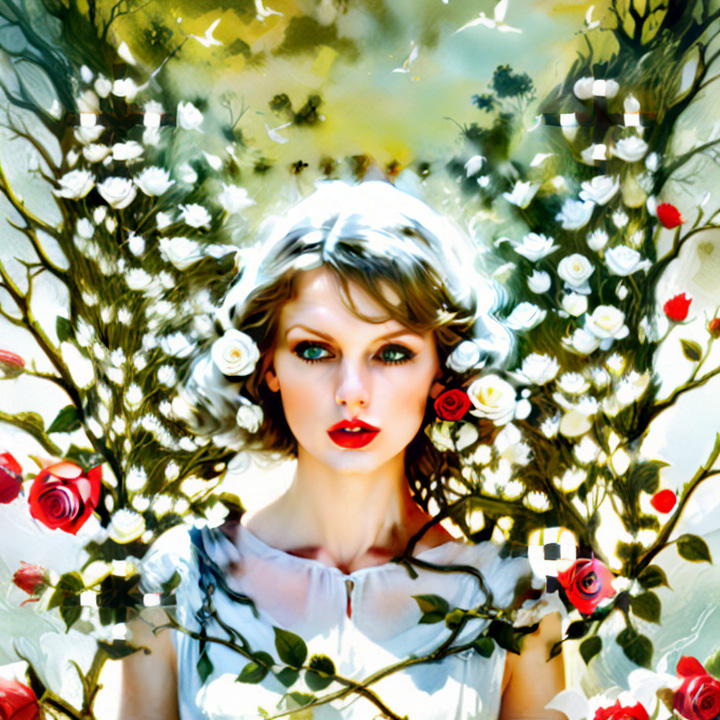}\\
    
        \bottomrule
        \end{tabularx}
    \end{minipage}\hfill
    \begin{minipage}[btp]{.335\linewidth}
        \caption{Bad $I^s$ results caused by conflict between face image and prompt.}
        \centering
        \begin{tabularx}{\linewidth}{>{\centering\arraybackslash}X *{2}{>{\centering\arraybackslash}X}}
    
        \toprule
        Input & $I^s$ \\
        \midrule
    
        \includegraphics[width=\picw, height=\picw]{figures/lyf/lyf-45_1.png} & 
        \includegraphics[width=\picw, height=\picw]{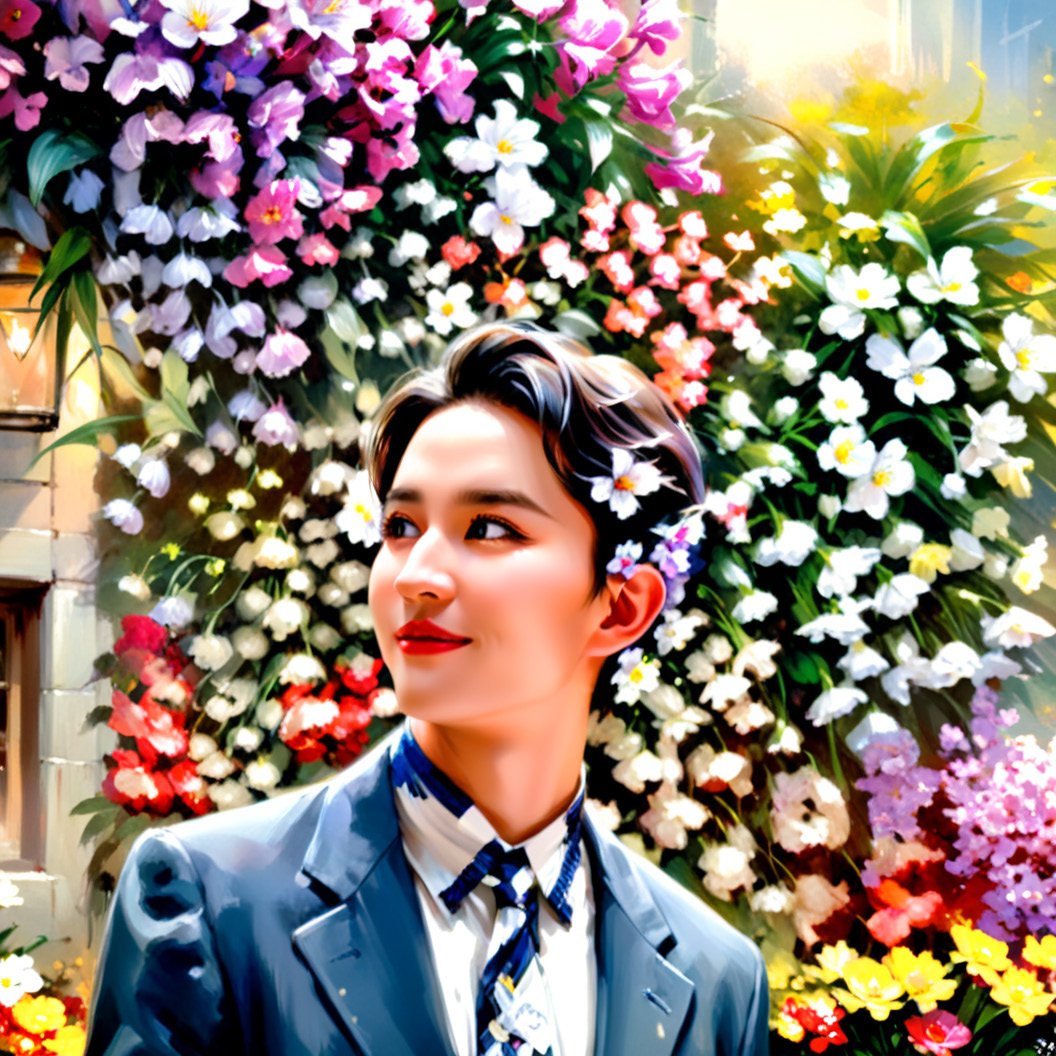} \\

        \includegraphics[width=\picw, height=\picw]{figures/GH/GH.png} & 
        \includegraphics[width=\picw, height=\picw]{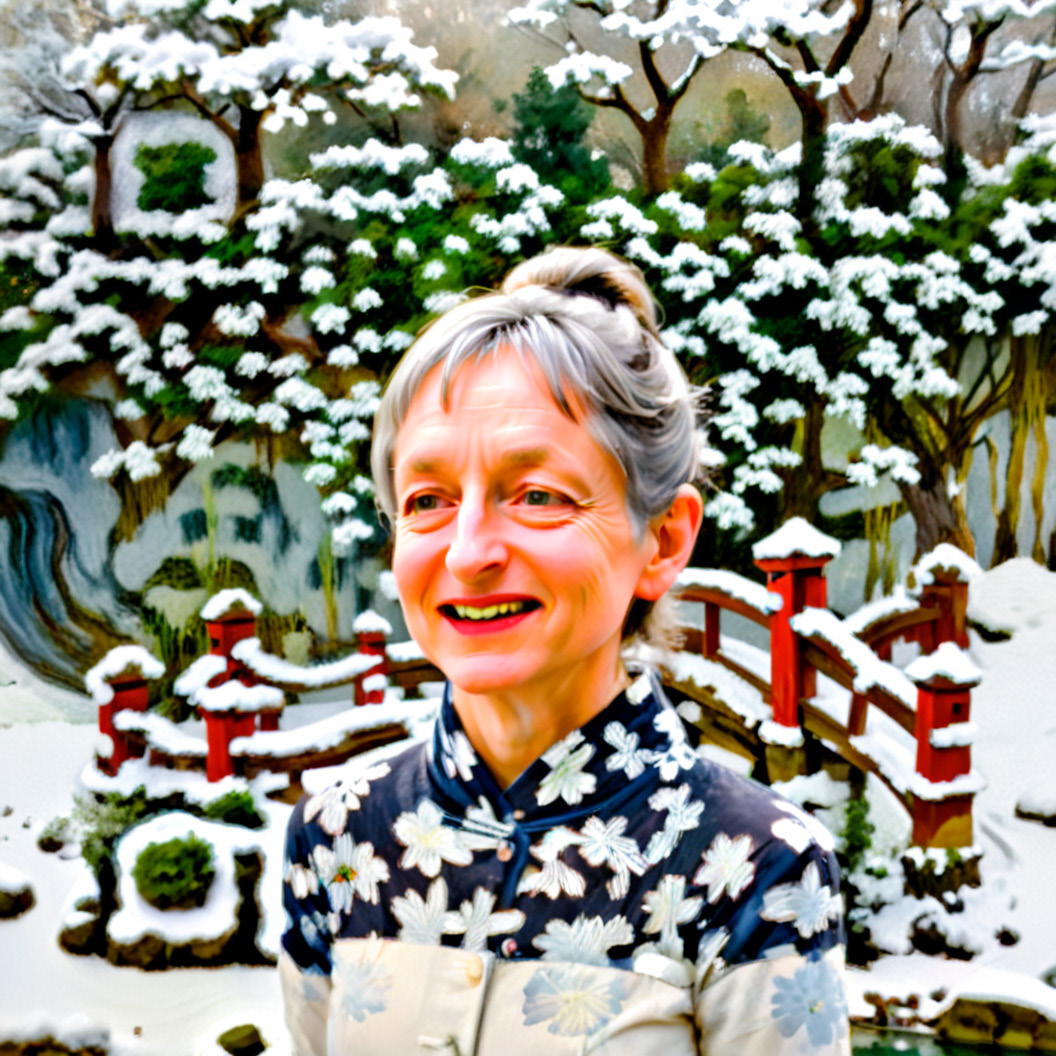} \\
        \includegraphics[width=\picw, height=\picw]{figures/Yann/yann.jpeg} & 
        \includegraphics[width=\picw, height=\picw]{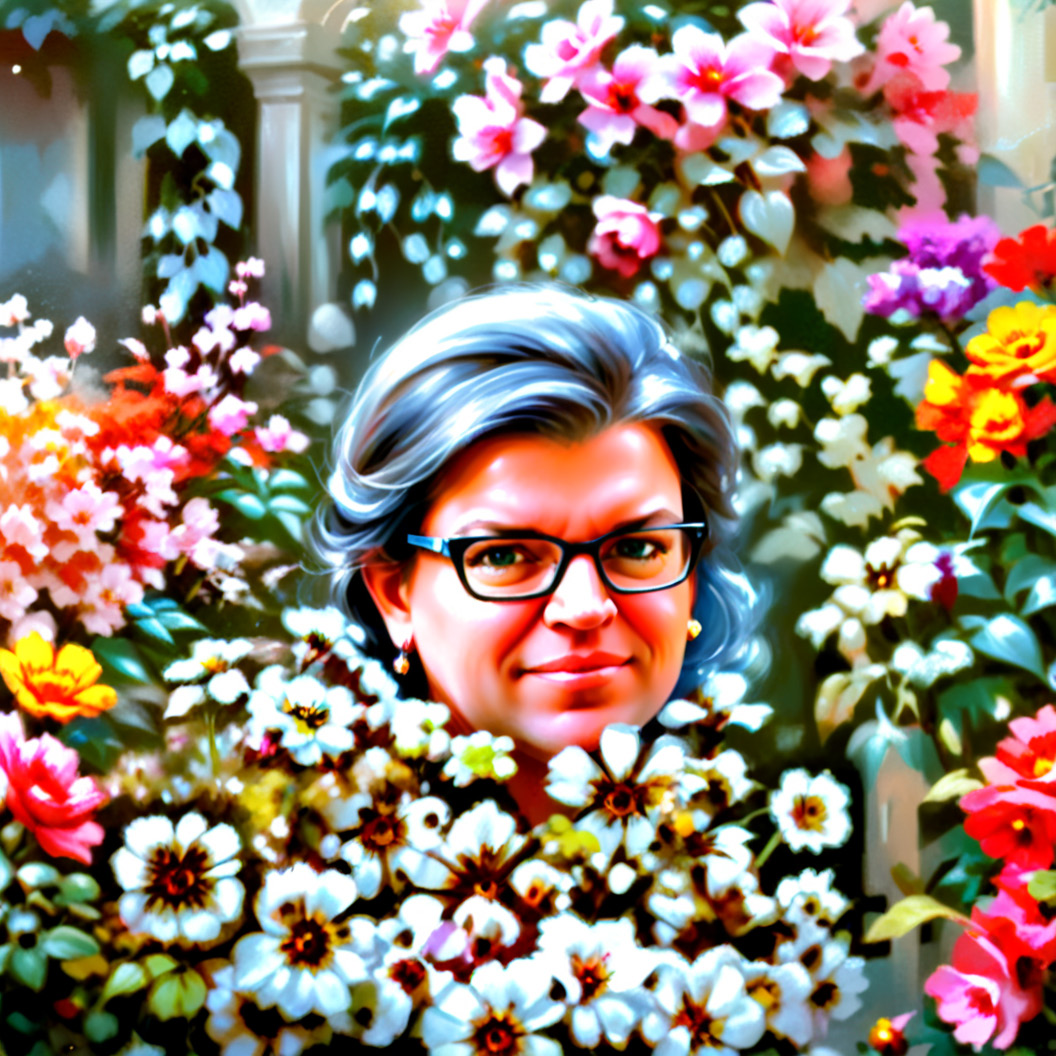} \\
        
        \bottomrule
        \label{fig:bad}
        \end{tabularx}
    \end{minipage}
\end{table*}

\newlength{\suppwidth}
\setlength{\suppwidth}{0.16\textwidth} 
\begin{table*}[tbp]
    \caption{Visualization of intermediate results during our aesthetic QR code generation pipeline.}
    \label{tab:comparison1}
    \centering
    \begin{tabularx}{\linewidth}{>{\centering\arraybackslash}X *{6}{>{\centering\arraybackslash}X}}

    \toprule
    Input & $I^{g}$ & $I_b$ & $I^s$ & $\hat{I^s}$ & $I^o$ \\ \midrule
    un-scannable & un-scannable & scannable & un-scannable & un-scannable & scannable \\
    \midrule

    \includegraphics[width=\suppwidth, height=\suppwidth]{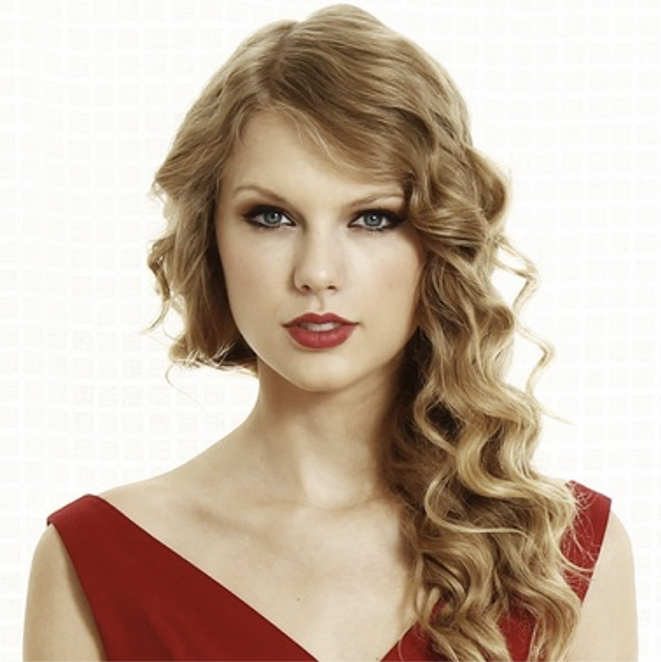} & 
    \includegraphics[width=\suppwidth, height=\suppwidth]{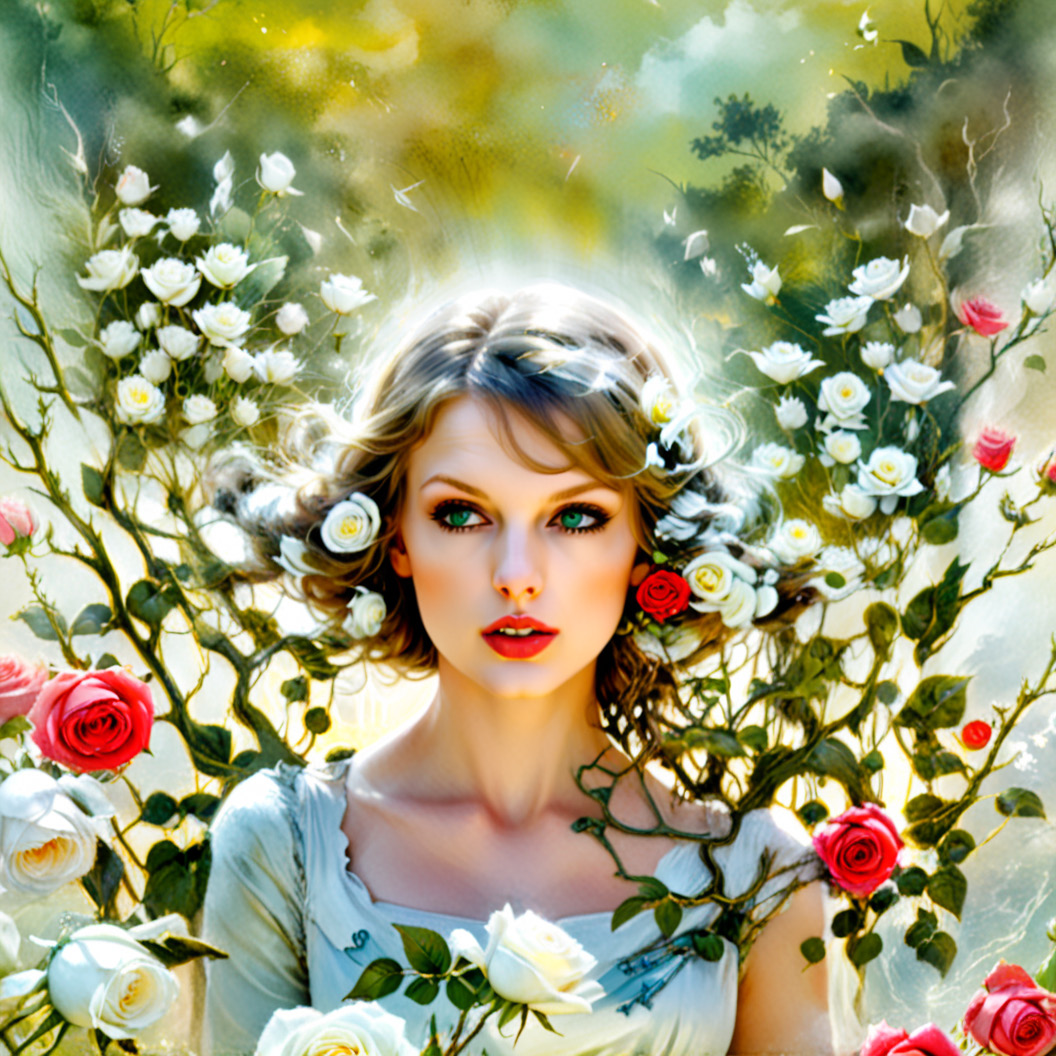} & 
    \includegraphics[width=\suppwidth, height=\suppwidth]{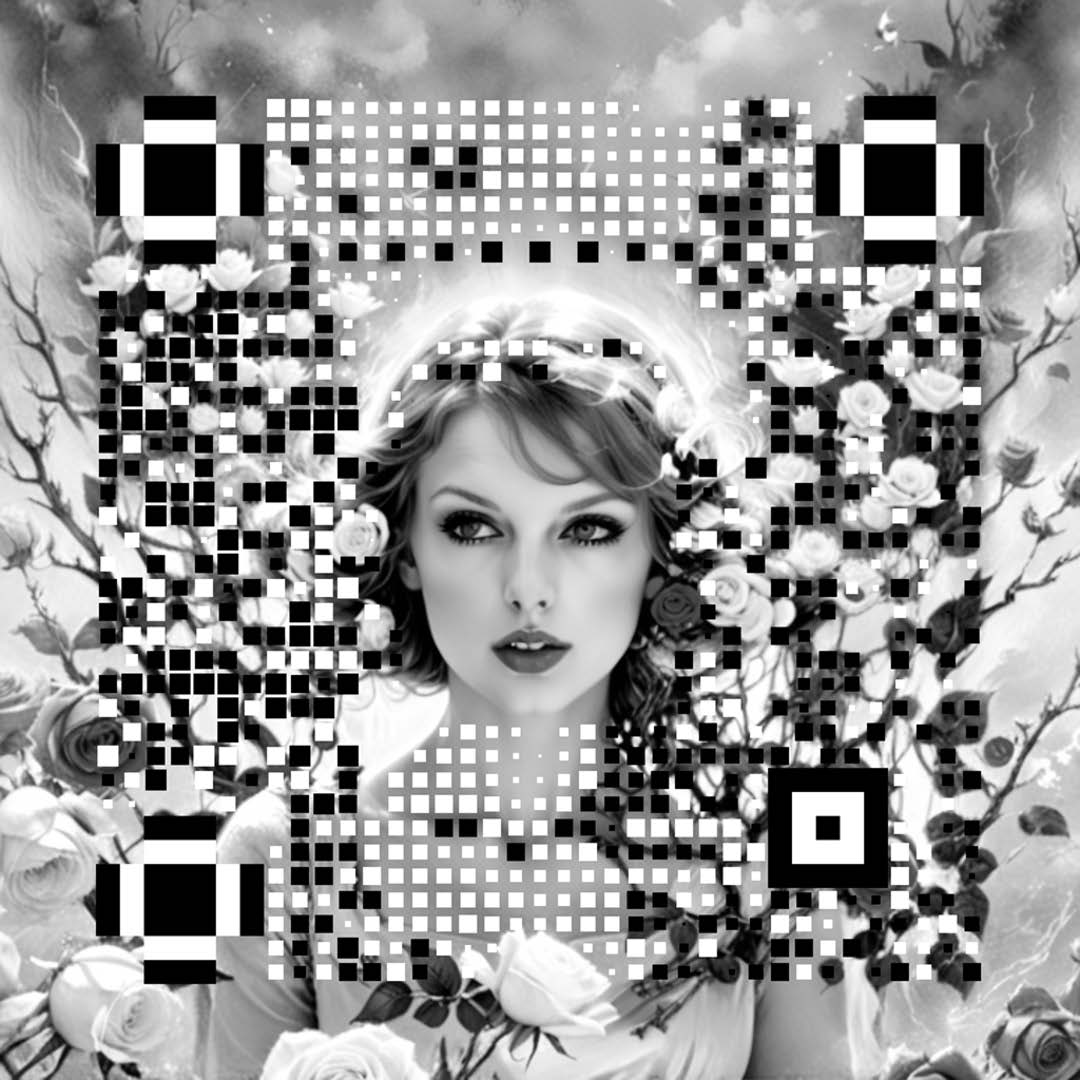} & 
    \includegraphics[width=\suppwidth, height=\suppwidth]{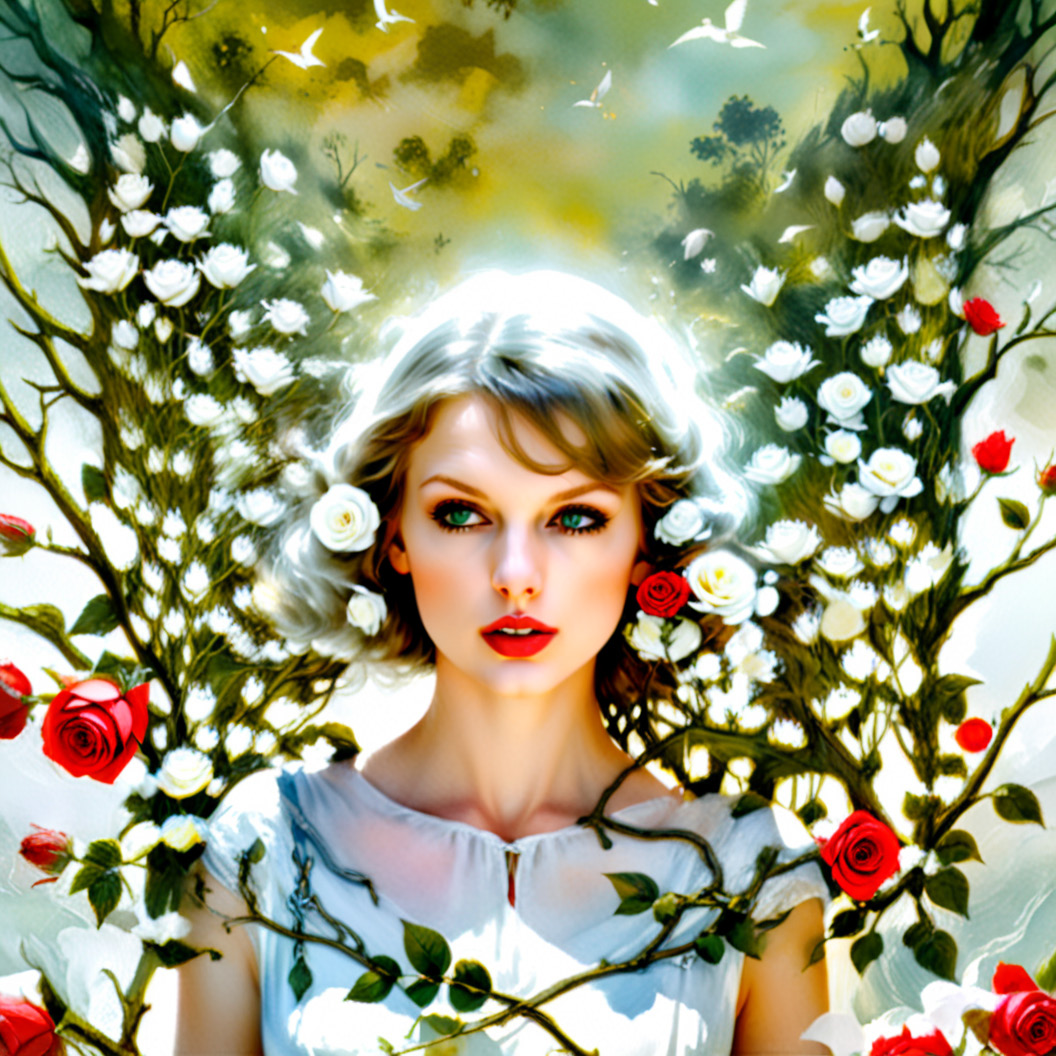} & 
    \includegraphics[width=\suppwidth, height=\suppwidth]{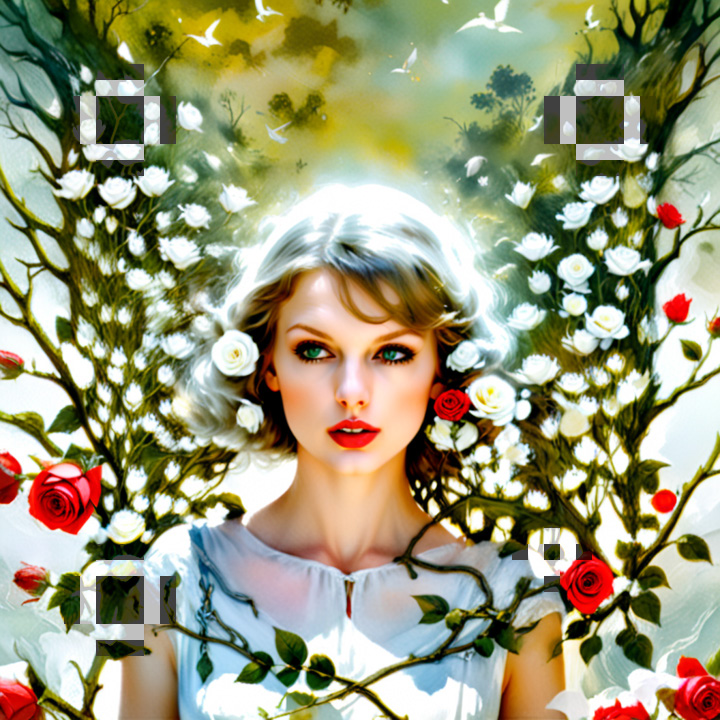} & 
    \includegraphics[width=\suppwidth, height=\suppwidth]{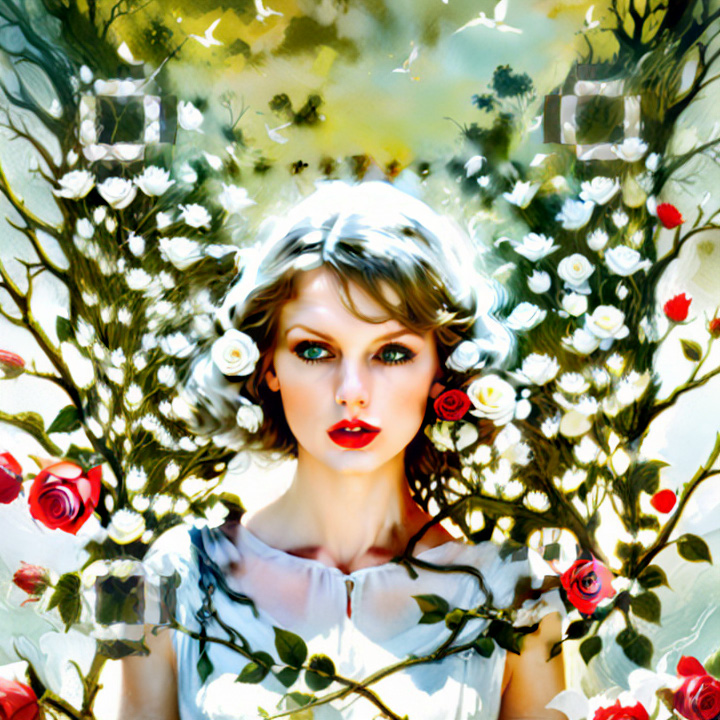} \\

    \includegraphics[width=\suppwidth, height=\suppwidth]{figures/GH/GH.png} & 
    \includegraphics[width=\suppwidth, height=\suppwidth]{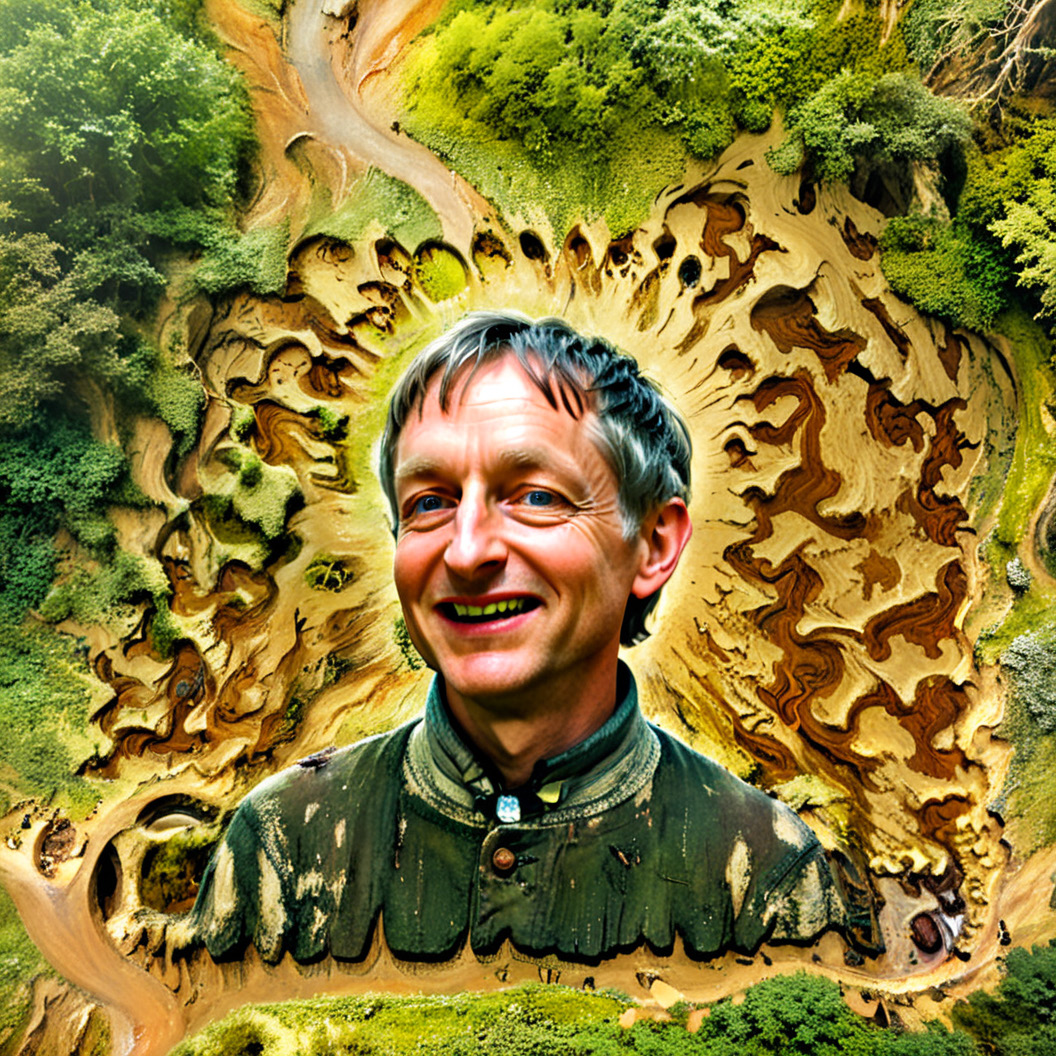} & 
    \includegraphics[width=\suppwidth, height=\suppwidth]{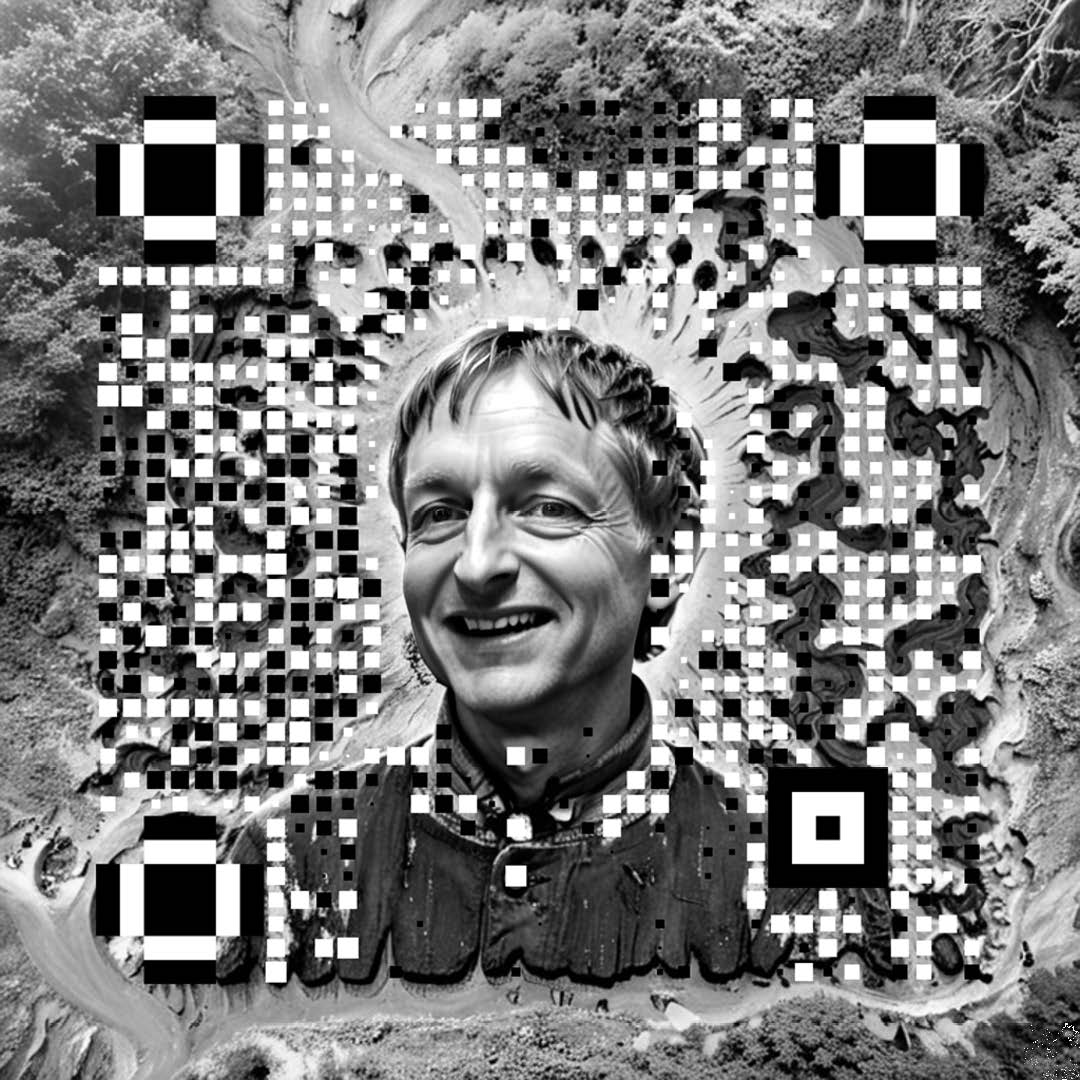} & 
    \includegraphics[width=\suppwidth, height=\suppwidth]{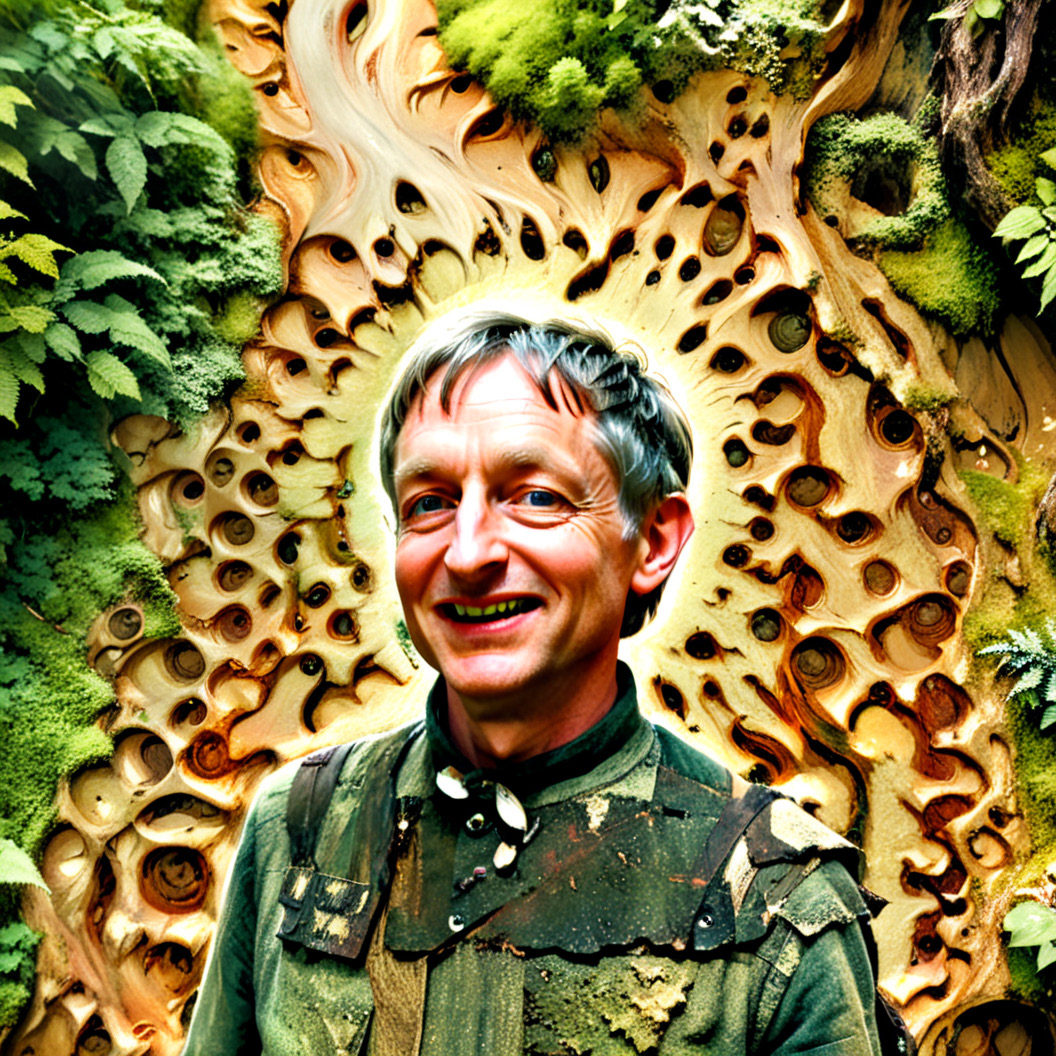} & 
    \includegraphics[width=\suppwidth, height=\suppwidth]{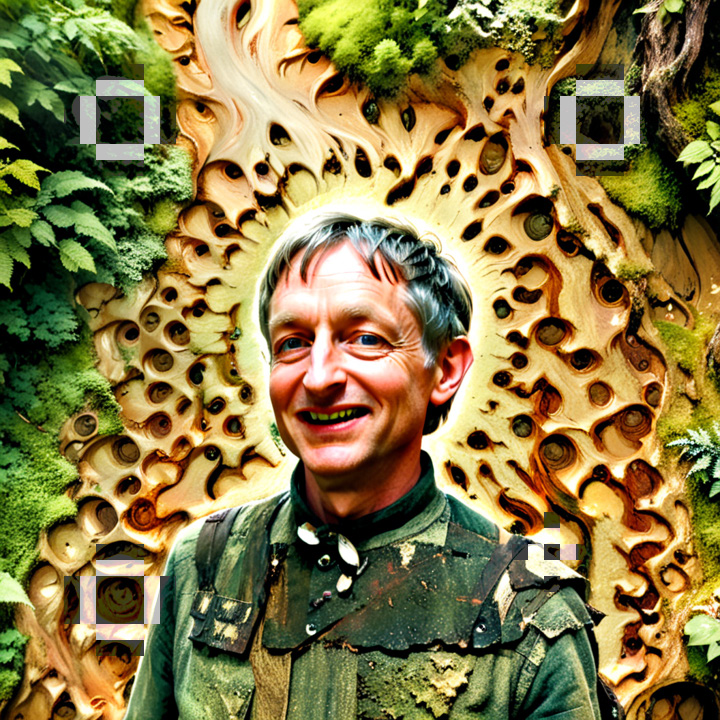} & 
    \includegraphics[width=\suppwidth, height=\suppwidth]{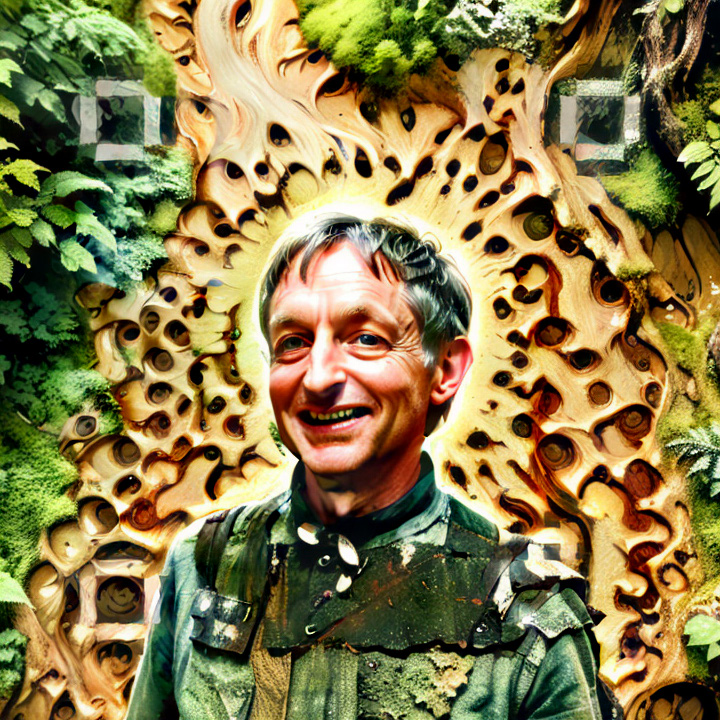} \\

    \includegraphics[width=\suppwidth, height=\suppwidth]{figures/lff/feifei.jpg} & 
    \includegraphics[width=\suppwidth, height=\suppwidth]{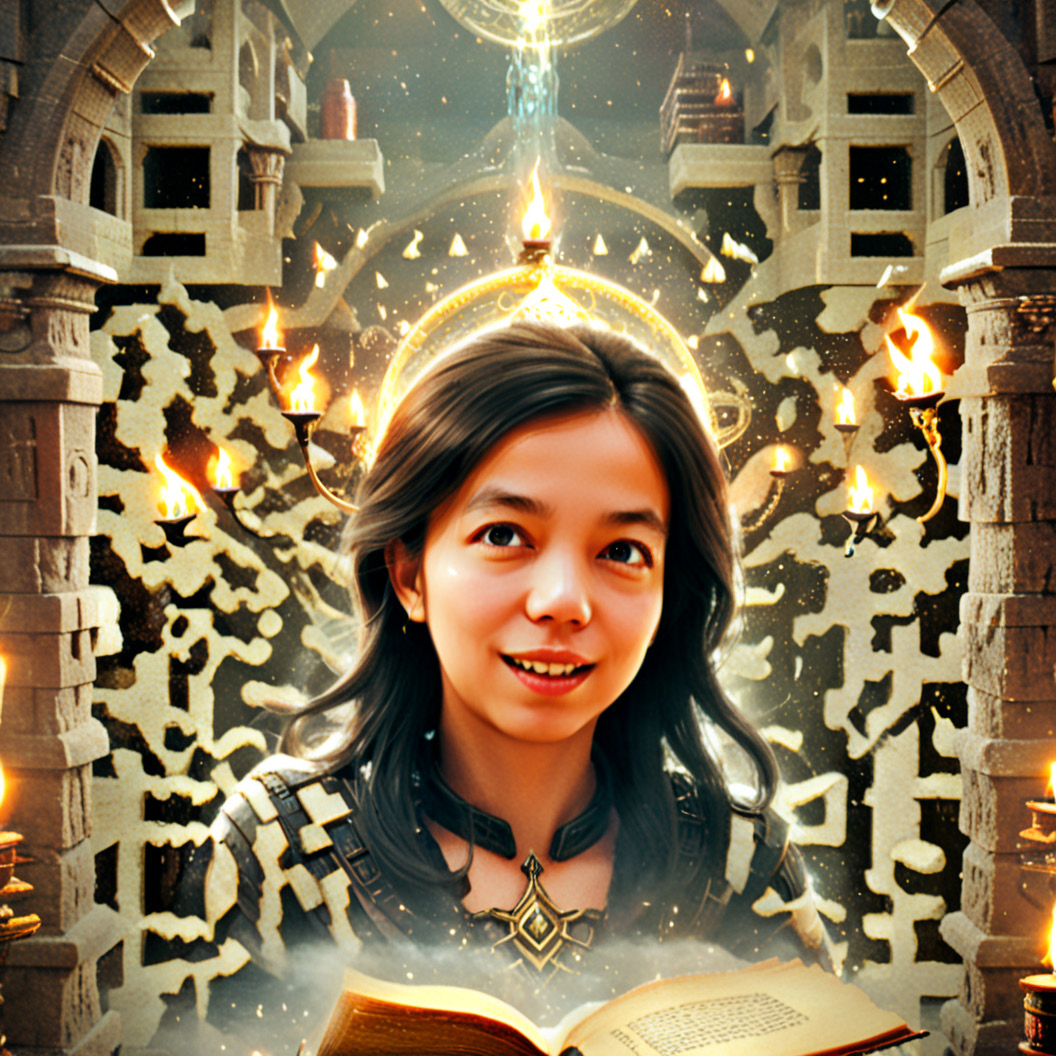} & 
    \includegraphics[width=\suppwidth, height=\suppwidth]{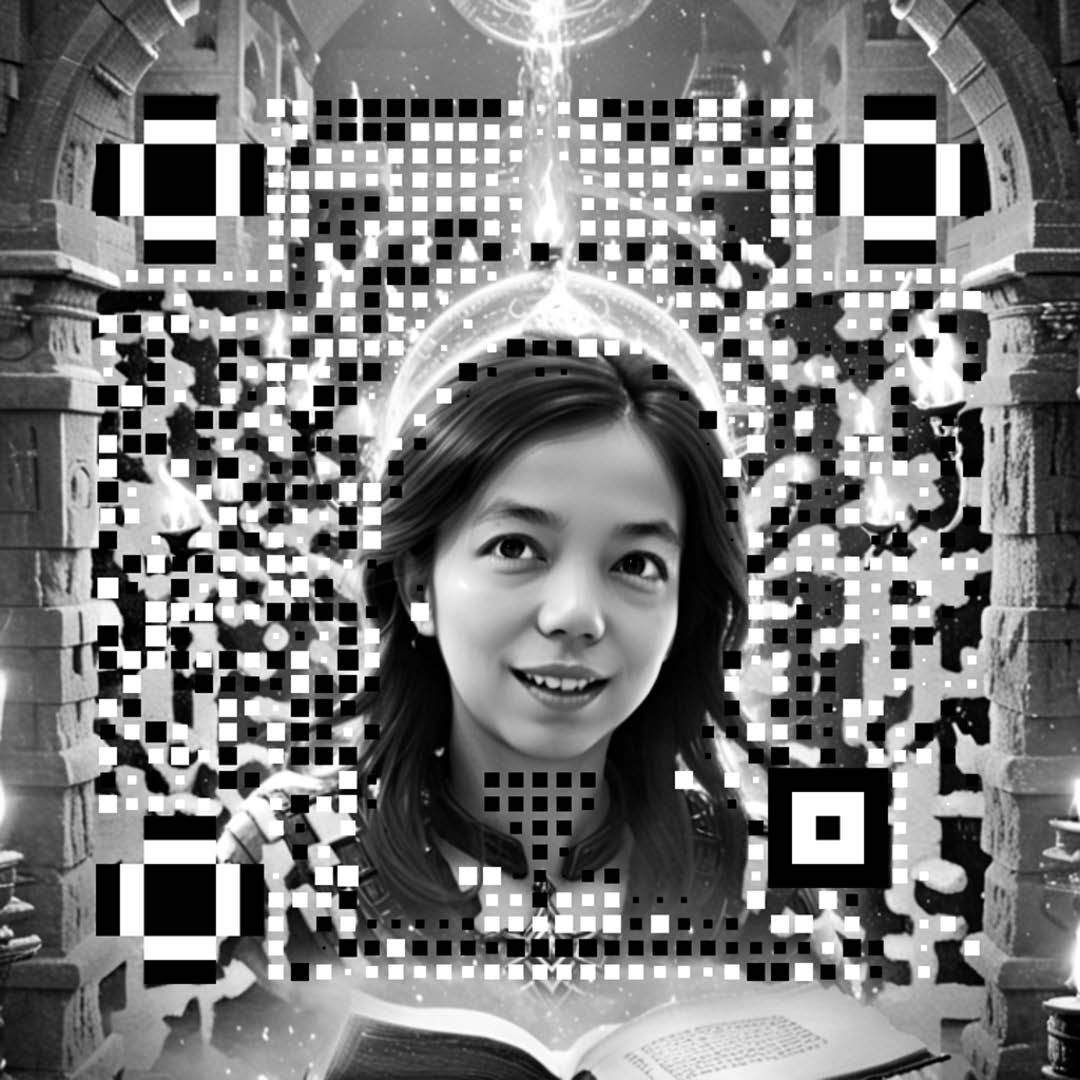} & 
    \includegraphics[width=\suppwidth, height=\suppwidth]{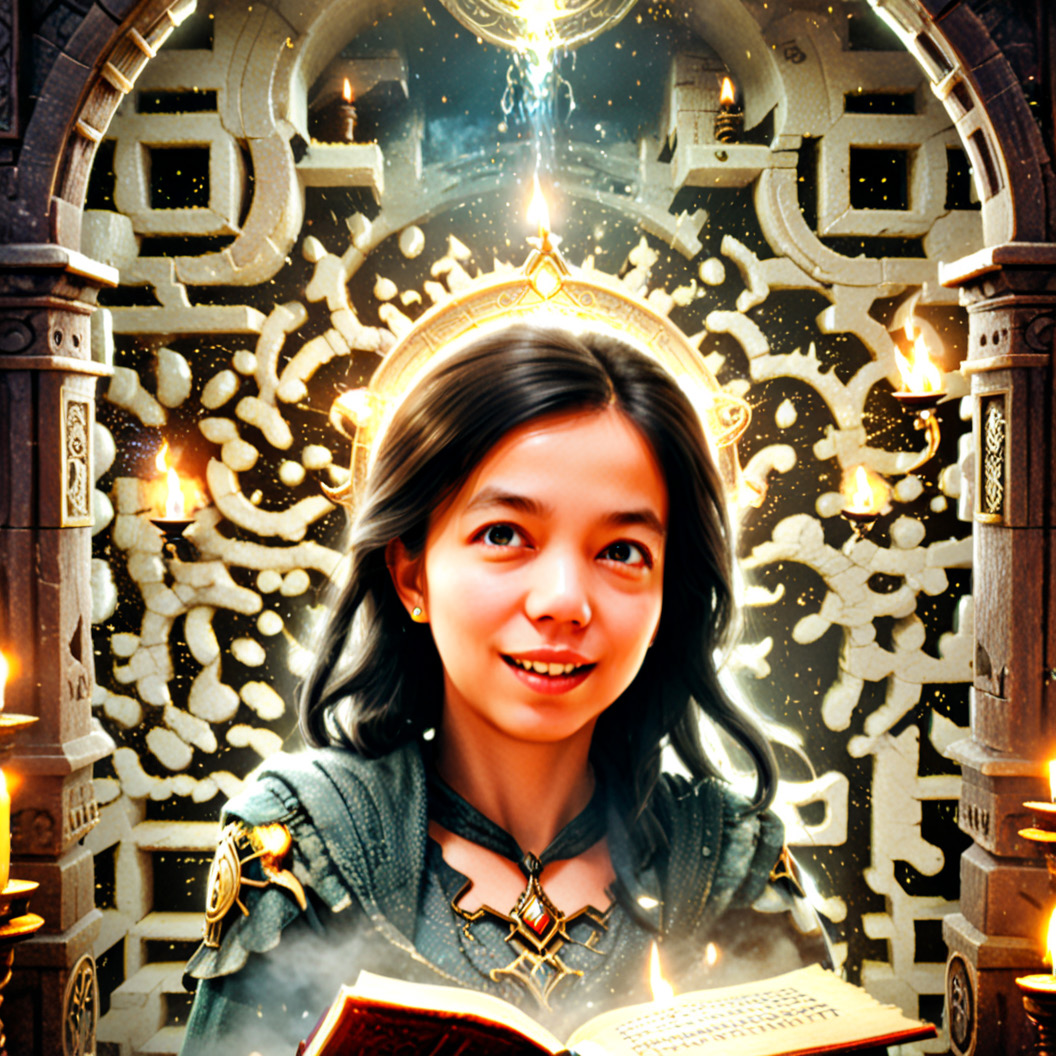} & 
    \includegraphics[width=\suppwidth, height=\suppwidth]{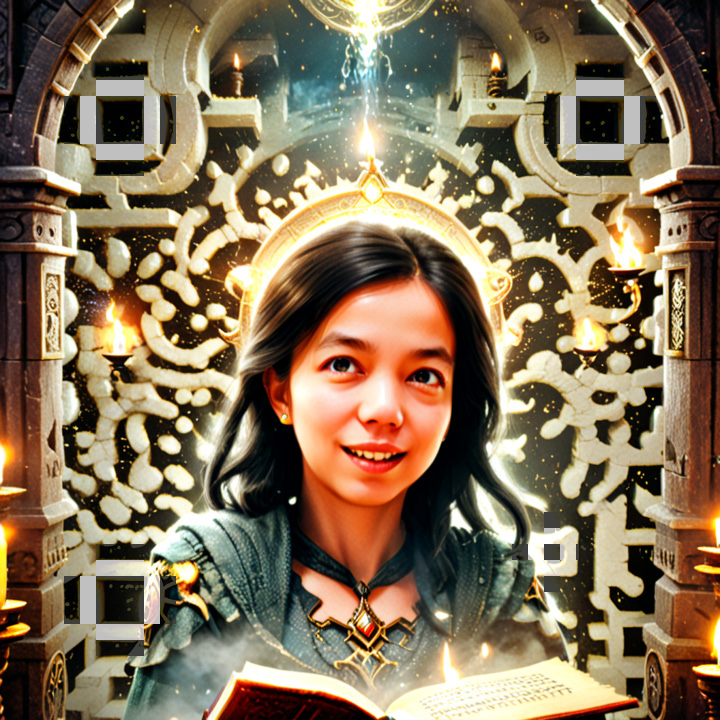} & 
    \includegraphics[width=\suppwidth, height=\suppwidth]{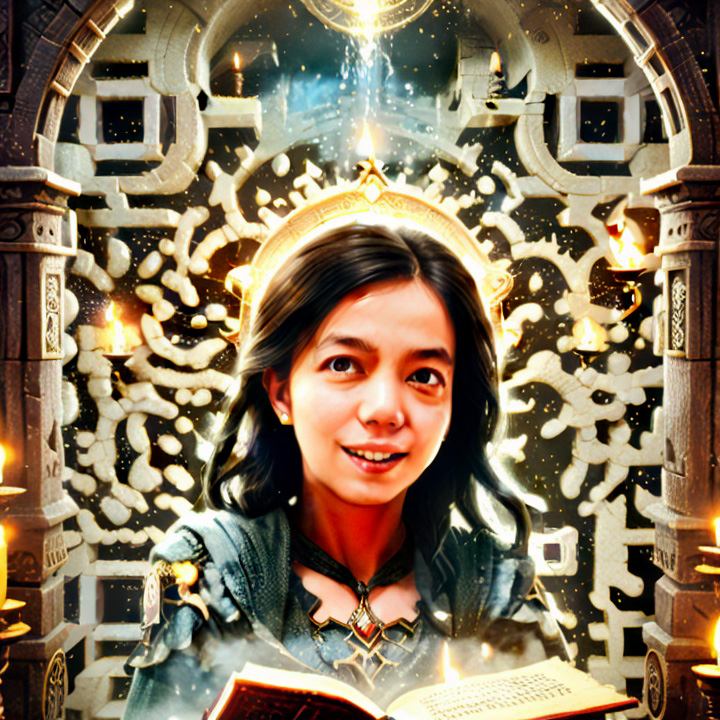} \\

    \includegraphics[width=\suppwidth, height=\suppwidth]{figures/Yann/yann.jpeg} & 
    \includegraphics[width=\suppwidth, height=\suppwidth]{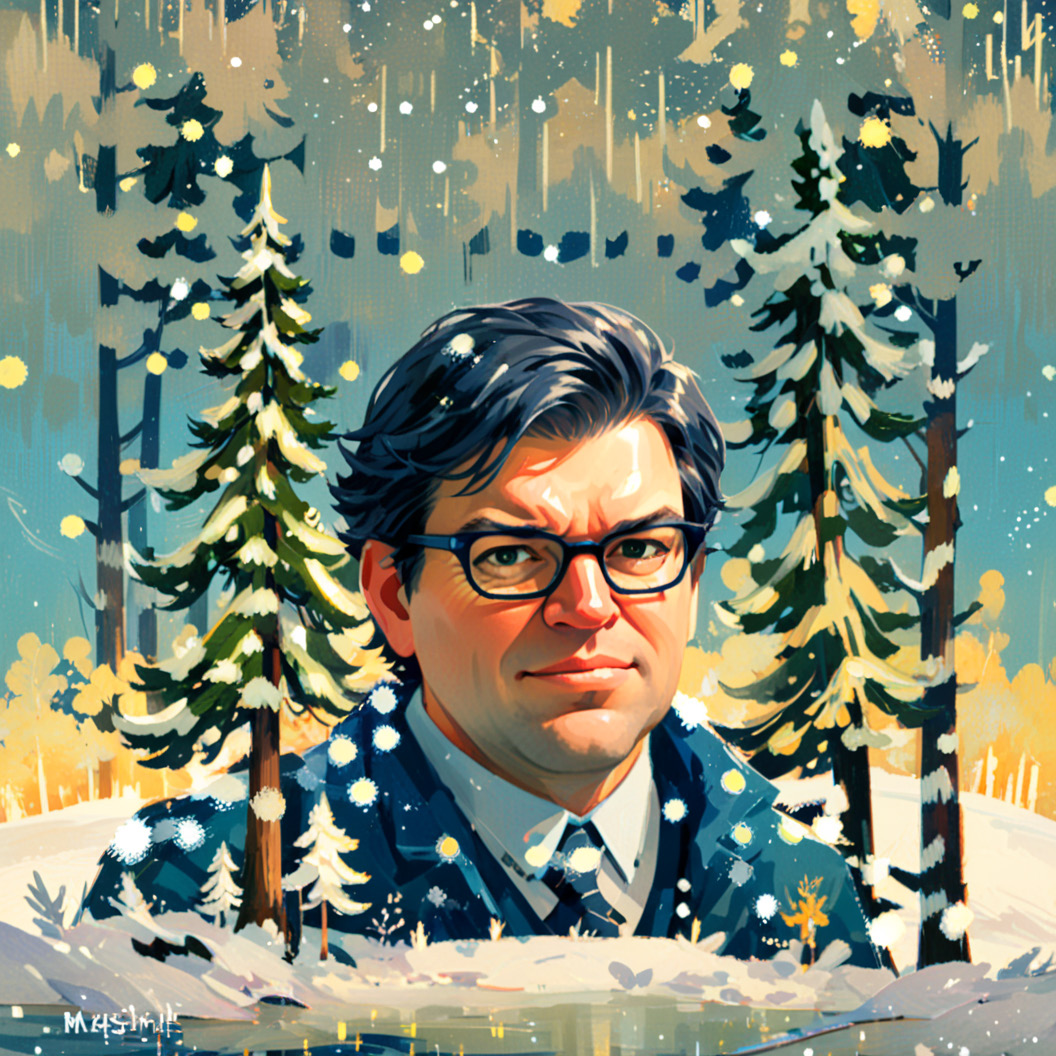} & 
    \includegraphics[width=\suppwidth, height=\suppwidth]{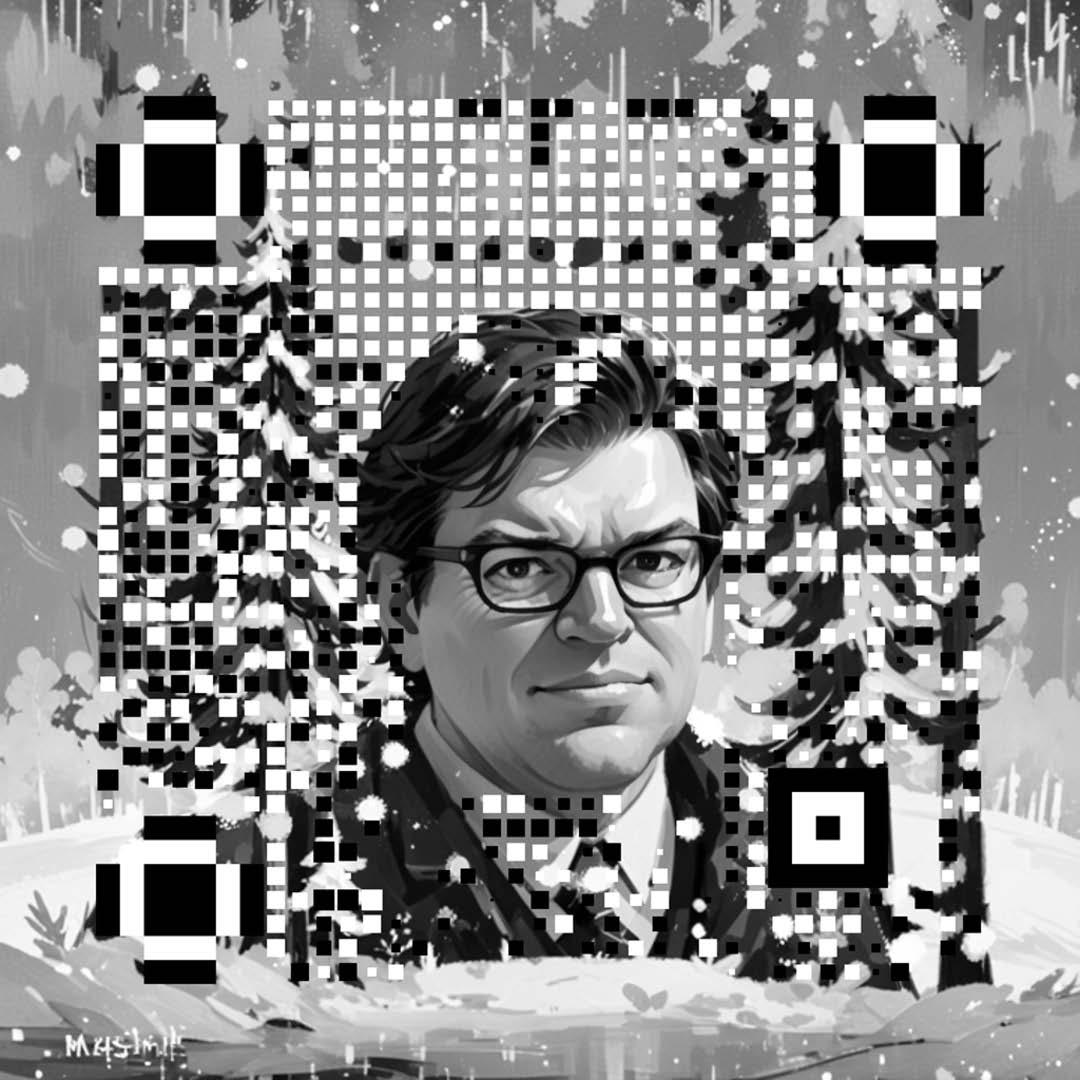} & 
    \includegraphics[width=\suppwidth, height=\suppwidth]{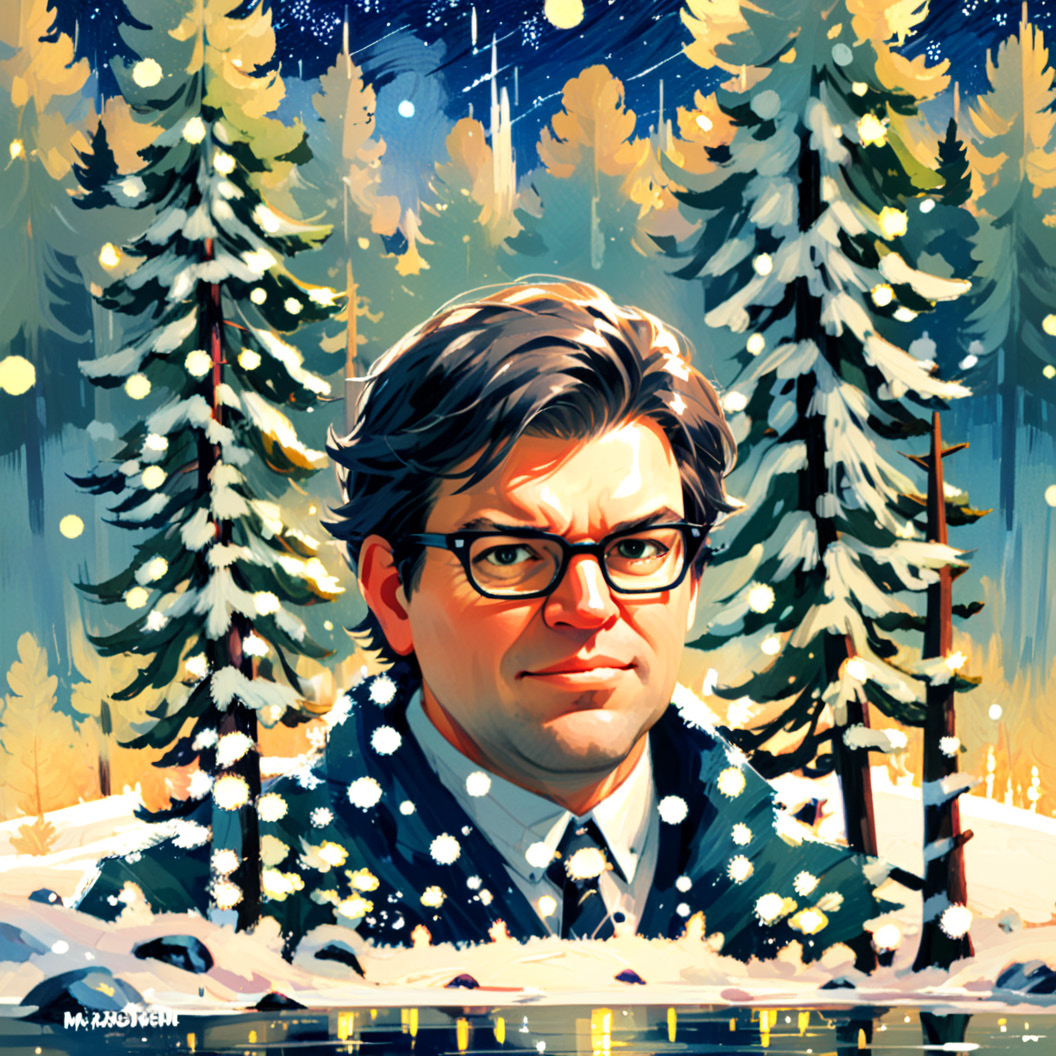} & 
    \includegraphics[width=\suppwidth, height=\suppwidth]{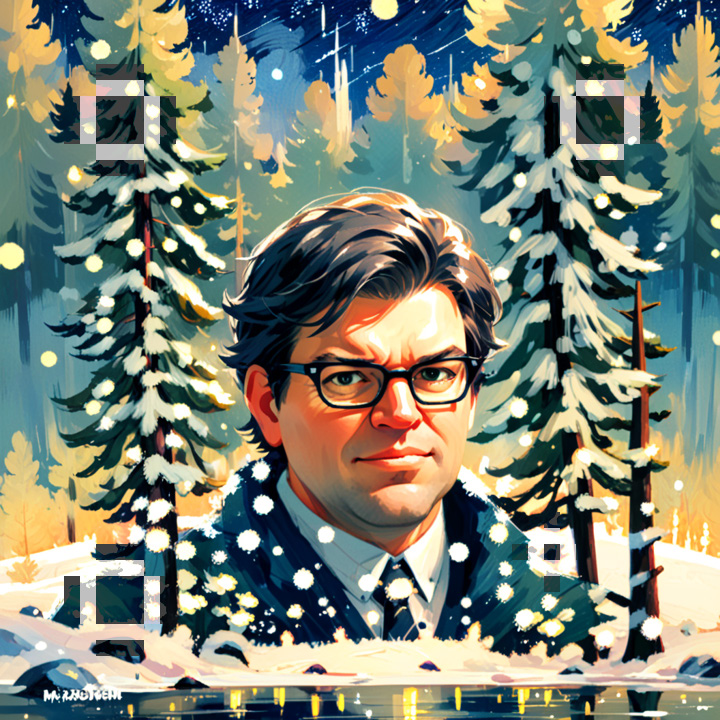} & 
    \includegraphics[width=\suppwidth, height=\suppwidth]{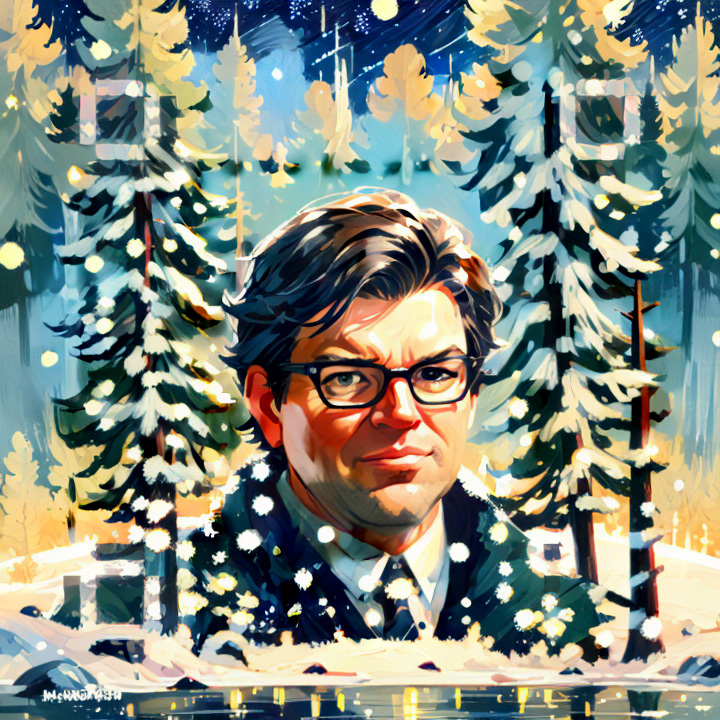} \\

    \includegraphics[width=\suppwidth, height=\suppwidth]{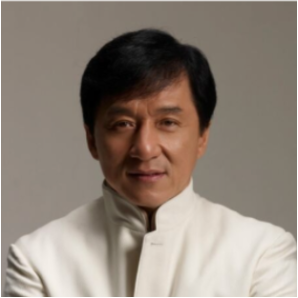} & 
    \includegraphics[width=\suppwidth, height=\suppwidth]{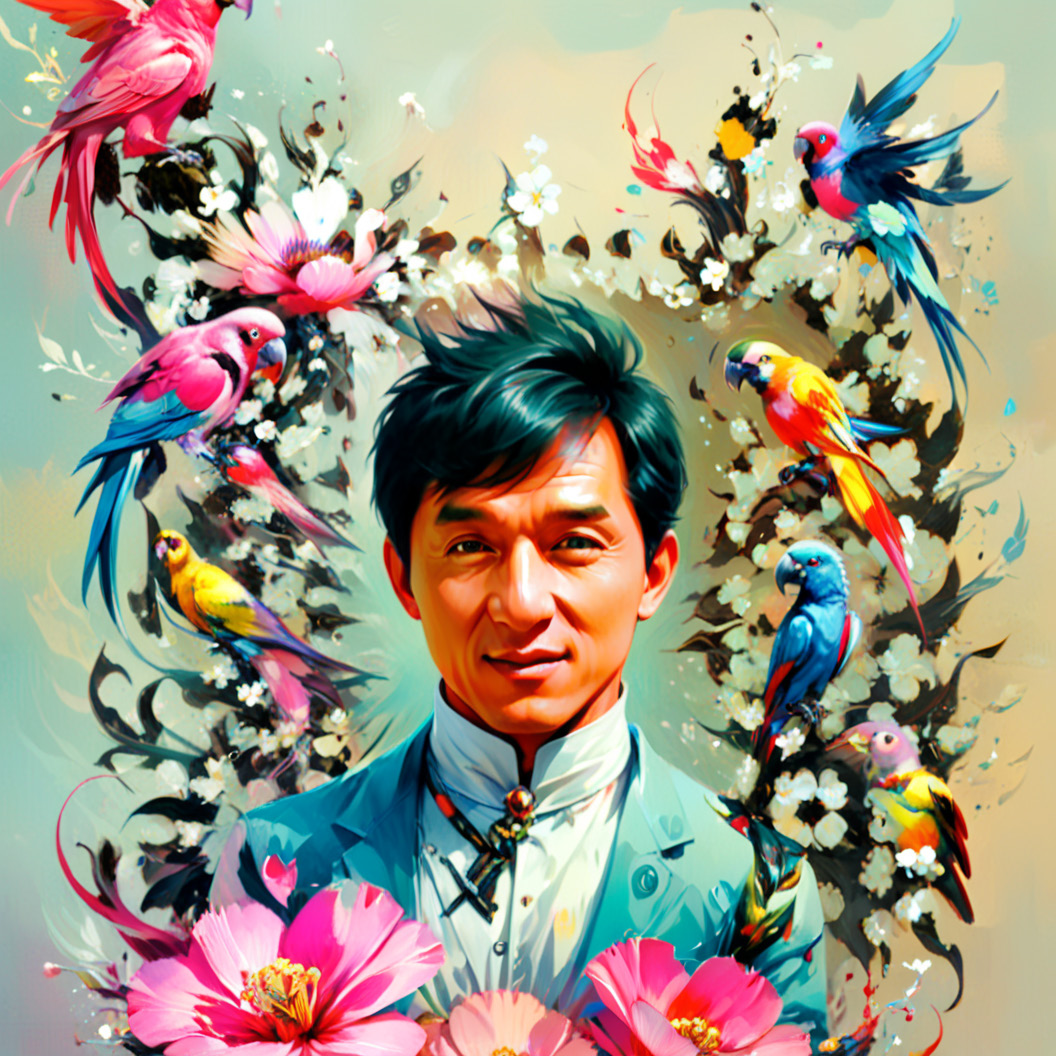} & 
    \includegraphics[width=\suppwidth, height=\suppwidth]{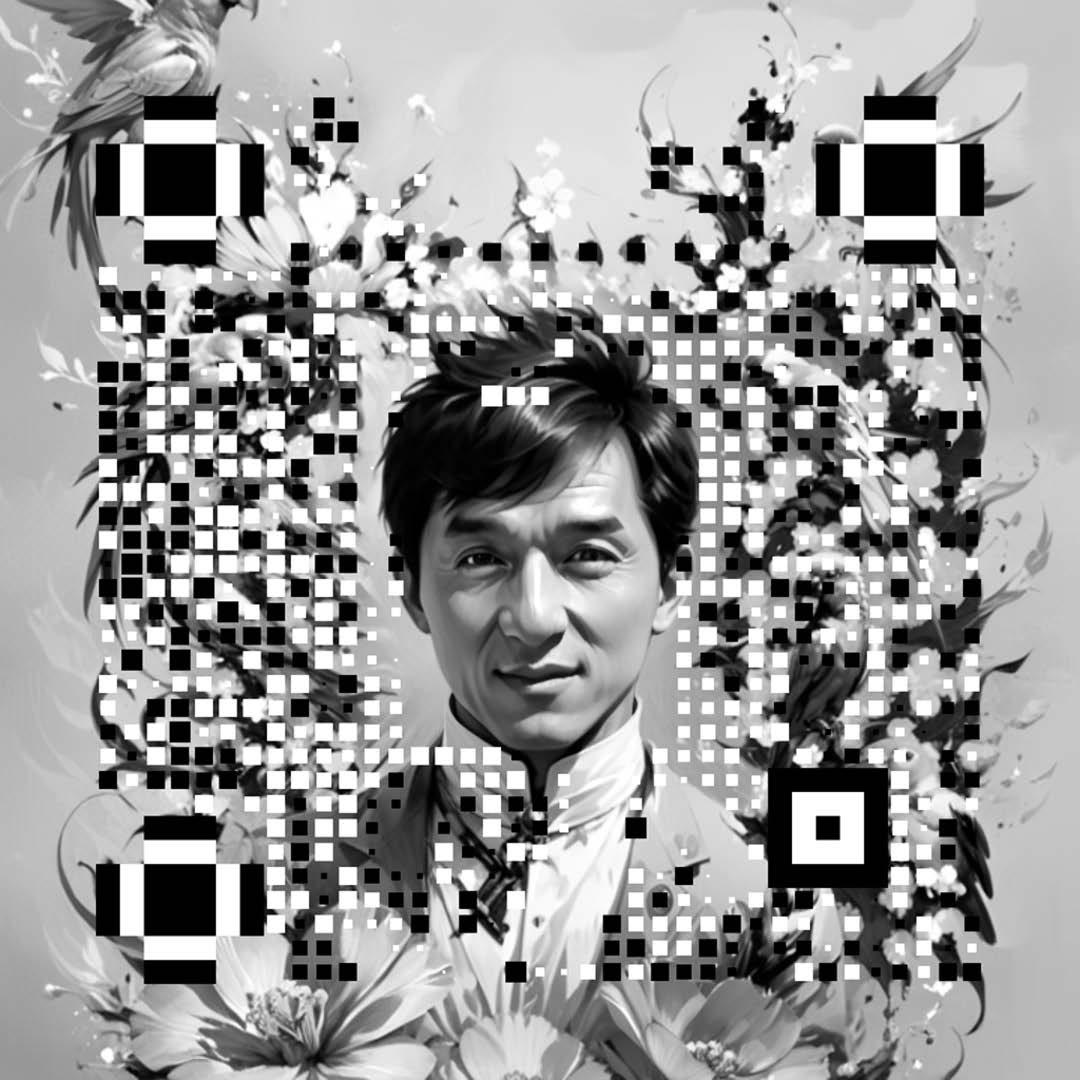} & 
    \includegraphics[width=\suppwidth, height=\suppwidth]{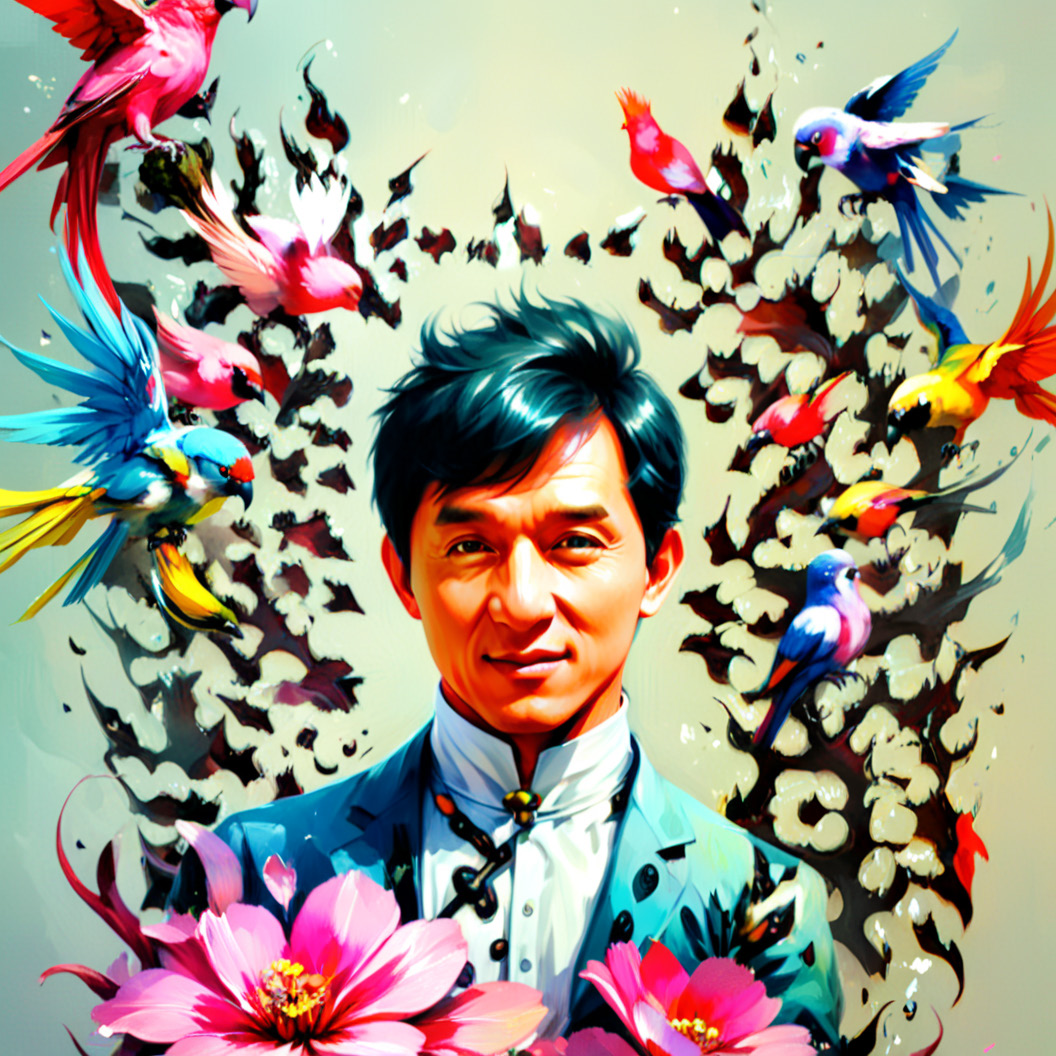} & 
    \includegraphics[width=\suppwidth, height=\suppwidth]{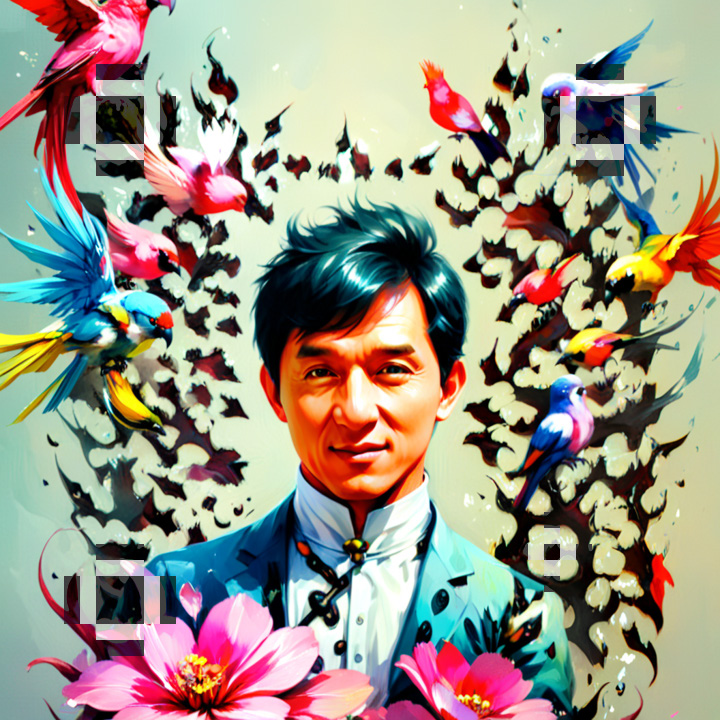} & 
    \includegraphics[width=\suppwidth, height=\suppwidth]{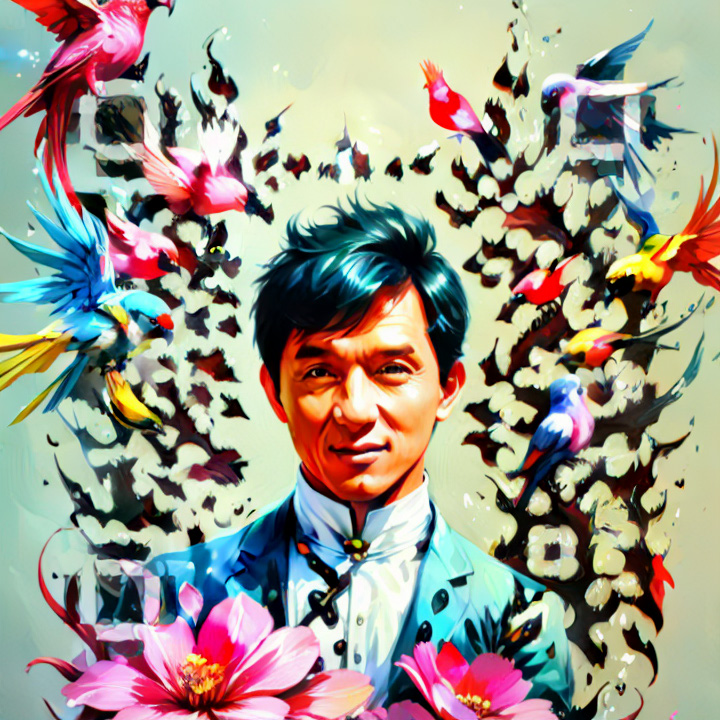} \\
    
    \includegraphics[width=\suppwidth, height=\suppwidth]{figures/lyf/lyf-45_1.png} & 
    \includegraphics[width=\suppwidth, height=\suppwidth]{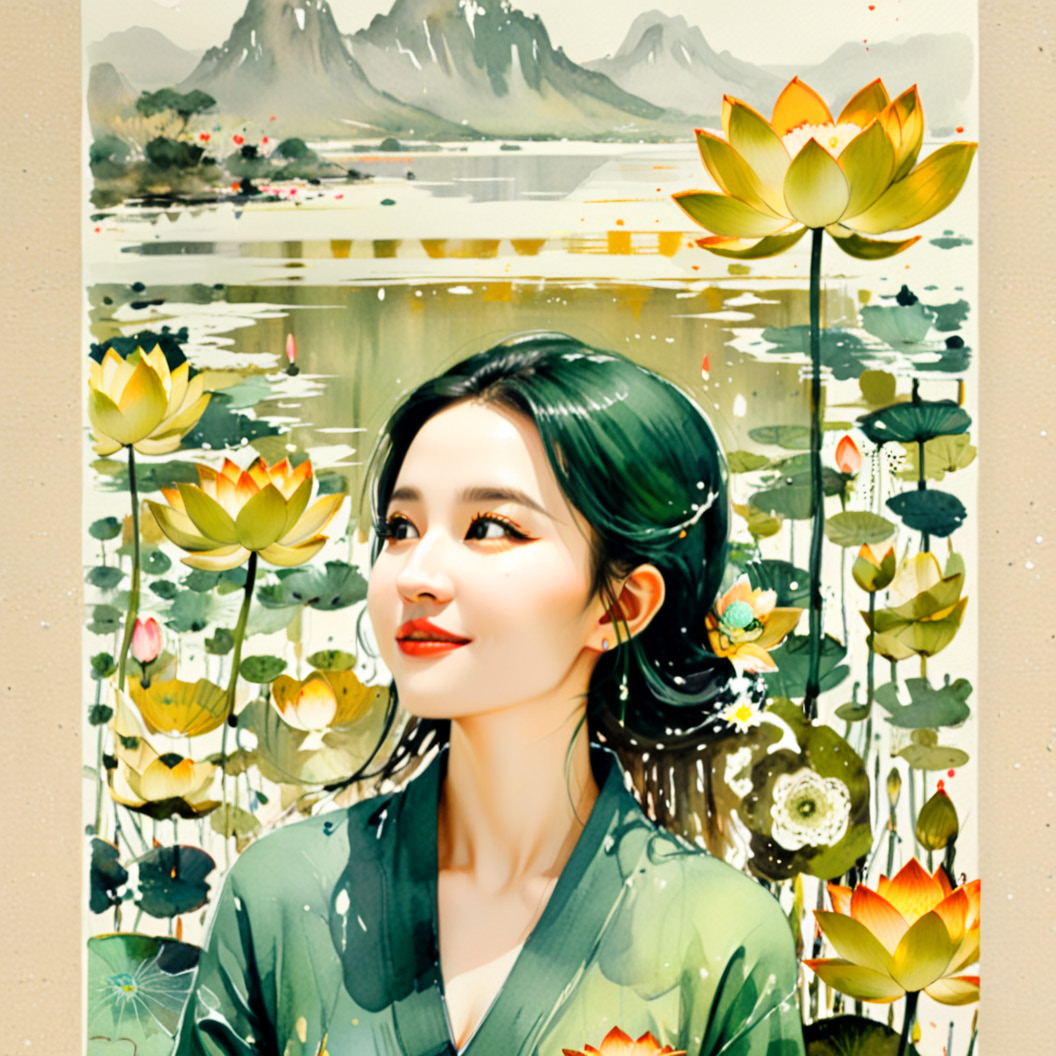} & 
    \includegraphics[width=\suppwidth, height=\suppwidth]{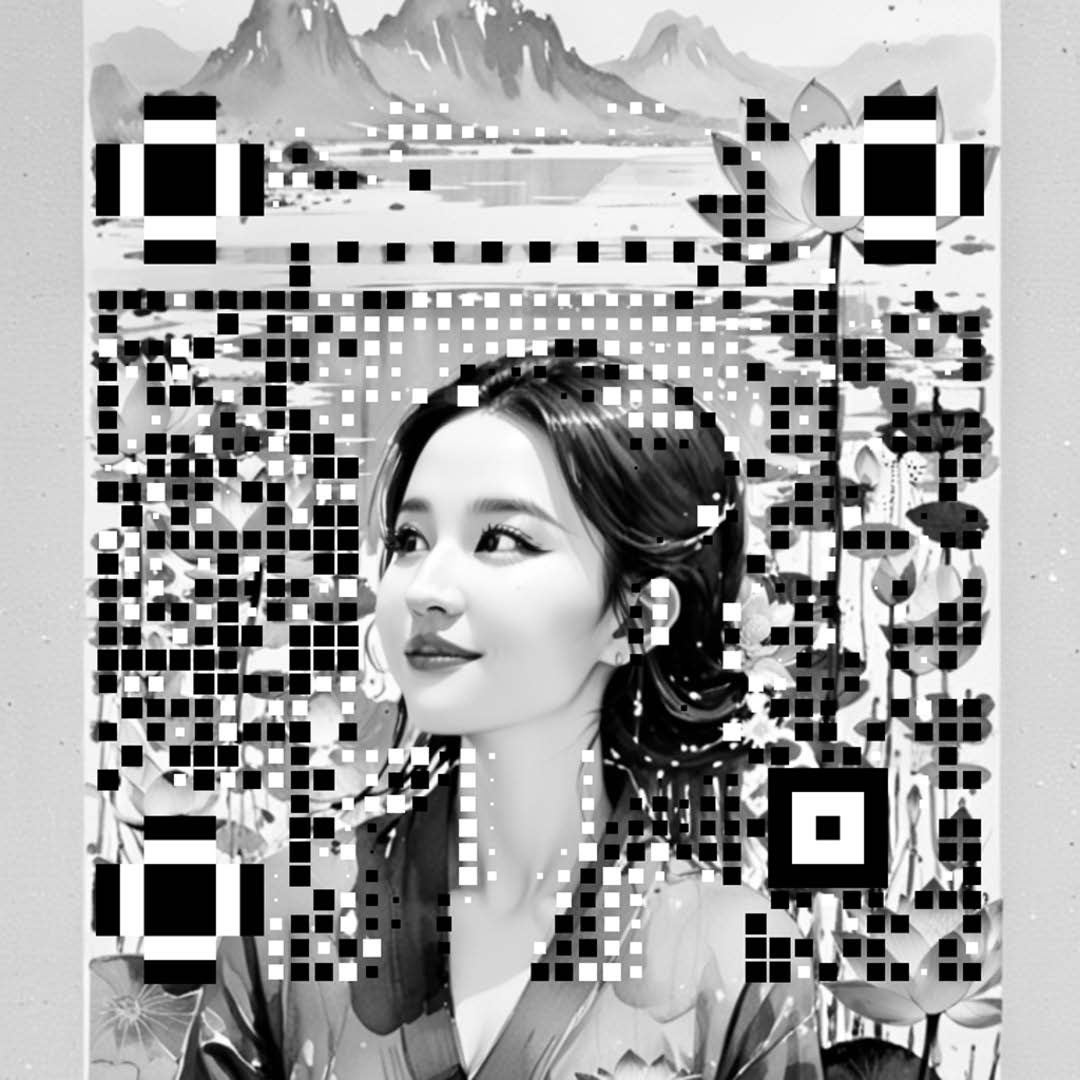} & 
    \includegraphics[width=\suppwidth, height=\suppwidth]{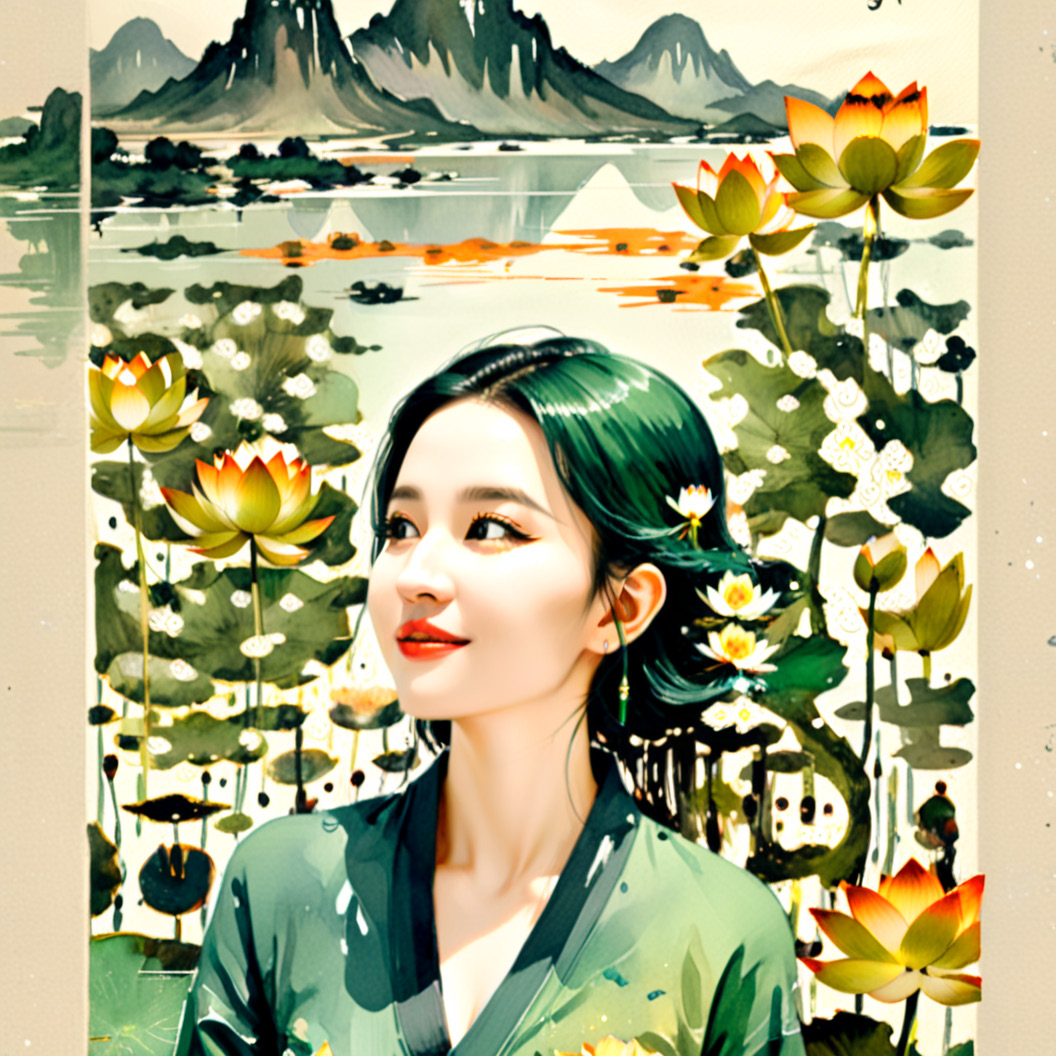} & 
    \includegraphics[width=\suppwidth, height=\suppwidth]{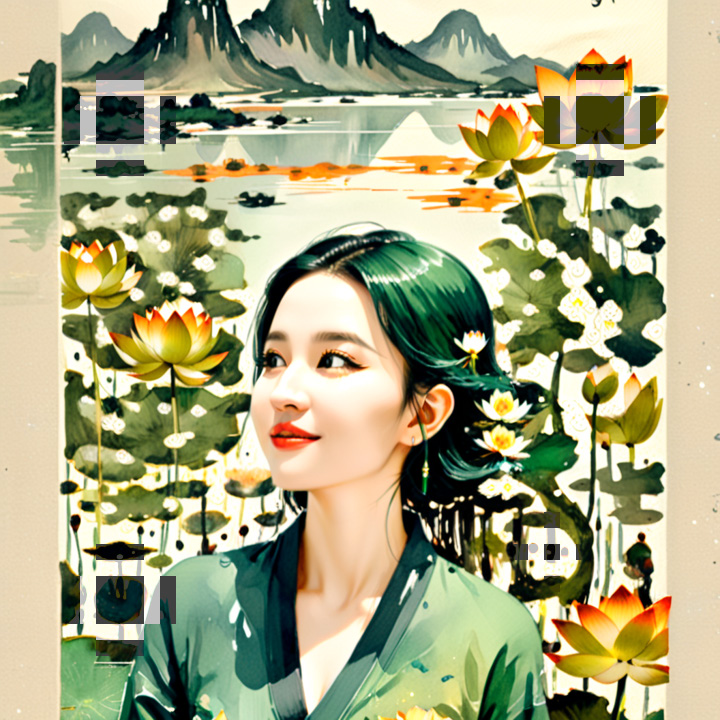} & 
    \includegraphics[width=\suppwidth, height=\suppwidth]{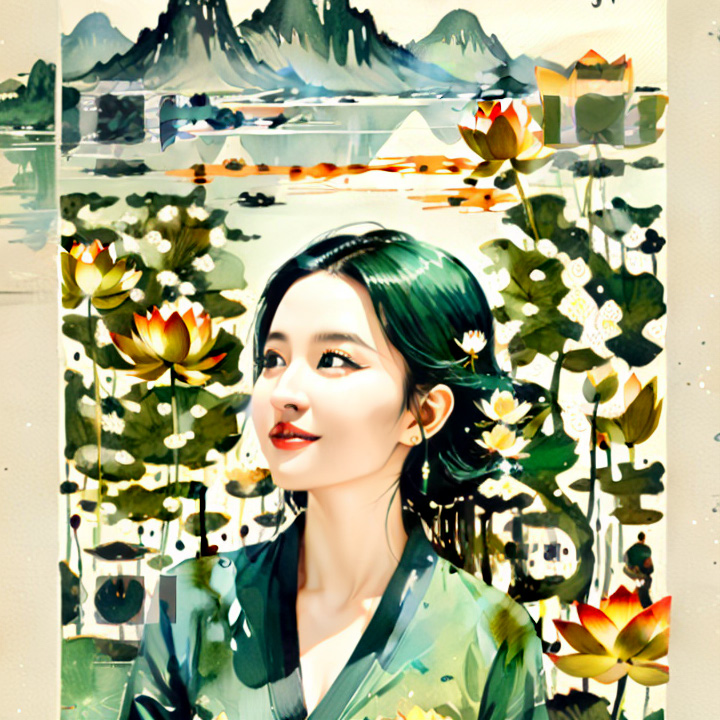} \\

    \includegraphics[width=\suppwidth, height=\suppwidth]{figures/lyf/lyf-0.png} & 
    \includegraphics[width=\suppwidth, height=\suppwidth]{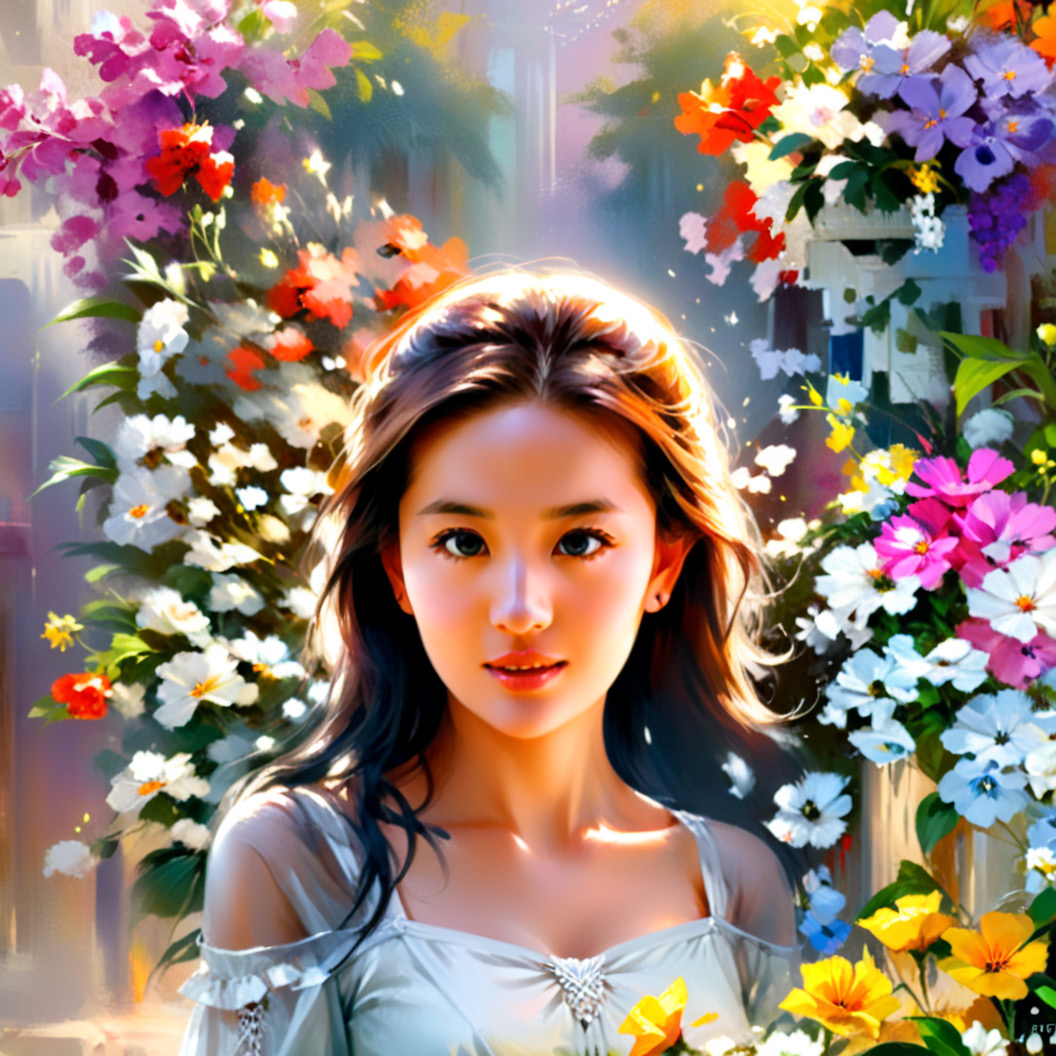} & 
    \includegraphics[width=\suppwidth, height=\suppwidth]{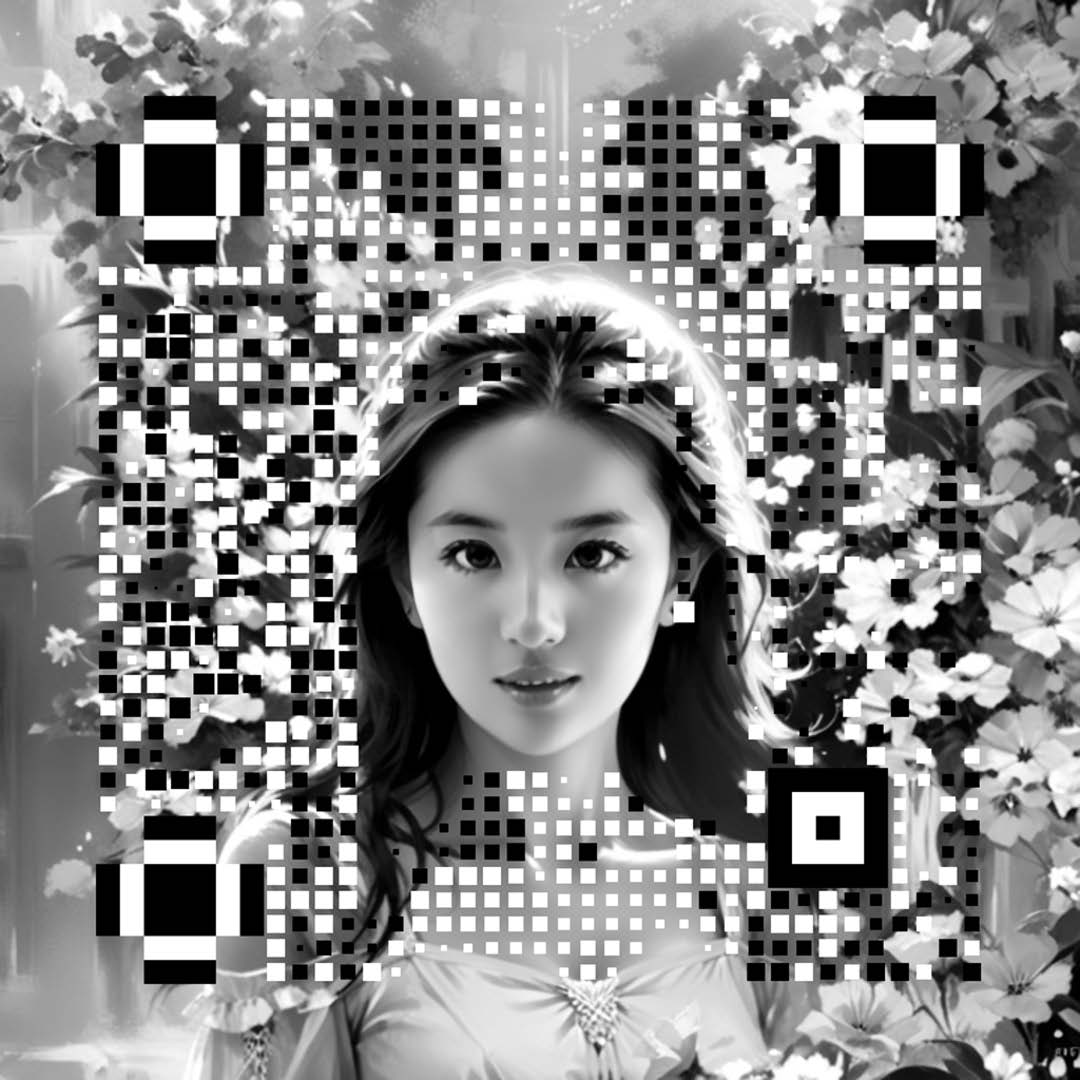} & 
    \includegraphics[width=\suppwidth, height=\suppwidth]{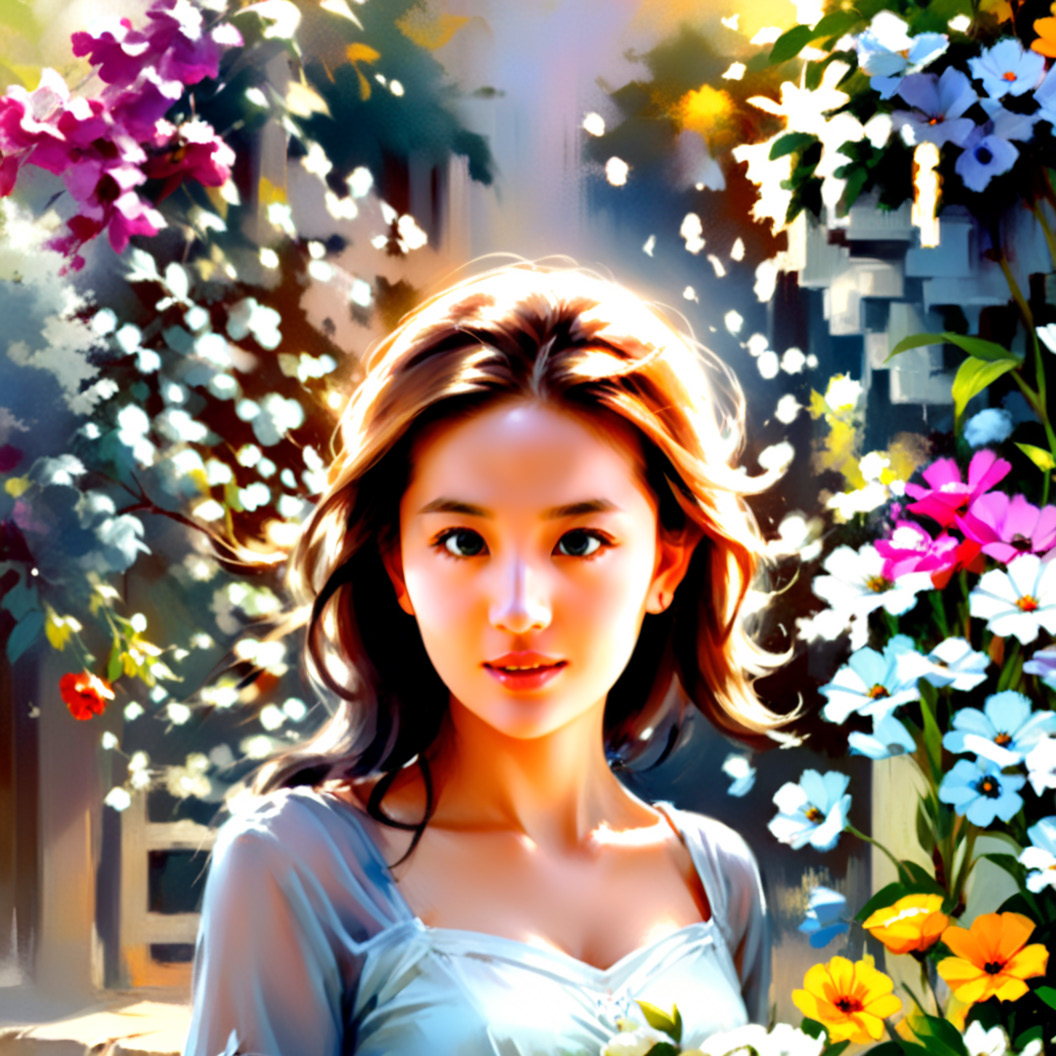} & 
    \includegraphics[width=\suppwidth, height=\suppwidth]{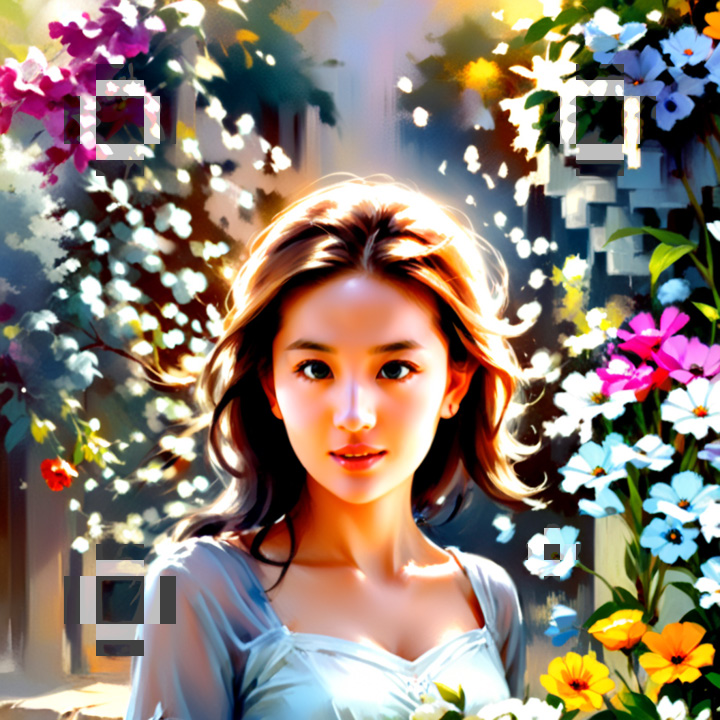} & 
    \includegraphics[width=\suppwidth, height=\suppwidth]{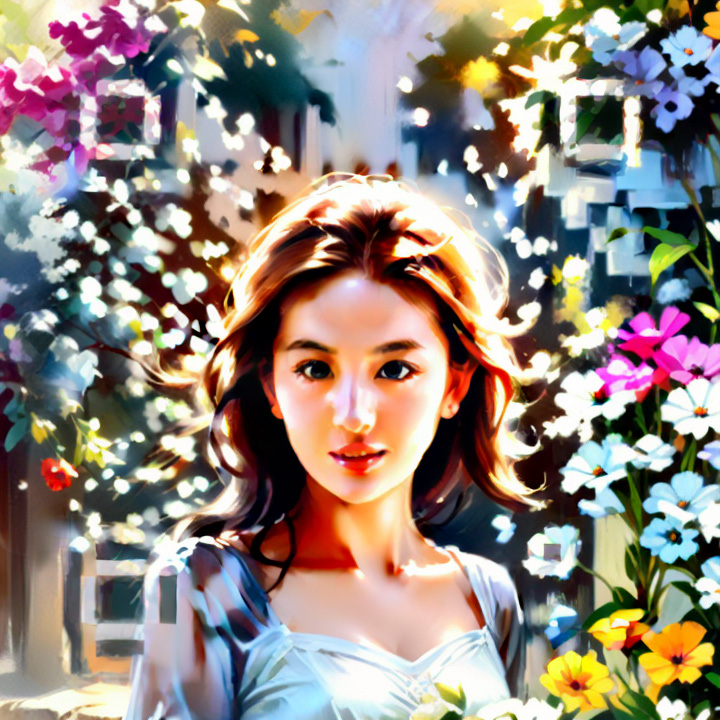} \\

    \includegraphics[width=\suppwidth, height=\suppwidth]{figures/lyf/lyf-90.png} & 
    \includegraphics[width=\suppwidth, height=\suppwidth]{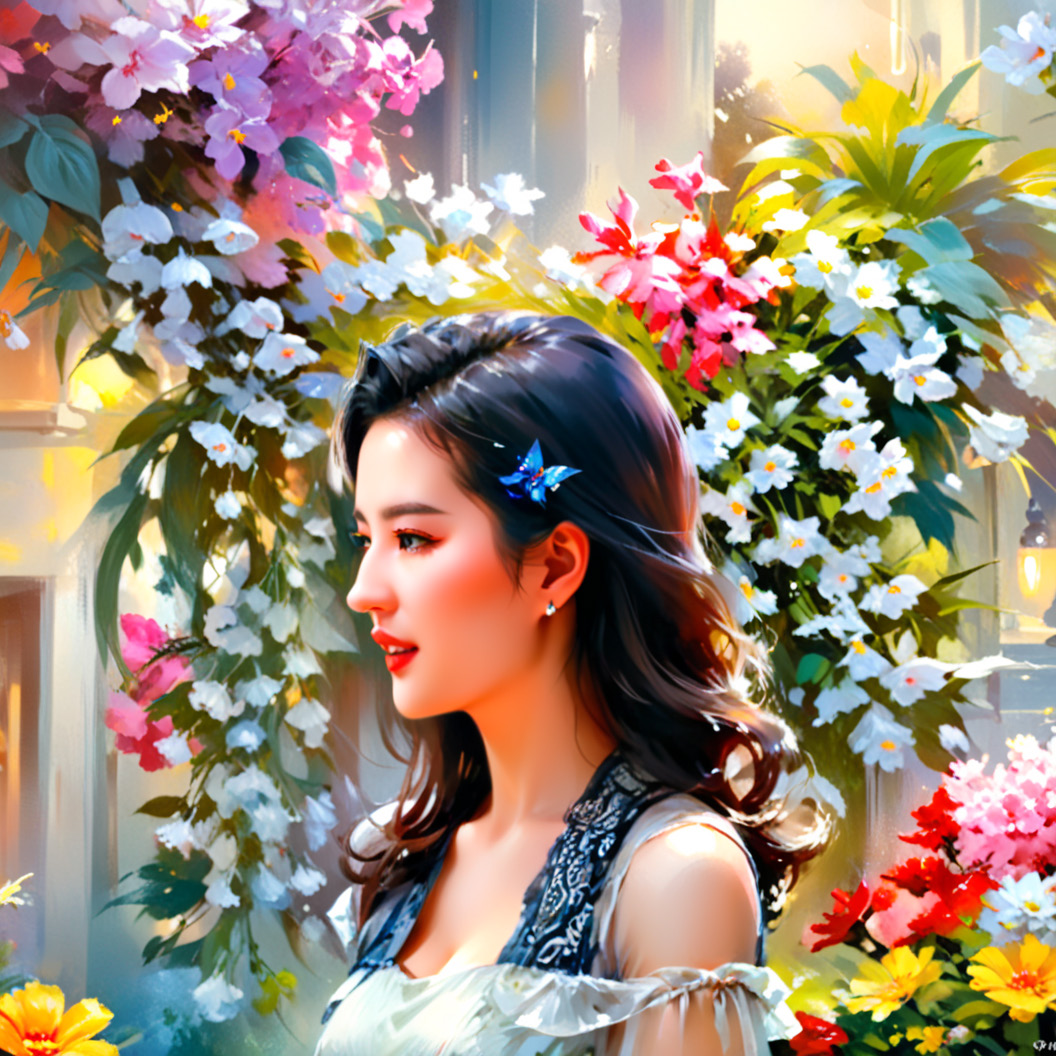} & 
    \includegraphics[width=\suppwidth, height=\suppwidth]{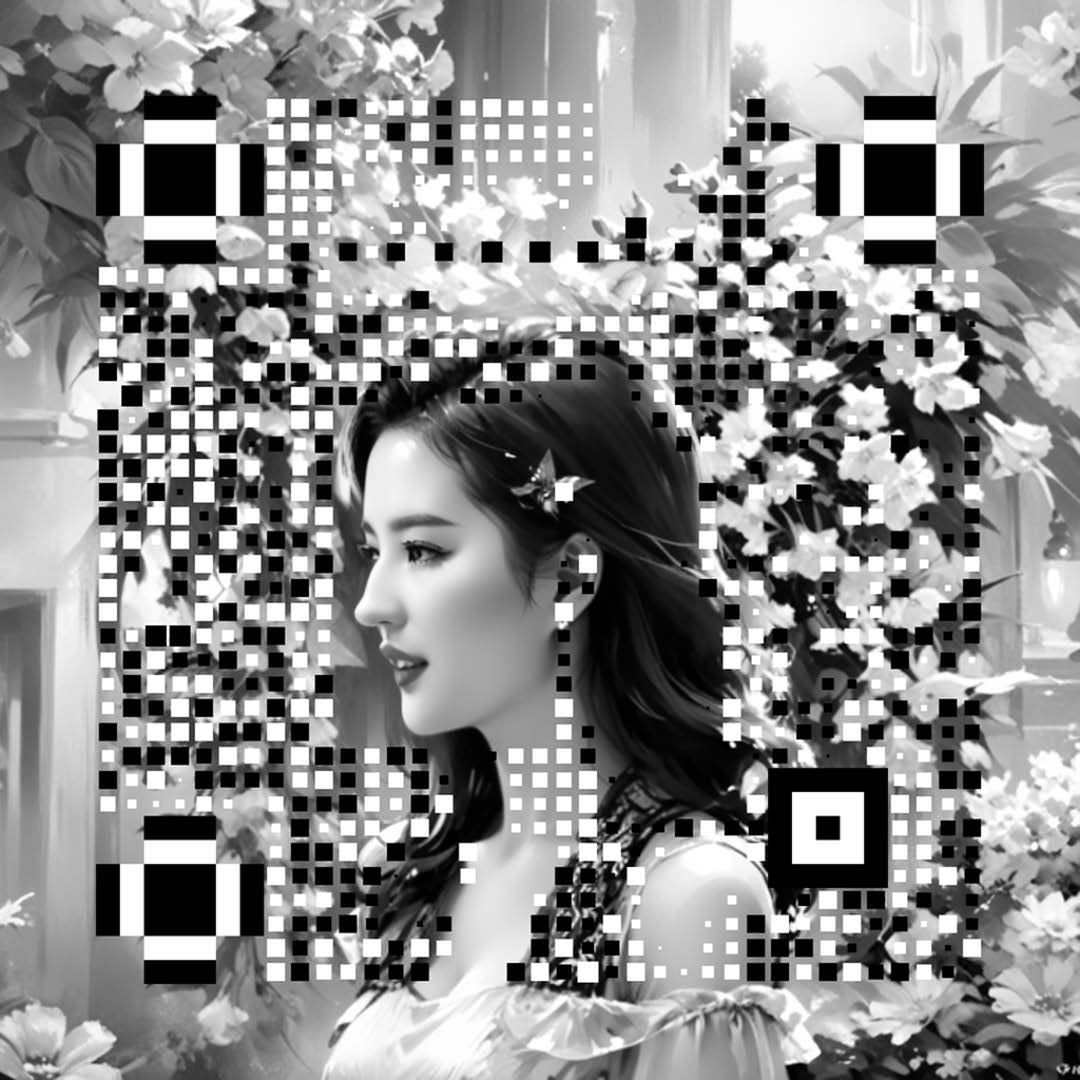} & 
    \includegraphics[width=\suppwidth, height=\suppwidth]{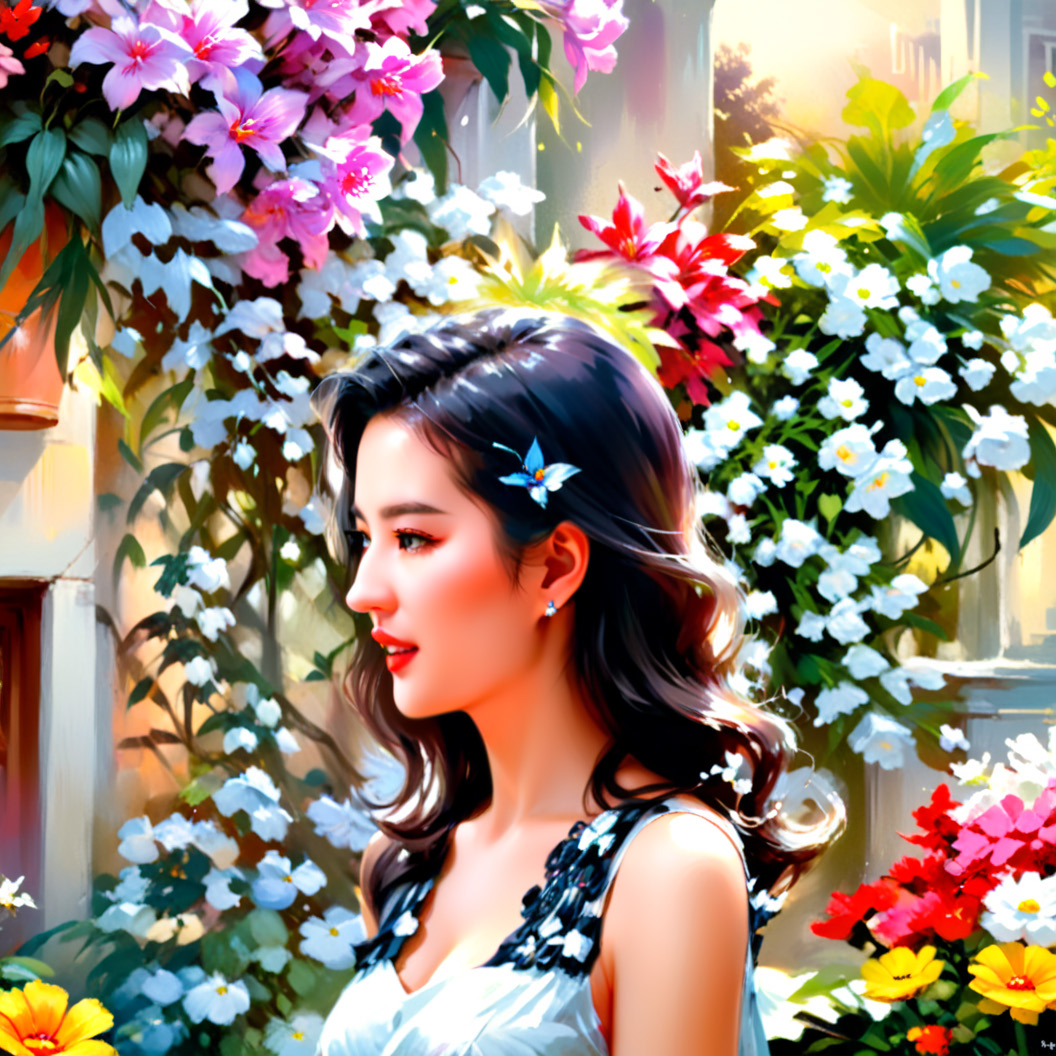} & 
    \includegraphics[width=\suppwidth, height=\suppwidth]{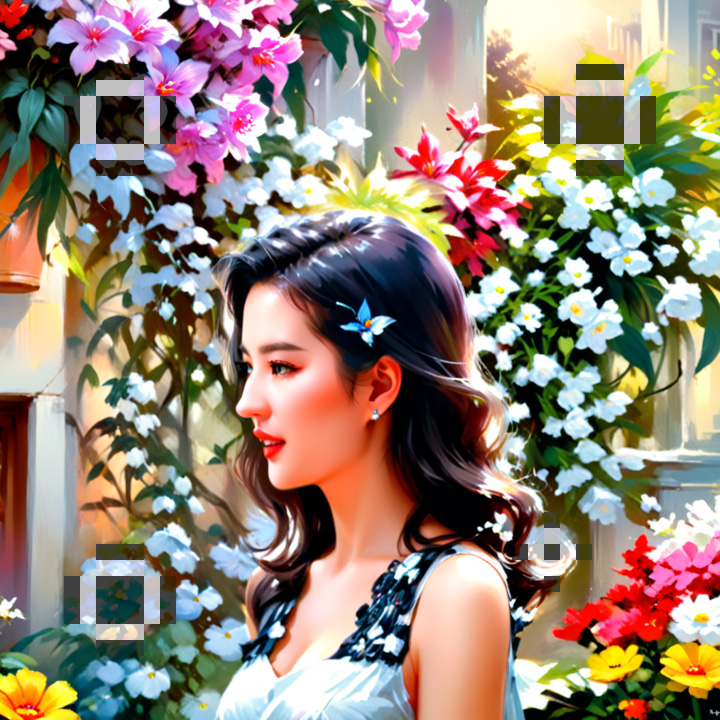} & 
    \includegraphics[width=\suppwidth, height=\suppwidth]{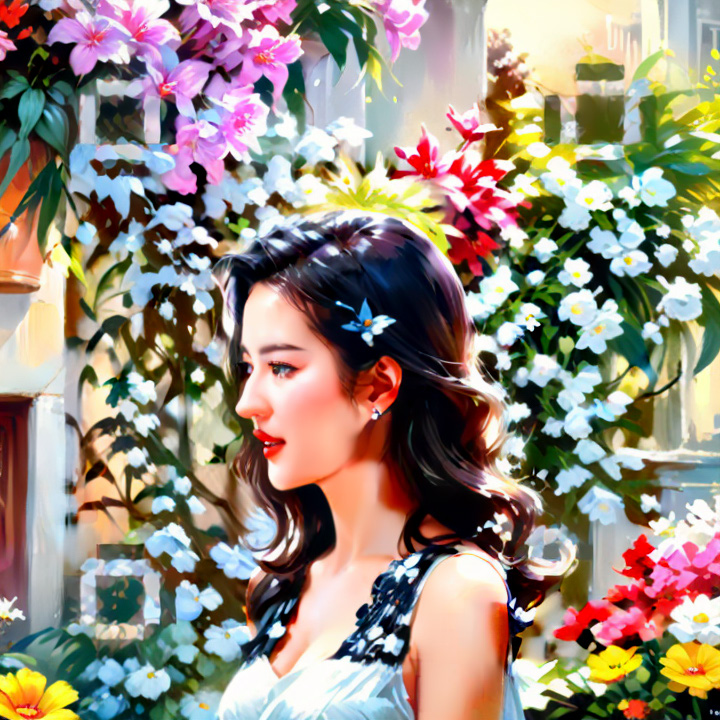} \\

    \bottomrule
    \end{tabularx}
    \vspace{-10pt}
    \label{pipeline_vis}
\end{table*}

\begin{table*}[tbp]
    \caption{Prompts for generated QR codes in the paper. All images are generated with size of 1,024$\times$1,024. The generative model is uniformly \href{https://civitai.com/models/84040/sdxl-unstable-diffusers-yamermix?modelVersionId=247214}{SDXL Unstable Diffusers YamerMIX}.}
    \label{tab:comparison3}
    \centering 
\begin{tabularx}{\textwidth}{
    >{\raggedright\arraybackslash}p{0.1\textwidth} 
    >{\raggedright\arraybackslash}p{0.9\textwidth} 
    >{\raggedright\arraybackslash}X 
    }

    \toprule
    Sample &  Prompt \\
    \midrule

    Figure~\ref{teaser} Col 1 & "A female woman, face in the middle. white bird sitting on a branch of roses, digital watercolor illustration of a meadow with white roses in the morning light, detailed fantastic background of Salvador Dali, waterhouse, Canaletto, watercolor art, intricate, complex contrast, HDR, sharp, soft cinematic volumetric lighting, the background is lost in haze. The foreground is brightly lit. 4k" \\
    \midrule

    Figure~\ref{teaser} Col 2 & "A male man, face in the middle. J. R. R. Tolkien-inspired landscape photo, a magical landscape inspired by J. R. R. Tolkien The Lord of the Rings, hilly path, bathed in a breathtaking play of sunlight splashing on surfaces, presents bark textures with color gradients in wood-earth tones, Jungle, mossy rock formations, complicated plants. HDR, Creating a photorealistic, asymmetrical composition, complicated details, very detailed, by Greg Rutkowski"\\
    \midrule

    Figure~\ref{teaser} Col 3 & "A female woman, face in the middle. olpntng style, ink wash in green and gold tones, Landscape of lotus flowers in the foreground over a lake, muted colours, wet on wet technique, sketch ink watercolor style with a hint of orange and white by Wu Guanzhong, Truong Lo, Mary Jane Ansell, Agnes Cecile, muted splatter art, gold ink splatter, faded dripping paints. green monochrome, soft impressionistic brushstrokes, oil painting, heavy strokes, dripping paint, oil painting, heavy strokes, paint dripping"\\
    \midrule

    Figure~\ref{teaser} Col 4 & "A male man, face in the middle. flat stylized pine trees, winter landscape with starry night sky and lake, painterly, acrylic painting, trending on pixiv fanbox, palette knife and brush strokes, style of makoto shinkai jamie wyeth james gilleard edward hopper greg rutkowski studio ghibli genshin impact"\\
    \midrule

    Figure~\ref{teaser} Col 5 & "A female woman, face in the middle. (best quality:1.5), (intricate emotional details:1.5), (sharpen details),  (ultra detailed), (cinematic lighting),   sorcerer's ancient library, ,floating candles, mystical artifacts, magical books, oxfort Key Elements:"\\
    \midrule

    Figure~\ref{teaser} Col 6 & "A male man, face in the middle. colorful birds sitting on top of a pink flower, fantasy, parrot by Adam MarczyÅ ski, fantasy art, art of alessandro pautasso, glowing oil,detailed beautiful animals, artwork in the style of guweiz"\\
    \midrule

    Table~\ref{tab:comparison} Row 1 & "A male man, face in the middle. painted clouds and landscape background, Watercolor, trending on artstation, sharp focus, studio photo, intricate details, highly detailed, by greg rutkowski"\\
    \midrule

    Table~\ref{tab:comparison} Row 2 & "A female woman, face in the middle. UHD, (masterpiece) Landscape of the Great Wall of China, smoke effects, trending on artstation, sharp focus, intricate details, highly detailed,"\\
    \midrule

    Table~\ref{tab:comparison} Row 3 & "A male man, face in the middle. A ocean of pastel pink blue and lilac ice cream, with a boat made of candy, waves"\\
    \midrule

    Table~\ref{fig:face} Col 2 & "A male man, face in the middle. Hatsune Mecha Tech Sense HD Wallpaper, ultra hd, realistic, vivid colors, highly detailed, UHD drawing, pen and ink, perfect composition, beautiful detailed intricate insanely detailed octane render trending on artstation, 8k artistic photography, photorealistic concept art, soft natural volumetric cinematic perfect light"\\
    \midrule

    Table~\ref{fig:face} Col 3 & "A male man, face in the middle. in the style of james gilleard, SamDoesArts, art by Sam Yang,  absolute beauty birth'd from fragile chaos, mandelbulb dress, insanely detailed, full of life, animated"\\
    \midrule

    Table~\ref{fig:face} Col 5 & "A male man, face in the middle. impressionist landscape of a Japanese garden in winter with a bridge over a pond"\\
    \midrule

    Table~\ref{pipeline_vis} Row 7\&8 & "A female woman, face in the middle. Highly detailed beautiful landscape, vintage style, bright colors, atmospheric lighting flowers, cinematic composition, digital painting, elegant, beautiful, high detail, by Willem Haenraets, trending on artstation, sharp focus, studio photo, intricate details, highly detailed, by greg rutkowski"\\
    \midrule

    \bottomrule
    \end{tabularx}
    \vspace{-10pt}
\end{table*}

\clearpage

\newpage
\section*{NeurIPS Paper Checklist}

\begin{enumerate}

\item {\bf Claims}
    \item[] Question: Do the main claims made in the abstract and introduction accurately reflect the paper's contributions and scope?
    \item[] Answer: \answerYes{} 
    \item[] Justification: In abstract, the main contributions of this paper are emphasized. Furthermore, in the last paragraph of the introduction, these contributions are clearly listed again.

\item {\bf Limitations}
    \item[] Question: Does the paper discuss the limitations of the work performed by the authors?
    \item[] Answer: \answerYes{} 
    \item[] Justification: Please refer to Section~\ref{sec:conclusion}.

\item {\bf Theory Assumptions and Proofs}
    \item[] Question: For each theoretical result, does the paper provide the full set of assumptions and a complete (and correct) proof?
    \item[] Answer: \answerNA{} 
    \item[] Justification: This paper does not include theoretical results.

    \item {\bf Experimental Result Reproducibility}
    \item[] Question: Does the paper fully disclose all the information needed to reproduce the main experimental results of the paper to the extent that it affects the main claims and/or conclusions of the paper (regardless of whether the code and data are provided or not)?
    \item[] Answer: \answerYes{} 
    \item[] Justification: We provide all needed information to reproduce the main experimental results of this paper in Section~\ref{sec:exp}. Our code will be released upon publication.

\item {\bf Open access to data and code}
    \item[] Question: Does the paper provide open access to the data and code, with sufficient instructions to faithfully reproduce the main experimental results, as described in supplemental material?
    \item[] Answer: \answerYes{} 
    \item[] Justification: We consider publishing the code at \url{https://github.com/cavosamir/Face2QR} once we complete our patent application process.

\item {\bf Experimental Setting/Details}
    \item[] Question: Does the paper specify all the training and test details (e.g., data splits, hyperparameters, how they were chosen, type of optimizer, etc.) necessary to understand the results?
    \item[] Answer: \answerYes{} 
    \item[] Justification: All experiments details are illustrated in Section~\ref{sec:exp4-1}.

\item {\bf Experiment Statistical Significance}
    \item[] Question: Does the paper report error bars suitably and correctly defined or other appropriate information about the statistical significance of the experiments?
    \item[] Answer: \answerNo{} 
    \item[] Justification: This paper mainly conducts qualitative comparisons and subjective experiments. Therefore, the corresponding error bars are not applicable.

\item {\bf Experiments Compute Resources}
    \item[] Question: For each experiment, does the paper provide sufficient information on the computer resources (type of compute workers, memory, time of execution) needed to reproduce the experiments?
    \item[] Answer: \answerYes{} 
    \item[] Justification: Computational resources have been described in Section~\ref{sec:exp4-1}.
    
\item {\bf Code Of Ethics}
    \item[] Question: Does the research conducted in the paper conform, in every respect, with the NeurIPS Code of Ethics \url{https://neurips.cc/public/EthicsGuidelines}?
    \item[] Answer: \answerYes{} 
    \item[] Justification: This work is conducted in accordance with the NeurIPS Code of Ethics.

\item {\bf Broader Impacts}
    \item[] Question: Does the paper discuss both potential positive societal impacts and negative societal impacts of the work performed?
    \item[] Answer: \answerYes{} 
    \item[] Justification: Please refer to the Section~\ref{sec:conclusion}.
    
\item {\bf Safeguards}
    \item[] Question: Does the paper describe safeguards that have been put in place for responsible release of data or models that have a high risk for misuse (e.g., pretrained language models, image generators, or scraped datasets)?
    \item[] Answer: \answerNA{} 
    \item[] Justification: The paper poses no such risks.

\item {\bf Licenses for existing assets}
    \item[] Question: Are the creators or original owners of assets (e.g., code, data, models), used in the paper, properly credited and are the license and terms of use explicitly mentioned and properly respected?
    \item[] Answer: \answerYes{} 
    \item[] Justification: The creators or original owners of assets (e.g., code, data, models), used in the paper, are properly credited, and the license and terms of use are explicitly mentioned and are properly respected.

\item {\bf New Assets}
    \item[] Question: Are new assets introduced in the paper well documented and is the documentation provided alongside the assets?
    \item[] Answer: \answerYes{} 
    \item[] Justification: The new assets introduced in the paper are well documented alongside the assets.

\item {\bf Crowdsourcing and Research with Human Subjects}
    \item[] Question: For crowdsourcing experiments and research with human subjects, does the paper include the full text of instructions given to participants and screenshots, if applicable, as well as details about compensation (if any)? 
    \item[] Answer: \answerYes{} 
    \item[] Justification: This paper includes the full text of instructions given to participants and screenshots, and the human subjects are paid at least the minimum wage in the country of the data collector, following the NeurIPS Code of Ethics. 

\item {\bf Institutional Review Board (IRB) Approvals or Equivalent for Research with Human Subjects}
    \item[] Question: Does the paper describe potential risks incurred by study participants, whether such risks were disclosed to the subjects, and whether Institutional Review Board (IRB) approvals (or an equivalent approval/review based on the requirements of your country or institution) were obtained?
    \item[] Answer: \answerYes{} 
    \item[] Justification: There is no such potential risks aware for research with human subjects in this paper. We have obtained the IRB approval and also adhere to the NeurIPS Code of Ethics.

\end{enumerate}

\end{document}